\documentclass[twoside,leqno,twocolumn,english]{article}
\usepackage{ltexpprt}

\usepackage{times}
\usepackage{graphicx}
\usepackage{array,booktabs,ragged2e}
\usepackage[tight,footnotesize]{subfigure}

\usepackage{algorithm}
\usepackage{algorithmic}
\usepackage{color}
\usepackage{amsmath}
\usepackage{amsfonts}
\usepackage{epsfig}
\usepackage{url}
\usepackage{multirow}
\usepackage{setspace}
\usepackage{wrapfig}
\usepackage{comment}
\usepackage{float}
\usepackage{tabularx}

\newcommand{\etal}{{\em et al.}}

\newcommand{\hide}[1]{}

\newcommand{\method}{{\sc RobustMultiSC}}
\newcommand{\problem}{{\sc RSCM Problem}}

\newcommand{\bit}{\begin{itemize}}
	\newcommand{\eit}{\end{itemize}}
\newcommand{\ben}{\begin{enumerate}}
	\newcommand{\een}{\end{enumerate}}
\newcommand{\beq}{\begin{equation}}
\newcommand{\eeq}{\end{equation}}

 \newtheorem{definition}{Definition}
 
\newcommand{\bfyl}{\mathbf{y_L}}
\newcommand{\bfyu}{\mathbf{y_U}} 
\newcommand{\bffl}{\mathbf{f_L}}
\newcommand{\bffu}{\mathbf{f_U}} 
\newcommand{\bfy}{\mathbf{y}}
\newcommand{\bff}{\mathbf{f}}
\newcommand{\bfw}{\mathbf{w}}
\newcommand{\mg}{\mathcal{G}}
\newcommand{\mgs}{\mathcal{GS}}
\newcommand{\bWk}{\mathbf{W_k}}
\newcommand{\bD}{\mathbf{D_k}}
\newcommand{\bLk}{\mathbf{L_k}}
\newcommand{\bLi}{\mathbf{L_i}}
\newcommand{\bI}{\mathbf{I}}
\newcommand{\bC}{\mathbf{C}}
\newcommand{\bxi}{\mathbf{\xi}}
\newcommand{\bbeta}{\mathbf{\beta}}

\newcommand{\tss}{{\sc TSS}}
\newcommand{\rob}{{\sc RobustLP}}
\newcommand{\gene}{{\sc GeneMania}}
\newcommand{\eql}{{\sc Eql-Wght}}
\newcommand{\perf}{{\sc Perf-Wght}}

\newcommand{\LINEIF}[2]{\STATE \algorithmicif\ {#1}\ \algorithmicthen\ {#2} }

\renewcommand{\algorithmicrequire}{\textbf{Input:}}
\renewcommand{\algorithmicensure}{\textbf{Output:}}

\title{Robust Semi-Supervised Classification for Multi-Relational Graphs}

\begin{document}

\author{
\makebox[230pt]{Junting Ye \quad \quad \quad  Leman Akoglu}\\
\makebox[230pt]{Stony Brook University}\\
\makebox[230pt]{Department of Computer Science}\\
\makebox[230pt]{\{juyye, leman\}@cs.stonybrook.edu} \\
\vspace{-0.3in}
}
\date{}
\maketitle

\begin{abstract}

Graph-regularized semi-supervised learning has been used effectively for classification
when ($i$) instances are connected through a graph, and ($ii$) labeled data is scarce.
If available, using \textit{multiple} relations (or graphs) between the instances can improve the prediction performance.
On the other hand, when these relations have varying levels of veracity and exhibit varying relevance for the task,
very noisy and/or irrelevant relations may deteriorate the performance.
As a result, an effective weighing scheme needs to be put in place.

In this work, we propose a robust and scalable approach for multi-relational graph-regularized semi-supervised classification.
Under a \textit{convex optimization scheme}, we simultaneously infer weights for the multiple graphs as well as a solution.
We provide a careful analysis of the inferred weights, based on which we devise an algorithm that filters out irrelevant and noisy graphs and produces \textit{weights proportional to the informativeness} of the remaining graphs. 
Moreover, the proposed method is \textit{linearly scalable} w.r.t. the number of edges in the union of the multiple graphs.
Through extensive experiments we show that our method yields superior results under different noise models, and under increasing number of noisy graphs and intensity of noise, as compared to a list of baselines and state-of-the-art approaches. 

%
\end{abstract}

\vspace{-0.05in}
\section{Introduction}
\label{sec:intro}

%
%
%
%
%

Given (1) a graph with \textit{multiple} different relations between its nodes, and (2) labels for a small set of nodes, how can we predict the labels of the unlabeled nodes in a \textit{robust} fashion?
Robustness is a key element especially when the data comes from sources with varying veracity, where some relations may be irrelevant for the prediction task or may be too noisy.

\begin{figure}[!t]
\centering
\begin{tabular}{ccc}
\hspace{-0.15in}\includegraphics[width=0.35\linewidth]{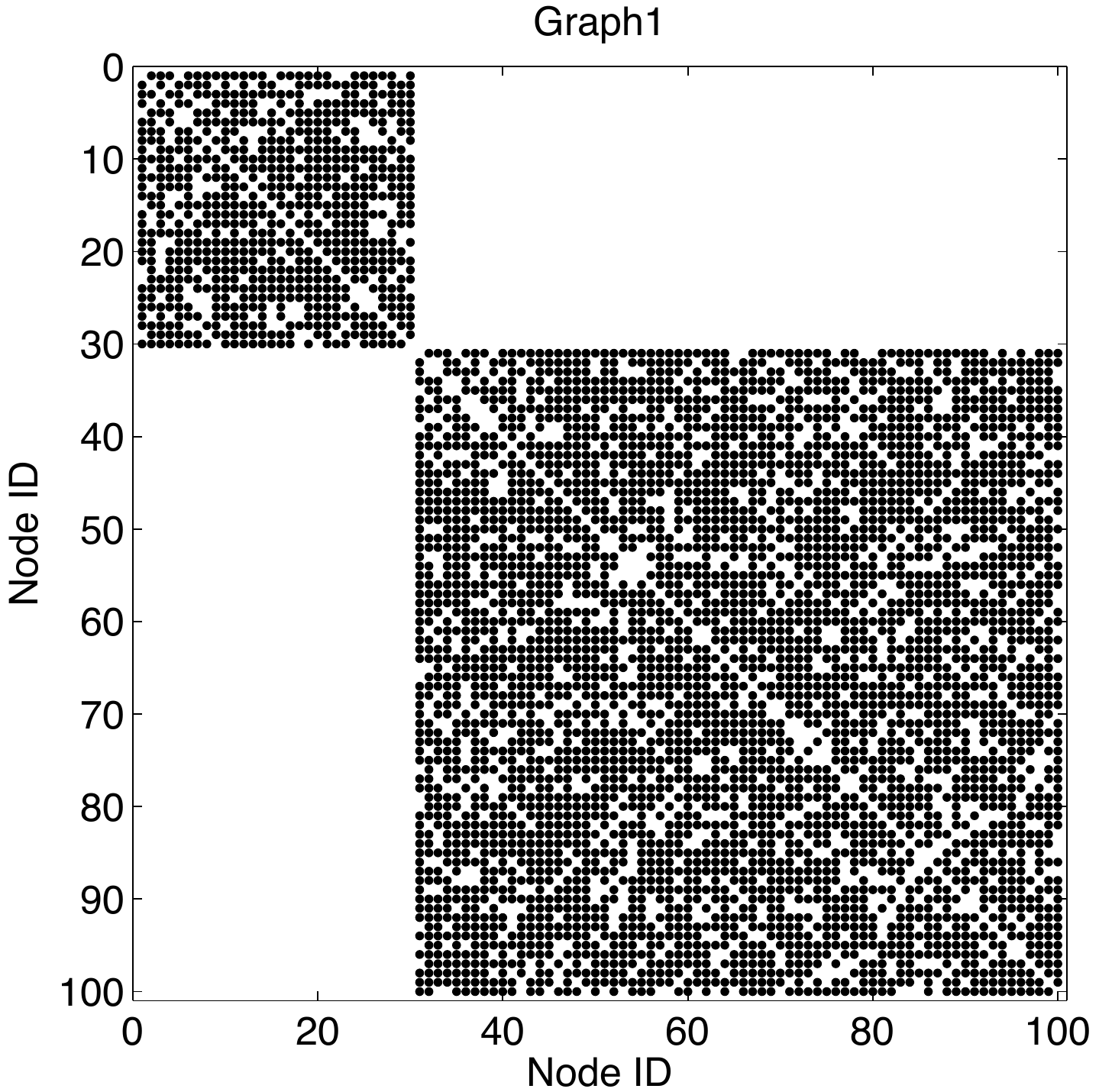} &
\hspace{-0.15in}\includegraphics[width=0.34\linewidth]{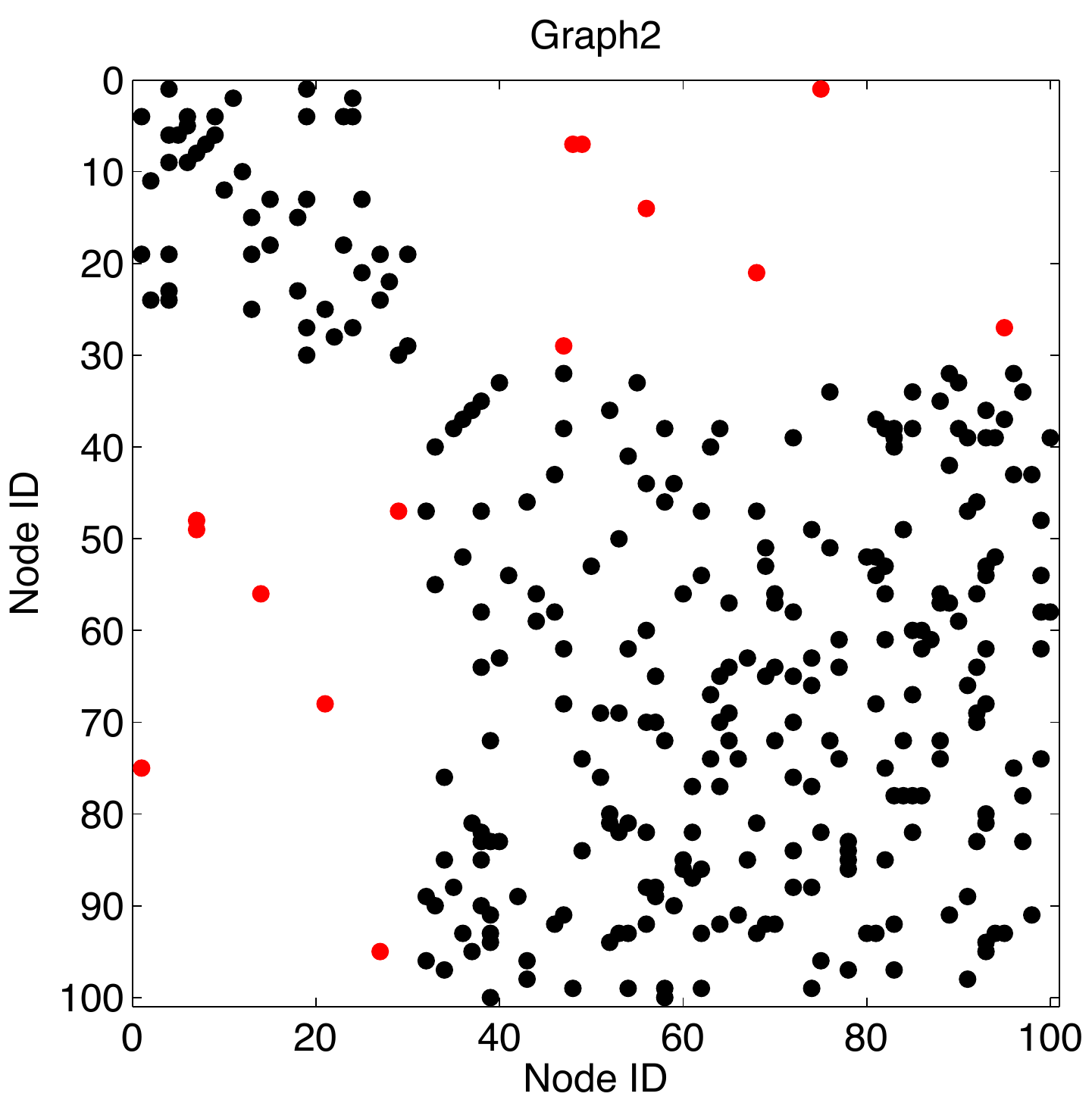} & 
\hspace{-0.15in}\includegraphics[width=0.34\linewidth]{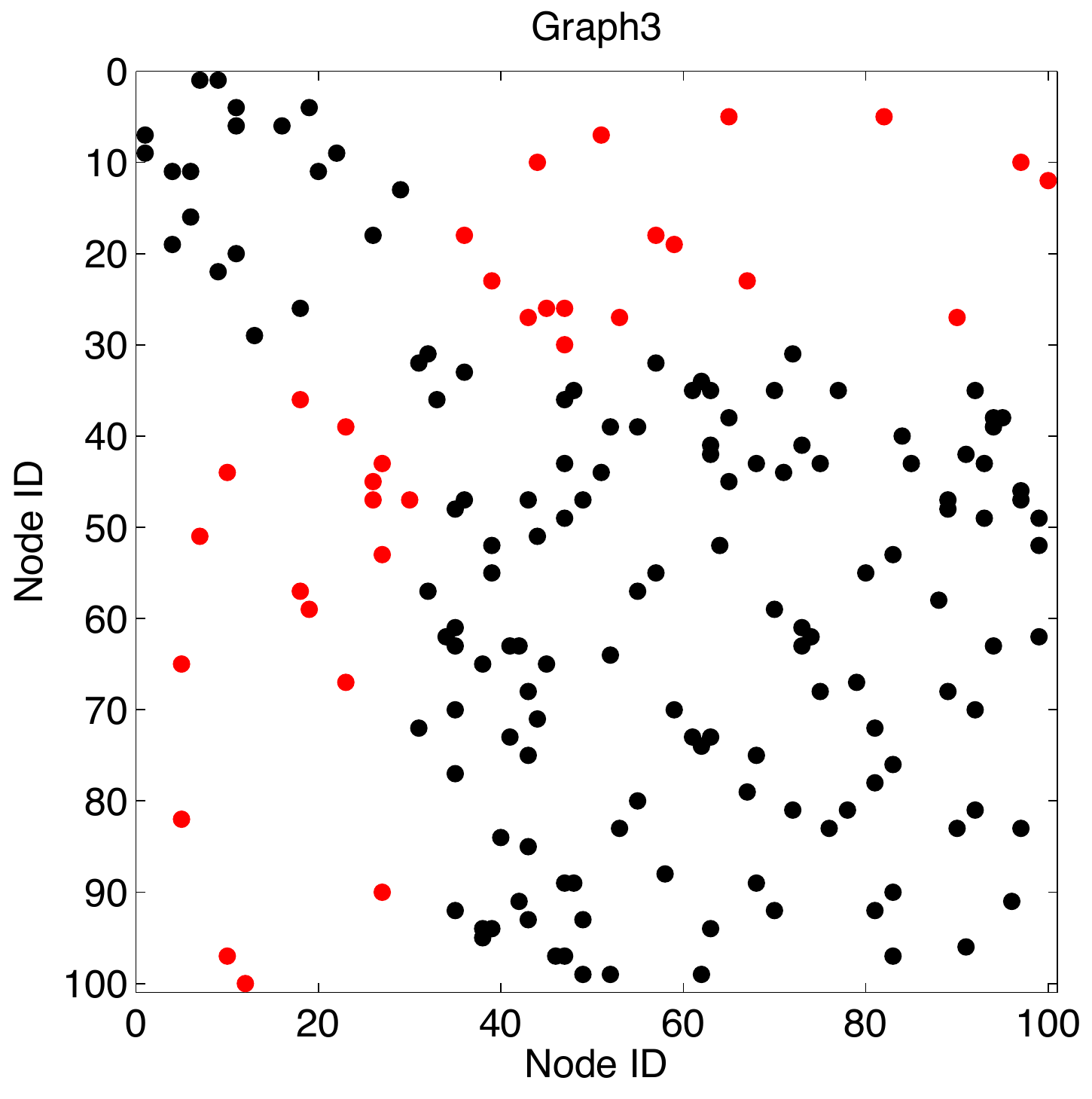} \\
\end{tabular}
\begin{tabular}{cc}
\hspace{-0.15in}\includegraphics[width=0.35\linewidth]{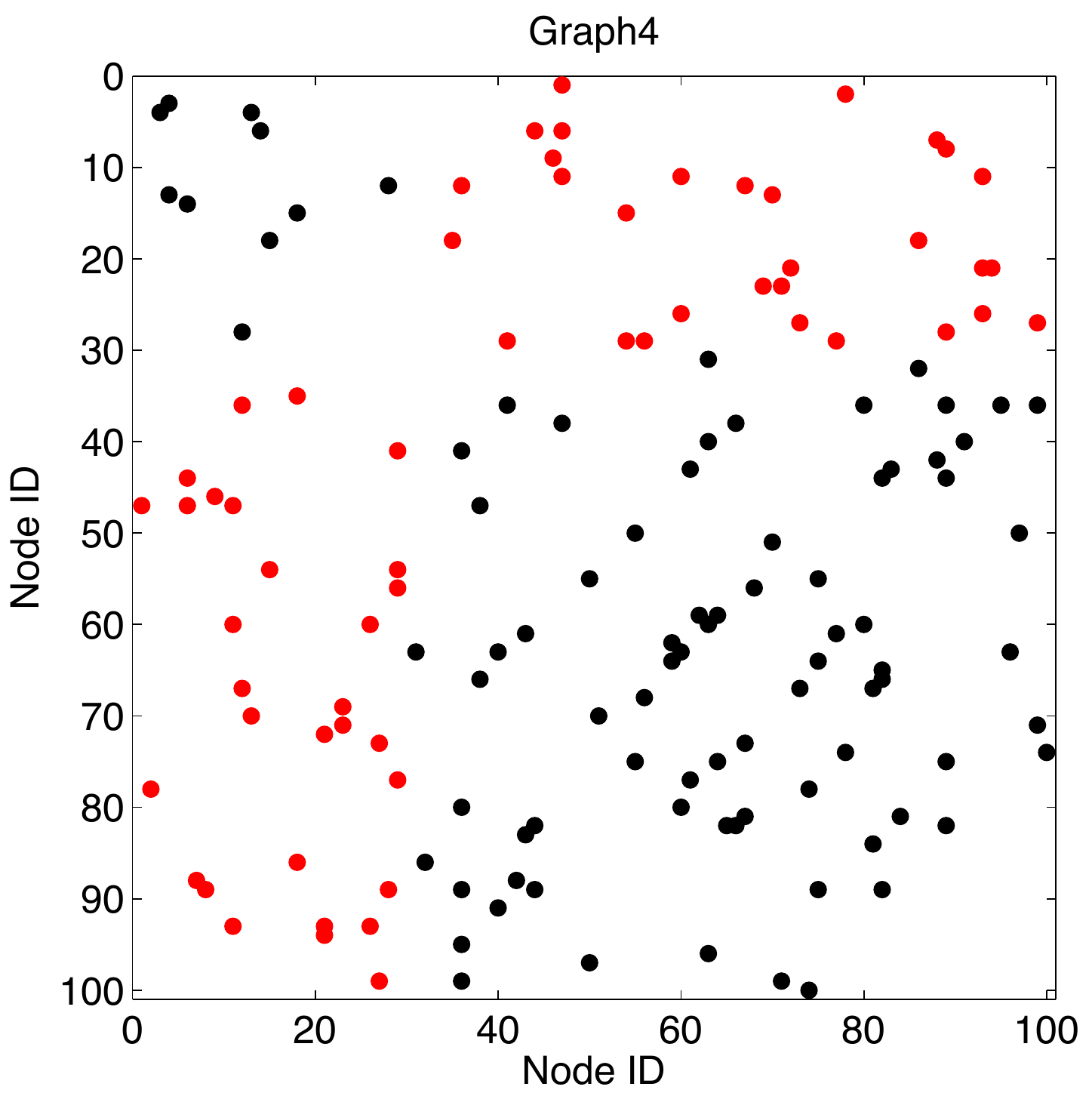} &
\hspace{-0.15in}\includegraphics[width=0.35\linewidth]{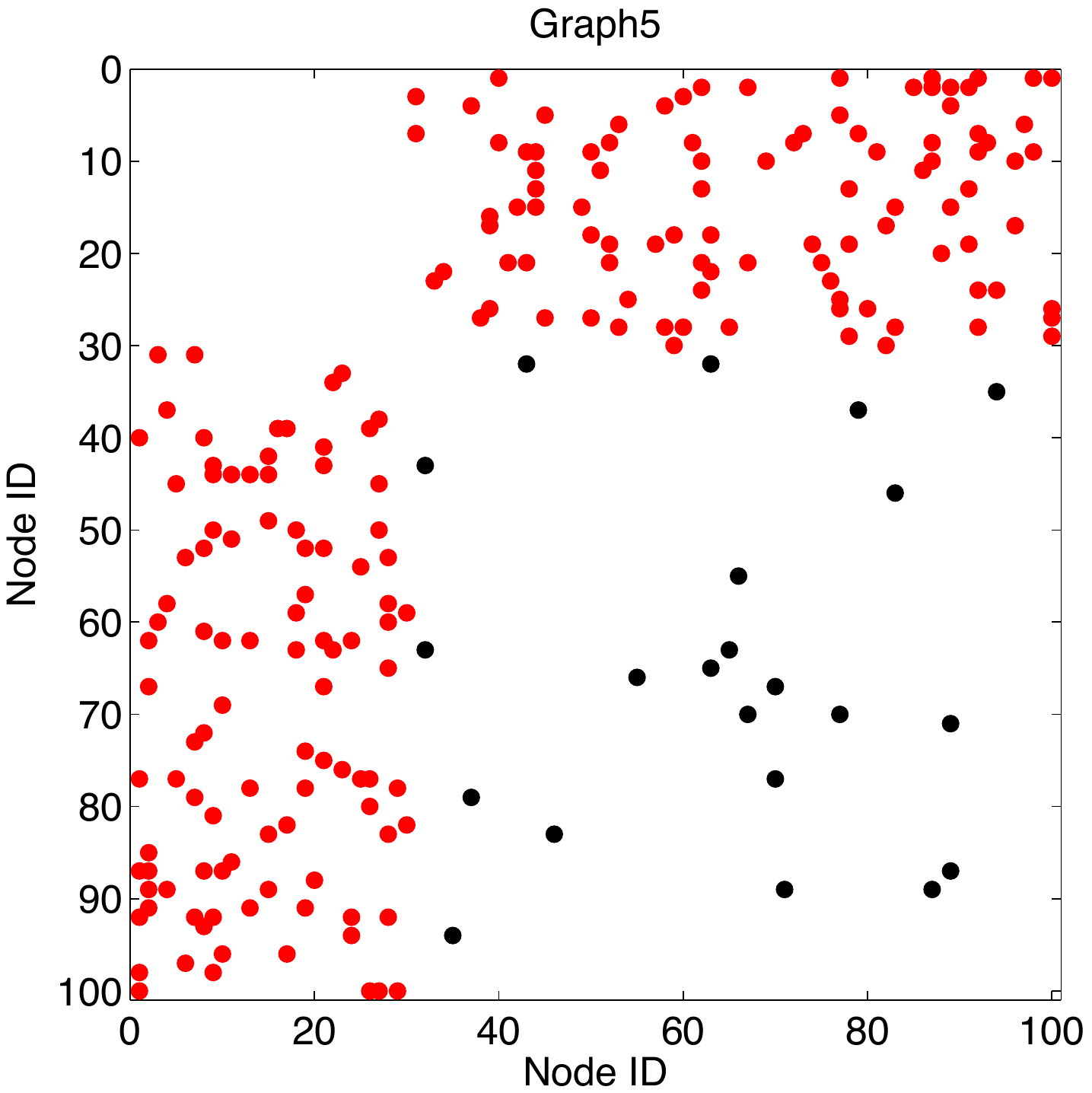} \\
\end{tabular}
\vspace{-0.1in}
\caption{\small Example multi-graph ($n=100$, $m=5$), with 3 informative (top) and 2 intrusive (bottom) graphs. Shown are adjacency matrices, red dots depict cross-edges between nodes from different classes. 
$G_1$--$G_3$ are in order of informativeness. $G_4$ depicts random noise. $G_5$ contains adversarial noise.
Inferred weights (all graphs): $[25.17, 16.54, 12.79, 17.82, 27.68]$ (AP = 0.734$\pm$0.035).
Weights after noisy graphs removed: $[0.5000, 0.3003, 0.1997, 0, 0]$ (AP = 0.974$\pm$0.004). AP: average precision.
 \label{fig:toy}}
\vspace{-0.1in}
\end{figure}

This abstraction admits various real-world applications. For example, in fraud detection one may try to classify individuals as fraudulent or not based on the phone-call, SMS, financial, etc. interactions between them. 
In biology, genes are classified as whether or not they perform a certain function through various similarity relations between them. 
An example is shown in Figure \ref{fig:toy}, where a multi-graph with five different relation types $G_1$-$G_5$ is depicted.

Accomplishing the above task requires addressing two main problems: (1) identifying and filtering out irrelevant and noisy relations, and (2) automatically weighing other relations by their informativeness for the task.
Existing methods either are vastly affected in the presence of noise \cite{tsuda2005fast}, produce locally optimal solutions due to their non-convex objective formulations \cite{journals/tnn/KatoKS09,journals/eswa/ShinTS09,conf/sdm/Wan15}, or are too expensive to compute \cite{conf/nips/ArgyriouHP05,LanckrietBCJN04} (See Section \ref{sec:related}).


In this work we introduce \method, a robust, scalable, and effective semi-supervised classification approach for multi-relational graphs. In the example case, it recognizes $G_4$ and $G_5$ as irrelevant/noisy, and estimates optimal weights for relations $G_1$-$G_3$ so as to leverage them collectively to achieve improved performance (See Figure \ref{fig:toy} caption).
Our contributions are listed as follows.

\vspace{-0.1in}
\bit
\setlength{\itemsep}{-0.75\itemsep}

\item {\bf Model formulation:} Under a {convex} formulation, we simultaneously estimate weights for the multiple relations (or graphs) as well as a solution that utilizes a weighted combination of them.

\item {\bf Analysis of weights:} We show that in the presence of noise, the inferred weights reflect the impact of different relations on the solution, where  both dense informative and irrelevant/noisy graphs receive large weights. 

\item {\bf Robustness:}  Analysis of weights enable us to devise a robust algorithm that filters out irrelevant/noisy graphs, so as to produce weights proportional to the informativeness of relevant graphs.  

\item {\bf Scalability:} Our proposed approach scales linearly w.r.t. the number of edges in the combined graph.

\item {\bf Effectiveness:} We show the efficacy of \method~on real-world datasets with varying number of relevant/noisy graphs, 
under different noise models, and varying intensity of noise, where it outperforms five baseline approaches including the state-of-the-art. 
\eit
\vspace{-0.075in}

\section{Problem Definition}
\label{sec:overview}

In this work we consider real-world problem settings in which 
(1) the problem is cast as a binary classification task, 
(2) data objects are related through multiple different relationships, and 
(3) ground-truth class labels are scarce.


Motivating examples of this setting can be found particularly in anomaly detection (fraud, spam, etc.) and biological applications.
For example, in bank fraud the goal is to classify users, i.e., account holders, as fraudulent or benign.
These users may relate to one another through various relations, e.g., phone calls, emails, social links, shared loans, etc.  
Moreover, known labels especially for the positive class is scarce since fraud is rare.
Besides, labeling a user as a fraudster or benign is a tedious process for domain analysts.

Similarly in biology, the problem of predicting gene function can be cast as a binary classification problem, in which a gene exhibits a certain function or not \cite{Mostafavi08}. The genes can be associated in multiple ways through heterogeneous sources of genomic and proteomic data. For example, function tends to be shared among genes with similar patterns of expression, similar three-dimensional structures, similar chemical sensitivity, with products that interact physically, etc. \cite{conf/psb/LanckrietDCJN04}.
Moreover, the labeled data is small, as the fraction of genes known to have a given function is relatively much smaller than the list of all genes.
In general, obtaining ground-truth labels in biological applications is often challenging due to the substantial effort required to verify annotations through physical lab experiments.

Overall, the data can be represented as a \textit{multi-graph}, in which the nodes represent the data objects (e.g., account holders or genes) and multiple sets of undirected edges between these nodes capture associations implied by the particular relations.
We give a formal definition as follows. 

\vspace{-0.025in}
\begin{definition}[Multi-Graph]
\sloppy{
A multi-relational graph (or a multi-graph) $\mg(V,\mathcal{E})$ consists of a set of graphs $\{G_1(V,E_1), \; G_2(V,E_2), \; \ldots, \; G_m(V,E_m)\}$, on the same node set $V$, $|V|=n$. Undirected (weighted) edges $\mathcal{E}= \{E_1, \ldots, E_m\}$ correspond to links implied by $m$ different relation types, where we denote $|\mg|=m$. 
}
\end{definition}

%

Using the various relationships between the data objects may provide more information for a given classification task, especially in the face of sparse input labels.
Collectively, more accurate predictions can be made by combining these \textit{multiple association networks}.
On the other hand, it is not realistic to assume that
all available relationships (i.e., graphs) between the data objects would be relevant for each particular prediction task. Relations that are either  ($i$) too noisy, 
or ($ii$)  irrelevant to a task
should be significantly down-weighted. 
In this paper, we simply refer to all such graphs as \textit{intrusive}.
Filtering intrusive relations is especially important when the data sources cannot be carefully controlled---for example, when data is collected from various Web repositories with varying veracity, when graphs are constructed using various (not carefully chosen) similarity functions between the objects, etc.
In addition, the relevant graphs may have varying degree of informativeness for a task, which necessitates a careful weighing scheme. 

Overall, it is essential to build robust classification models that can effectively leverage multiple relationships by carefully weighing relevant graphs  while  filtering out the intrusive ones.
Our work addresses this problem of Robust Semi-supervised Classification  for Multi-Graphs ({\sc RSCM}): Given a binary classification task, a multi-graph, and a (small) set of labeled objects, the goal is to build an effective classifier that is \textit{robust to noisy and irrelevant data}.
We give the formal problem definition as follows.

\begin{definition}[\problem]
\label{def:problem}
Given a multi-graph $\mg(V,\mathcal{E})$, $|\mg|=m$, and
a subset of labeled seed nodes $L \subset V$;
devise a learning procedure to infer the labels of unlabeled nodes $V\backslash L$, which assigns an optimal set of weights $\bfw = \{w_1, \ldots, w_m\}$ to individual graphs where
($i$)  intrusive graphs are filtered (i.e., $w_k=0$), and
($ii$) relevant graphs receive weights relative to their informativeness.
\end{definition}

\section{Robust SsC for Multi-Graphs: Formulation}
\label{sec:proposed}

%
%
%

%
%

In this section we describe our formulation for the \problem~and later provide a learning procedure in Section \ref{sec:algo}. Our formulation particularly focuses on the inference and interpretation of individual graph weights $\bfw = \{w_1, \ldots, w_m\}$ for the different relations. Following on Definition \ref{def:problem} we first provide notation used in our formulation.

\vspace{0.05in}
\paragraph{Notation.$\;$}
Let $L = \{v_1, \ldots, v_l\}$ denote the set of labeled nodes, and $U = \{v_{l+1}, \ldots, v_n\}$ denote the set of unlabeled nodes.
We define a vector $\bfy = (y_1, \ldots, y_l, y_{l+1}, \ldots, y_n)$, where
$y_i \in \{+1, -1\}$ for $1\leq i\leq l$, indicating node belongs to positive or negative class, and $y_j = 0$ for $l+1\leq j \leq n$ for unlabeled nodes.

The estimation of $y_i$ from our model is denoted by $f_i$, $i=\{1,\ldots,n\}$.
Then, $\bffl=(f_1, \ldots, f_l)$ denotes estimations of $\bfyl$, and 
$\bffu=(f_{l+1}, \ldots, f_n)$ denotes estimations of $\bfyu$; where $\bff = (\bffl, \bffu)$.

Given a weighted graph $G_k(V,E_k)$, we denote its adjacency matrix by $\bWk$ and its
Laplacian matrix by $\bLk = \bD^{-\frac{1}{2}}(\bD - \bWk)\bD^{-\frac{1}{2}}$,
where $\bD$ is a diagonal matrix, in which $\bD(i,i) = \sum_j \bWk(i,j)$ and 0 elsewhere.

Throughout text, we use lowercase bold letters for vectors, uppercase bold letters to denote matrices, and plain font for scalars. We use apostrophe (e.g., $\bff'$) to denote transpose.

\subsection{Graph-Regularized Semi-supervised Classification}

Generalizing traditional semi-supervised learning objectives \cite{zhu2003semi} to multi-graphs, we can write
\vspace{-0.1in}
\beq
\label{generic}
\min_\bff \|\bff-\bfy\|_2^2 + \lambda \sum_{k=1}^m \bff'\bLk \bff
\eeq
where $\lambda$ is a regularization parameter. The first term enforces {\em both} fit to the training data {\em and} small norm on the estimation of unlabeled nodes, as we can write

\vspace{-0.25in}
$$\| \bff-\bfy \|_2^2 = (\bff-\bfy)'(\bff-\bfy) = \sum_{i=1}^{l} (f_i-y_i)^2 + \sum_{i=l+1}^{n} f_i^2 \;,$$
since $y_i=0$ for unlabeled nodes.
The second term enforces smoothness with respect to the composite graph structure, since

\vspace{-0.25in}
$$
\sum_{k=1}^m \bff'\bLk \bff = \sum_{k=1}^m \sum_{i,j\in V} 
\bWk(i,j) (\frac{f_i}{\sqrt{\bD(i,i)}}  - \frac{f_j}{\sqrt{\bD(j,j)}})^2 \;.
$$
\vspace{-0.1in}

Taking the derivative w.r.t. $\bff$ and setting it to zero, we obtain the solution
$$
\bff^* = (\bI+\lambda \sum_k \bLk)^{-1} \bfy \;.
$$ 
\vspace{-0.1in}

In the above derivation, all the graphs have equal impact on solution $\bff$. To enable different impact (based on informativeness), one can instead use a \textit{weighted} sum of the Laplacians in \eqref{generic} and obtain 
$$
\bff^* = (\bI+\lambda \sum_k w_k \bLk)^{-1} \bfy \;.
$$ 
\vspace{-0.15in}

Unfortunately, one often does not know the weights $w_k$ a priori, which is also hard to set manually, especially for large multi-graphs with many relations.
As such, one may try to infer \textit{both} $\bff$ and $\bfw$ by solving e.g., the following optimization problem:

\vspace{-0.2in}
\beq
\label{nonconvex}
\begin{split}
\min_{\bff,\bfw}\;\; & \|\bff-\bfy\|_2^2 + \lambda \sum_k \bff' w_k \bLk \bff\\
& s.t. \;\; w_k \geq 0, \; \sum_k w_k=1
\end{split}
\eeq
\vspace{-0.1in}

The above objective, however, is non-convex in both $\bff$ and $\bfw$. To get around this issue, several previous approaches have proposed alternating optimization schemes for similar objective functions \cite{journals/tnn/KatoKS09,conf/sdm/Wan15}.

In this work, inspired by the TSS approach \cite{tsuda2005fast}, we introduce a scheme that infers $\bff$ and $\bfw$ \textit{together} under a \textit{convex} setup.
The graph weights we infer (i.e., $w_k$'s) capture the {\em impact} that each graph has on the solution (i.e., on $\bff$).
Building on this interpretation, we 
devise a learning procedure that estimates $\bff$ which is \textit{robust} to intrusive graphs. 
In the experiments, we show the superiority of our method to TSS \cite{tsuda2005fast},
its robust extension \cite{journals/tnn/KatoKS09}, and others such as using equal weights (See Eq. \eqref{generic}) or weights estimated solely based on labeled data \cite{Mostafavi08} (See Section \ref{sec:eval}).

\subsection{Objective Formulation$\;\;$}
\label{sec:objective}

We start by reorganizing the objective in Eq. \eqref{generic} 
as
\beq
\label{main0}
\begin{split}
\min_{\bff} \;\; & (\bff-\bfy)' (\bff-\bfy)  \\
& s.t. \;\; \bff' \bLk \bff \leq c \;, \forall k = 1, \ldots, m 
\end{split}
\eeq
The above aims to find a solution, where error of smoothness on each graph is bounded by a positive constant $c$ (as $\bff' \bLk \bff$ is non-negative). Considering some graphs are less informative than others, we 
can allow for some slack, and rewrite
\beq
\label{main}
\begin{split}
\min_{\bff,\bxi} \;\; & (\bff-\bfy)' (\bff-\bfy) + c_0 \sum\limits_{k=1}^{m}\xi_k \\
& s.t. \;\; \bff' \bLk \bff \leq c + \xi_k, \; \xi_k \geq 0 \;, \forall k = 1, \ldots, m 
\end{split}
\eeq
By introducing the Lagrange multipliers ($w_k$'s, $\beta_k$'s), we get:
\beq
\label{lagr}
\begin{split}
\min_{\bff,\bxi} \;  \max_{\bfw,\bbeta} & \;\; (\bff-\bfy)' (\bff-\bfy)  + c_0 \sum\limits_{k=1}^{m}\xi_k  \\
& + \sum\limits_{k=1}^{m} w_k (\bff' \bLk \bff -c -\xi_k)  - \sum_k \beta_k \xi_k \\
& s.t. \;\; w_k \geq 0, \; \beta_k \geq 0   \;, \forall k = 1, \ldots, m 
\end{split}
\eeq
Setting derivative with respect to $\xi_k$ to zero we get:
\beq
c_0 - w_k - \beta_k = 0 \nonumber
\eeq
As such, $c_0 \geq w_k \geq 0$ as $\beta_k \geq 0$.
Substituting $c_0 - w_k = \beta_k$ to (\ref{lagr}), we get:
\beq
\label{lagrs}
\begin{split}
 \max_{\bfw} \; \min_{\bff} & \;\; (\bff-\bfy)' (\bff-\bfy)
 + \sum\limits_{k=1}^{m} w_k (\bff' \bLk \bff - c)  \\
& s.t. \;\; c_0 \geq w_k \geq 0  \;, \forall k = 1, \ldots, m 
\end{split}
\eeq
Note that in the above we can swap $\max$ and $\min$ in the optimization \eqref{lagr} due to the min-max theorem \cite{Sion1958}. Setting the derivative with respect to $\bff$ to zero we have:
\beq
\label{fvalue}
\bff - \bfy + \sum_k w_k \bLk \bff = 0 \Rightarrow \bff = (\bI+\sum_k w_k \bLk)^{-1} \bfy
\eeq
Using \eqref{fvalue} we can rewrite \eqref{lagrs} as
\vspace{-0.25in}
\beq
\label{lagrsre}
\begin{split}
 \max_{\bfw} \min_{\bff}\;\; & \bfy \bfy' + \bff' (\bff - 2\bfy) + \bff' (\sum_k w_k \bLk) \bff - \sum_k w_k c\\
& = -2\bff'\bfy + \bfy'\bfy + \bff'(\bI + \sum_k w_k \bLk) \bff - c \sum_k w_k \\
& = -2\bff'\bfy + \bfy'\bfy + \bff'\bfy - c \sum_k w_k \\
& = \bfy'\bfy - \bff'\bfy - c \sum_k w_k \equiv  \\
\max_{\bfw}\;\; &  \bfy'\bfy  - \bfy' (\bI+\sum_k w_k \bLk)^{-1} \bfy - c \sum_k w_k \nonumber
\end{split}
\eeq
\vspace{-0.1in}

Therefore, the dual problem becomes: 
\beq
\label{dual}
\begin{split}
\min_\bfw & \;\; \bfy' (\bI+\sum_k w_k \bLk)^{-1} \bfy + c \|\bfw\|_1 \\
& s.t. \;\; c_0 \geq w_k \geq 0  \;, \forall k = 1, \ldots, m
\end{split}
\eeq
The dual program  \eqref{dual} is convex and can be solved (e.g., using the projected gradient descent method) to infer the graph weights $\bfw$.
One can then plug in those weights directly into Eq. \eqref{fvalue} to estimate $\bff$.
However, this procedure as followed in \cite{tsuda2005fast}, yields inferior results in the presence of irrelevant and noisy graphs. This has to do with what the inferred weights capture.
We discuss the interpretation of $w_k$'s in Section \ref{GraphWeightsInterpreted}, which motivates our devised algorithm \method~in Section \ref{sec:algo}.
Before, we provide a modification of our formulation in the face of class bias (i.e., imbalanced class distribution).

\hide{
We discuss :

\subsubsection{Choice of graph Laplacian}

\textit{Claim: } If increasing the weight of edges by a constant factor, the symmetric normalized Laplacian Matrix L is the same.

\textit{Proof: } Assume the graph weights matrix before change is $W$, the one after change is $W'$.

We know the Laplacian Matrix 
$$
L = D^{-\frac{1}{2}}(D - W)D^{-\frac{1}{2}}
$$
where $D$ is a diagonal matrix, in which $d_{ii} = \sum_j W_{ij}$.

After we change the weights matrix by factor $\alpha$, i.e. $W' = \alpha W$, we have $D' = \alpha D$.

Therefore, 
\begin{align}
\begin{split}
L' {}& = D'^{-\frac{1}{2}}(D' - W')D'^{-\frac{1}{2}}  \\
    &= (\alpha D)^{-\frac{1}{2}}\alpha(D - W) (\alpha D)^{-\frac{1}{2}} \\
    &= D^{-\frac{1}{2}}(D - W)D^{-\frac{1}{2}}\\
    &= L
\end{split}
\end{align}
}

\vspace{0.05in}
\paragraph{Class bias.$\;$}

When there is bias in the class distribution (without loss of generality, let `+1' and `-1' respectively depict minority and majority classes), we assign large penalties to the misclassification of `+1' instances. Otherwise, the minority class is largely ignored when optimizing the loss function.
This is often the case in anomaly detection where the anomalous class is relatively much smaller (e.g., fraudsters are much fewer than benign users).

In semi-supervised learning, only part of the nodes are labeled for training, and the rest are unlabeled (depicted with `0'). For each node type (`+1',`0', `-1'), we assign a different penalty coefficient, $c_+$, $c_u$, $c_-$ respectively. Let $\bC$ be a $n\times n$ diagonal matrix, called the {\em class penalty matrix}, where $\bC(i,i) = c_+$ if $y_i = 1$, $c_-$ if $y_i = -1$, and $c_u$ if $y_i = 0$.  
As such, the criterion in \eqref{main}  can be reformulated as:

\vspace{-0.15in}
$$
\min_{\bff,\bxi} (\bff-\bfy)' \bC (\bff-\bfy) + c_0 \sum\limits_{k=1}^{m}\xi_k 
$$
Following the same derivation as before, we obtain the 
 dual problem 
 
 \vspace{-0.25in}
\beq
\label{main2}
\begin{split}
 \min_\bfw \;\; & \bfy' \bC ( \bC - \sum_k w_k \bLk)^{-1} \bC \bfy + c \|\bfw\|_1 \\
  & s.t.  \;\; c_0\geq w_k\geq 0 \;, \forall k = 1, \ldots, m
  \end{split}
\eeq  
  and the solution
\beq
\label{fstar}
\bff^* = (\bC+\sum_k w_k \bLk)^{-1} \bC \bfy \;.
\eeq



\vspace{-0.2in}
\subsection{Graph Weights Interpreted$\;\;$}
\label{GraphWeightsInterpreted}

Next we provide a detailed discussion on
the interpretation of the inferred weights by \eqref{main2}.
In a nutshell, we show that in the presence of intrusive graphs, the weights do {\em not} reflect the 
relative {\em informativeness} of individual graphs---but rather the relative {\em impact} of each graph on the solution.


Ideally, we want to infer a weight $w_k$ for each graph $G_k$ proportional to its informativeness for the task, where the weights for intrusive graphs are zero.
For example, in Figure \ref{fig:toy} we illustrate a toy multi-graph with five views. The ideal weights would  look like $w_1 > w_2 > w_3 > w_4=w_5=0$.
As we show in the following, however, the estimated weights should be interpreted carefully when we have intrusive graphs in the data.

\vspace{0.05in}
\paragraph{Graphs $G_k$ with larger $\bff'\bLk\bff$ tend to get larger $w_k$:}
We have the dual problem $d(\bfw)$ in \eqref{dual} when learning the weights.
We know from basic calculus that

\vspace{-0.2in}
\beq
\frac{\partial}{\partial x} Y^{-1} = - Y^{-1}(\frac{\partial}{\partial x}Y)Y^{-1} \;.
\eeq
\vspace{-0.2in}

Thus we derive the derivative of $d(\bfw)$ w.r.t $w_k$ as

\vspace{-0.1in}
\beq
\label{gradient}
\frac{\partial d ({\bfw})}{\partial w_k} = - \bfy'(\bI+\sum\limits_{i=1}^{m}w_i \bLi)^{-1}\bLk(\bI+\sum\limits_{i=1}^{m}w_i \bLi)^{-1} \bfy + c
\eeq
\vspace{-0.1in}

Since $\bff= (\bI+\sum_{i=1}^{m}w_i \bLi)^{-1}\bfy$, we obtain

\vspace{-0.2in}
\beq
\label{incDeriv}
\frac{\partial d({\bfw})}{\partial w_k} = - \bff' \bLk \bff +c
\eeq
\vspace{-0.1in}

Based on (\ref{incDeriv}), we make the following inferences:

\vspace{-0.075in}
\bit
\setlength{\itemsep}{-0.75\itemsep}
\item The value of $\bff' \bLk \bff$ (i.e., the smoothness penalty) determines the changing rate in each dimension $w_k$ of the gradient descent, i.e., 

\vspace{-0.1in}
$$w_k^{t+1} \leftarrow w_k^t - \frac{\partial d({\bfw})}{\partial w_k} = w_k^t + \bff' \bLk \bff - c \;.$$
\vspace{-0.1in}

\item For graphs with larger $\bff' \bLk \bff$, the rate is larger. As a result, those graphs tend to receive larger weights.

\eit

Further intuition as to why graphs with large $\bff' \bLk \bff$  tend to have large weights is as follows. 

{\em SVM intuition:} Graphs that are important for the solution (correct instances) as well as those that are irrelevant (erroneous instances) all become support vectors, and receive non-zero dual coefficients (i.e., weights).

{\em Constraints intuition:} Graphs that ``obstruct'' the road to the solution more, i.e., cause the constraints ($\bff' \bLk \bff \leq c$) to be violated the more, will get larger $w_k$.

\vspace{0.05in}
\paragraph{Both dense informative and intrusive graphs $G_k$ has large $\bff'\bLk\bff$---and hence large $w_k$:}

In this part, we first show that dense informative graphs have large $\bff' \bLk \bff$. 
Consider a graph with no noisy edges (i.e., no edges between nodes from different classes) but with high edge density among nodes that belong to the same class. 
For such a graph, $\bff'\bLk\bff =  \sum_{i,j\in V} 
\bWk(i,j) (\frac{f_i}{\sqrt{\bD(i,i)}}  - \frac{f_j}{\sqrt{\bD(j,j)}})^2$ can be large due to the numerous non-zero (although likely small) quadratic terms in the sum.

Next, we argue that it is not only the dense informative graphs that have large $\bff' \bLk \bff$, but \textit{also the intrusive graphs}. This is mainly due to the many cross-edges that irrelevant and noisy graphs have between the nodes from different classes, which yield large quadratic terms.

We demonstrate this through the inferred weights on our example multi-graph in Figure \ref{fig:toy}. Notice that while the highly informative $G_1$ and $G_2$ receive large weight, the noisy graphs $G_4$ and $G_5$ also obtain comparably large weights.

Finally, we show that the graphs with larger weights have higher impact on the final solution.

\vspace{0.05in}
\paragraph{The larger the $w_k$, the higher the impact of $G_k$ on $\bff$:}

When $w_k$'s are fixed in \eqref{lagr}, the term $w_k (\bff'\bLk \bff-c-\xi_k)$ needs to be smaller for larger $w_k$ as we are to minimize w.r.t. $\bfw$. As a result, the graphs that have large weights ``pull'' $\bff$ towards themselves, i.e., force the estimations $f_i$'s to be smooth over their structure. As such, larger weight graphs have larger impact on the solution.
This is also evident from $\bff = (\bI+\sum_k w_k \bLk)^{-1} \bfy$ in \eqref{fvalue}---the larger the $w_k$, the more the $\bLk$ is integrated into and influences the solution $\bff$.

These arguments show that in general the estimated {\em weights are  not indicators of graph quality or informativeness but of impact} that each graph has on the solution.
When intrusive graphs are present in the data, they receive large weights. As a result they inflict large impact on the solution, by ``pull''ing the solution toward fitting to their structure.
As intrusive graphs have structure that is \textit{not} compliant with the class labels, we would obtain a poor solution (note the low initial performance in Figure \ref{fig:toy}).

On the other hand, the inferred weights would be as desired in the \textit{absence} of intrusive graphs. Then, the dense informative graphs get large weights and are the sole claimers of high impact on $\bff$ (cf. Figure \ref{fig:toy}). 
This suggests one needs to ``weed out'' the intrusive graphs to obtain reliable estimates.
We introduce our proposed algorithm for this goal next. 


\hide{
	
	%

	
	The first type are irrelevant graphs to remove. However, the second type consists of both irrelevant graphs and relevant graphs. Therefore, if we remove the largest weight graph in each iteration, the performance might decrease significantly because the removal of relevant but dense graph. 
}

\vspace{-0.05in}
\section{Robust SsC for Multi-Graphs: Algorithm}
\label{sec:algo}


To briefly reiterate Section \ref{GraphWeightsInterpreted}, we find that 
graphs with large weights highly influence the solution.
Moreover, intrusive graphs (if any) are among those with large weights and should be carefully filtered out to obtain a reliable solution.

In this section, we propose a robust algorithm for semi-supervised classification in multi-graphs. 
The pseudo-code is given in Algorithm \ref{alg:SA}. We first describe the steps of the algorithm and then provide details on the parameters and computational complexity.

The goal is to successfully remove the intrusive graphs.
The main idea is to explore the search space through simulated annealing by carefully removing large-weighted graphs one at a time.
We start with introducing a queue of graph-sets, which initially includes the set of all graphs (line 2). We process the graph-sets in the queue one by one until the queue becomes empty (line 3). For each graph-set $\mgs$ that we dequeue (line 4), we compute its cross-validation performance $cvP$ on the labeled data (line 5). In our experiments, we use average-precision (AP; area under the precision-recall curve) as our performance metric.  This metric is more meaningful than accuracy, especially in the face of class bias (one can achieve high accuracy by always predicting the majority class).

\begin{algorithm}[t]
\caption{\method~ (proposed algorithm for robust semi-supervised classification for multi-graphs)} \label{alg:SA}
\begin{algorithmic}[1]
\REQUIRE Multi-graph $\mg = \{G_1, \ldots, G_m\}$, labeled nodes $L$, 
 initial temperature $t$, class penalty matrix $\bC$
\ENSURE Label estimations $\bff$
\STATE Construct $\bfy$ as described in Notation (\S\ref{sec:proposed})
\STATE $best\bff \leftarrow \emptyset$, $bestP = 0$, $m=|\mg|$, $Q \leftarrow \mg$ 
\WHILE{$Q$ is not empty}
	\STATE $\mgs \leftarrow dequeue(Q)$ 
	\STATE $cvP =$ Compute cross validation performance of $\mgs$
	\IF{$rand(0,1) \leq \exp({\frac{cvP - bestP}{t^{m-|\mgs|+1}}})$} 
		\STATE $\bfw_{\mgs} \leftarrow$ Solve \eqref{main2} using $\mgs$ and input $\bC$
		\STATE $\bff_{\mgs} \leftarrow$ Compute solution using \eqref{fstar} and $\bfw_{\mgs}$
		\STATE Cluster the weights: $(W_s,W_l) \leftarrow 2$-$means(\bfw_{\mgs})$ 
		\FOR{{\bf each} $G_k\in \mgs$ for which $w_k \in W_l$ }
			\STATE $v \leftarrow hash(\mgs\backslash G_k)$
			\LINEIF{$v$ is null}{$Q \leftarrow Q\; \cup \; \mgs\backslash G_k$}
		\ENDFOR
	\LINEIF{$cvP > bestP$}{$best\bff \leftarrow \bff_{\mgs}$, $bestP = cvP$}	
	\ENDIF
\ENDWHILE
\RETURN $best\bff$
\end{algorithmic}
\end{algorithm}
\setlength{\textfloatsep}{0.2in}

%

We record the best AP as $bestP$
during the course of our search (line 14).
With probability $\exp({\frac{cvP - bestP}{t^{m-|\mgs|+1}}})$, we ``process'' the graph-set in hand (lines 7-13, which we will describe shortly), otherwise we discard it.
In line 6, $t\leq 1$ is the temperature parameter of simulated annealing and $(m-|\mgs|)$ denotes the number of removed graphs from the original set.
If the graph-set $\mgs$ in hand yields a $cvP$ that is larger than $bestP$, we always process the set further, since when $(cvP - bestP)\geq 0$, $\exp({\frac{cvP - bestP}{t^{m-|\mgs|+1}}})\geq 1$. 
On the other hand, if $\mgs$ yields inferior performance, we still process it with some probability that is proportional to the size of the graph-set. That is, the probability of processing a set decreases as they have more graphs removed from the original set.
The probability is also inversely proportional to the performance distance $(cvP - bestP)$. The larger the gap, the higher the chance that $\mgs$ will be discarded. 

Next we describe the steps to ``process'' a graph-set $\mgs$. We first solve the optimization problem \eqref{main2} using $\mgs$ for the graph weights $\bfw_\mgs$ and compute the solution based on $\bfw_\mgs$ (lines 7-8). Next we cluster the weights into two groups, those with small weights $W_s$ and those with large weights $W_l$ (line 9). We know, through the analysis in \S\ref{GraphWeightsInterpreted}, that intrusive graphs are {\em among} the large-weighted graphs. The issue is we do not know in advance which ones, as dense informative ones are likely to also belong to this group. As such, we create from $\mgs$ candidate graph-sets that contain all but each large-weighted graph and add those to the queue. Note that we maintain a hash table of the candidate graph-sets (line 11), so that we avoid re-considering the same sets that might be generated through different removal paths. At the end, we return the solution $best\bff$ with  the $bestP$.

\vspace{-0.075in}
\subsection{Parameters}


Our algorithm expects two main parameters; the initial simulated annealing temperature $t$, and the class penalty matrix $\bC$. We describe how we carefully set these in the following. 
Note that our objective function \eqref{main2} involves two further parameters $c$ and $c_0$. Those are hyper-parameters, which we choose through cross-validation.



\paragraph{Initial temperature $t$.$\;$}

As we remove more and more graphs from the input multi-graph, the probability of further considering a set  with inferior performance should decrease. That is when $d=(cvP - bestP)<0$, $p=\exp({\frac{d}{t^{m-|\mgs|+1}}})$ should decrease as $r = (m-|\mgs|+1)$ increases. As such, we need $t\leq 1$.

Assume that we have an expected range $[m_l,m_u]$ for the number of intrusive graphs in the data where $m_l$ and $m_u$ respectively denote the minimum and maximum number.
We would then want the probability $p = \exp(\frac{d}{t^r})$ to approach zero as $r$ gets closer to $m_u$ even for a considerably small $d$. That is, as $r\rightarrow m_u$ and $0 > d \geq d_{thresh}$ for small $d_{thresh}$, we want $p_{thresh}>p>0$ for small $p_{thresh}$.

Since $t = (\frac{d}{\ln p})^{\frac{1}{r}}$, the range for $t$ satisfying the above constraints can be given as
\beq
t \in [(\frac{d_{thresh}}{\ln p_{thresh}})^{\frac{1}{m_u}},(\frac{d_{thresh}}{\ln p_{thresh}})^{\frac{1}{m_l}}]
\eeq
Empirically, we let $d_{thresh} = -0.1$ and $p_{thresh} = 0.01$. Thus if we expect $m_l = 5$ and $m_u = 10$, then the initial temperature is chosen randomly from $t \in [0.465, 0.682]$.

%

\paragraph{Class penalty matrix $\bC$.$\;$}

As described in \ref{sec:objective}, we can normalize biased class distribution by assigning larger penalty to minority-class mis-classification. 
Recall that $c_+$, $c_u$, $c_-$ denote penalty coefficients for classes `+1', `0' (unlabeled), and `-1', respectively. We set these parameters as
$
c_+ = 1 + const * sign(1-2f) * \max(f,1-f)
$,
$
c_u = 1
$, and 
$
c_- = 1 + const * sign(2f-1)* \max(f,1-f)
$,
where $const$ is a constant set to 0.7, and $f$ is the fraction of class `+1' instances in the labeled set. For example, if $f=0.2$, then $c_+ = 1.56, c_2 = 1, c_- = 0.44$.
One can also treat $const$ as a hyper-parameter and select it through cross-validation under a performance metric of interest.

\subsection{Computational Analysis}

While our algorithm as presented uses a queue to process candidate graph-sets sequentially, it is amenable to parallelization. In fact, our publicly available source code employs a parallel implementation, see {{\small \url{www.cs.stonybrook.edu/~juyye/#code}}}.

In particular, we can think of the explored graph-sets to form a search tree structure, rooted at the original multi-graph containing all the graphs. Each removal of a graph from a given graph-set produces a new leaf node in the tree attached to its superset. As such, a series of successive removals forms a path from the root to a leaf.
Each of these search paths can be executed independently in parallel ($bestP$ and the hash table are shared across processes). 

As such, the computational complexity is proportional to the $depth$ of the search tree. Since we control the temperature parameter $t$ such that the exploration probability approaches zero as we remove an expected maximum number $m_u$ of graphs, we have $depth \leq m_u$.

At each node of the tree, we solve the optimization problem \eqref{main2} using projected gradient descent, where the main computation involves computing the gradient (See \eqref{gradient} in \S\ref{GraphWeightsInterpreted}).  The gradient involves the term $(\bI+\sum_{i=1}^{m}w_i \bLi)^{-1}$, i.e., the inverse of a $(n\times n)$ matrix which is $O(n^3)$ if done naively.
The same is true for the solution $\bff$ which requires a similar inverse operation (See \eqref{fvalue} or \eqref{fstar}). Luckily, however, we do not need to compute the inverse explicitly, because it always appears as the vector $\mathbf{x} = (\bI+\sum_{i=1}^{m}w_i \bLi)^{-1}\bfy$.
We can obtain $\mathbf{x}$ as a solution of sparse linear systems, where the computational cost of the derivative is nearly linear in the number of non-zero entries of $\sum_{i=1}^{m}w_i \bLi$, i.e., proportional to the number of edges of the multi-graph \cite{DBLP:conf/stoc/SpielmanT04}.

Computing the dual objective then takes $O(s|\mathcal{E}|)$, where $s$ is the number of steps of the gradient descent algorithm. 
All in all, the total time complexity of a parallel implementation becomes $O(s|\mathcal{E}|m_u)$, which is tractable for most sparse real-world multi-graphs where $m_u$ and $s$ are often small.

\paragraph{Pruning computation.$\;$}

In the following, we describe several strategies we use that help us keep the size of the search tree tractable, even for a serial implementation. 

\vspace{-0.1in}
\bit
\setlength{\itemsep}{-0.75\itemsep}
\item \textit{Branching factor}: During our search, we only consider graphs with large weights as candidates to remove. This way the branch-out factor of the search tree becomes smaller than considering all possible removals.
\item \textit{Simulated annealing}: Our search terminates when there is significant decrease in the cross-validation performance. As such, some search paths take shorter than others, i.e., the search tree is not fully-balanced.
\item \textit{Tree depth}: 
As discussed earlier, we control the depth of the search tree through $m_u$, expected upper bound on the number of intrusive graphs. As often $m_u \ll m$, the search tree is terminated significantly earlier than e.g., a brute-force search.
\eit




\section{Evaluation}
\label{sec:eval}

We evaluated our \method~on real-world datasets, and compared it to a list of baselines and state-of-the-art methods. We also ``noise-tested'' all methods under varying level, intensity, and models of noise.


\vspace{-0.1in}
\subsection{Experiment Setup}

\paragraph{Datasets.$\;$}
The real-world multi-graphs used in our work are publicly available, as listed in Table \ref{tab:realData}. 
\textit{RealityMining} \cite{eagle2009inferring} contains 4 different relations between two classes of students (in business school and in computer science): phone call, Bluetooth scans, SMS, and friendship. \textit{Protein} \cite{tsuda2005fast} consists of Yeast proteins, associated through 5 relations, where those with function ``transport facilitation'' constitute the positive class, and others the negative. \textit{Gene1} and \textit{Gene2} contain different sets of Yeast genes, each associated through 15 different genomic sources. Those are obtained from \cite{Mostafavi08} which we refer to for details. Also see Appendix \ref{sec:viz} for further description and visualization of our multi-graphs.

\paragraph{Baselines.$\;$}
We compare \method~against three state-of-the-art: {\sc TSS} \cite{tsuda2005fast}, {\sc RobustLP} \cite{journals/tnn/KatoKS09}, and {\sc GeneMania} \cite{Mostafavi08}. We also introduce two simple baselines, {\sc Eql-Wght} that assigns equal weight to all graphs and {\sc Perf-Wght} that assigns weights proportional to the cross-validation accuracy of individual graphs on labeled nodes.

\paragraph{Noise-testing.$\;$}
To test the robustness of the methods, we tested them in the presence of injected intrusive graphs with varying level, model, and intensity of noise.

\vspace{-0.1in}
\bit
\setlength{\itemsep}{-0.75\itemsep}
\item {\em Number of intrusive graphs}: We tested the effect of increasing noise level  by injecting 2, 4 and 6 intrusive graphs at a time.
\item {\em Noisy graph models}:  We adopted 3 strategies to generate intrusive graphs, (1) Erdos-Renyi random graphs (ER), (2) edge-rewired original graphs (RW), and what we call (3) adversarial graphs (AV) (where most edges are cross-edges between the different classes).
\item \textit{Noise intensity} (low/high): Intensity reflects injected graph density (5\%/50\%) for ER, ratio of within-class edges rewired to be cross-edges (60\%/80\%) for RW, and ratio of cross-edges (60\%/80\%) for the AV model.
\eit
\vspace{-0.1in}

Overall, there are 3 different number of injected graphs, 3 models, and 2 noise intensities. As such, we ``noise-tested'' the methods under 18 (3*3*2) different settings.

\vspace{-0.1in}
\subsection{Evaluation Results}

To perform semi-supervised classification, we label 5\% of the nodes in {\em Protein},
{\em Gene1}, and {\em Gene2} and 30\% in RM as it is a smaller dataset.
We randomly sample the labeled set 10 times, and report the mean Average Precision (area under precision-recall curve) in Table \ref{tab:allresults} (notice in Table \ref{tab:realData} that datasets are class imbalanced, hence accuracy is not a good measure to report). Also see Appendix \ref{sec:ap} for the precision-recall plots.

We see that \method~outperforms all baselines across all the datasets. Its superior performance is especially evident in the presence of noise, when the performance of others degrade dramatically (See Appendix \ref{sec:rm} for similar results on RM injected with 2 and 4 intrusive graphs).

\begin{table}[t]
\small{
	\begin{center}
		\caption{\textit{Real-world multi-relational graphs used in work.}\label{tab:realData}}
		\vspace{-0.1in}
		\begin{tabular}{|c|c|r|r|r|}
			\hline  \textbf{Dataset} &  \textbf{\#Graphs $m$} & \textbf{ \#Nodes $n$} & \textbf{\#Pos.} & \textbf{\#Neg.}\\ 
			\hline  \textit{RealityMining} \cite{eagle2009inferring} &  4 &  78 & 27 & 51 \\ 
			\hline  \textit{Protein} \cite{tsuda2005fast} &  5 &  3,588 & 306 & 3,282 \\ 
			\hline   \textit{Gene1} \cite{Mostafavi08} & 15 & 1,724 & 185 & 1,539\\ 
			\hline   \textit{Gene2} \cite{Mostafavi08} &  15 & 3,146 & 214 &  2,932\\ 
			\hline 
		\end{tabular} 
	\end{center}
	}
	\vspace{-0.2in}
\end{table}


\begin{table*}[t!]
	\begin{center}
		\caption{\textit{Performance of all methods on real-world multi-graphs. RM also injected with 6 intrusive graphs with various settings. Values depict mean and standard deviation Average Precision (AP) (over 10 runs with different labeled set).}
	}
	\vspace{-0.1in}
		\label{tab:allresults}
		{\fontsize{8}{9.5}
		\selectfont
		\begin{tabular}{p{0.38in}|p{0.3in}|p{0.57in}|p{0.3in}|>{\bfseries} p{0.54in}|p{0.65in}|p{0.65in}|p{0.54in}|p{0.62in}|p{0.65in}}
			\hline   Dataset 					& \#Graphs 	  & NoiseModel   & Intensity & {\sc Proposed}  & \perf & \eql &  \tss &  \rob &  \gene \\ 
			\hline   \multirow{7}{*}{RM} & 4				&   ------------ 			& ------	   & 0.970$\pm$0.004 &  0.944$\pm$0.004 &  0.939$\pm$0.018 &  \textbf{0.970$\pm$0.004} &  0.947$\pm$0.010 &  0.951$\pm$0.021  \\  
			\cline{2-10}   						&  4+6				&  Adversarial    & Low		   & 0.970$\pm$0.004    &  0.389$\pm$0.049 &  0.277$\pm$0.024 &  0.354$\pm$0.036 &  0.284$\pm$0.026 &  0.217$\pm$0.012 \\ 
			\cline{2-10}  						&  4+6				&  Adversarial    & High		& 0.970$\pm$0.004   &  0.257$\pm$0.026 & 0.223$\pm$0.011   & 0.577$\pm$0.041  &   0.225$\pm$0.010 & 0.197$\pm$0.003   \\
			\cline{2-10}   						&  4+6				&  Rewire 			& Low   	 & 0.930$\pm$0.004   &  0.468$\pm$0.049 & 0.396$\pm$0.033  & 0.597$\pm$0.041 & 0.371$\pm$0.036  & 0.235$\pm$0.011      \\ 
			\cline{2-10}   						&  4+6				&  Rewire 			& High        & 0.907$\pm$0.073   &  0.292$\pm$0.026 &  0.267$\pm$0.020 &  0.571$\pm$0.044 &  0.264$\pm$0.023 &  0.202$\pm$0.004\\ 
			\cline{2-10}   						&  4+6				&  Erdos-Renyi  &  Low		   & 0.970$\pm$0.004  &  0.860$\pm$0.029 &  0.756$\pm$0.050 & 0.494$\pm$0.072  &  0.810$\pm$0.043 &  0.645$\pm$0.077 \\ 
			\cline{2-10}   						&  4+6				&  Erdos-Renyi  &  High		   & 0.970$\pm$0.004  &  0.937$\pm$0.002 &  0.896$\pm$0.036  & 0.621$\pm$0.078  &  0.907$\pm$0.033 &  0.773$\pm$0.085\\ 
			\hline   Protein					    &  5		 		& ------------ 	  	&  ------		   & 0.457$\pm$0.065   & 0.452 $\pm$0.067 & 0.441 $\pm$0.070 &  \textbf{0.457$\pm$0.065} & 0.439$\pm$0.063  & 0.424$\pm$0.058\\ 
			\hline Gene1				 &   15		 		& ------------ 		&  ------		   &\textbf{0.703$\pm$0.056} & 0.658$\pm$0.048  & 0.632$\pm$0.053  & 0.648$\pm$0.059  & 0.628$\pm$0.054 & 0.509$\pm$0.077  \\ 
			\hline Gene2				 &   15		 		& ------------		& ------		   &\textbf{0.838$\pm$0.031}  & 0.830$\pm$0.048 & 0.809$\pm$0.046  & 0.734$\pm$0.080  & 0.460$\pm$0.072  & 0.229$\pm$0.069\\ 
			\hline 
		\end{tabular}}
	\end{center}
	\vspace{-0.35in}
\end{table*}

\begin{figure}[!t]
\centering
\begin{tabular}{cc}
\hspace{-0.1in}\includegraphics[width=0.53\linewidth,height=1.22in]{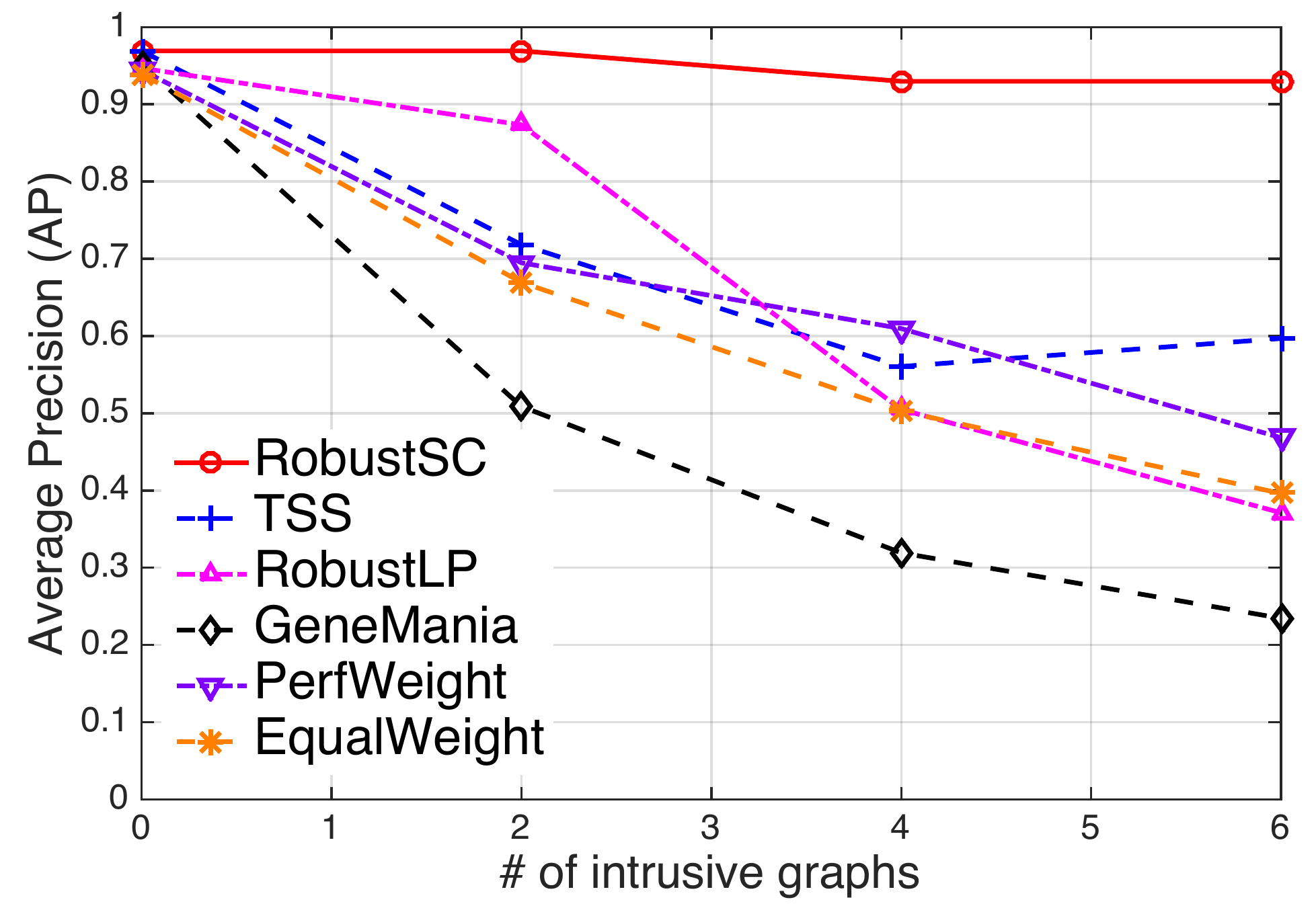} &
\hspace{-0.12in}\includegraphics[width=0.53\linewidth,height=1.24in]{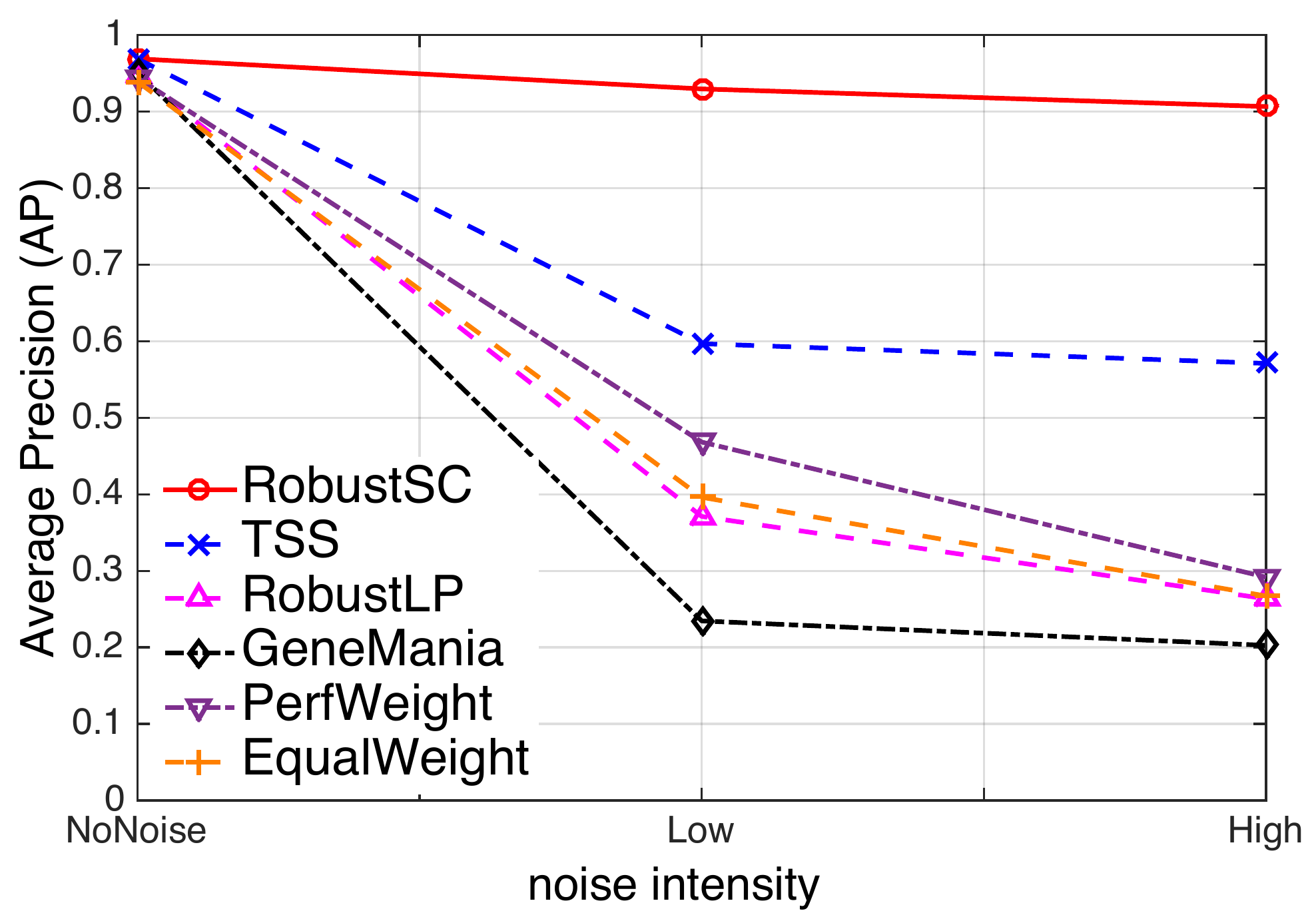}  \\
\end{tabular}
\vspace{-0.2in}
\caption{\textit{Performance by (\textit{left}) increasing number of intrusive graphs, and (\textit{right}) increasing intensity of noise.}
 \label{fig:noise1}}
\vspace{-0.1in}
\end{figure}

\begin{figure}[h]
	\vspace{-0.1in}
\centering
\begin{tabular}{c}
\includegraphics[width=0.75\linewidth,height=1.3in]{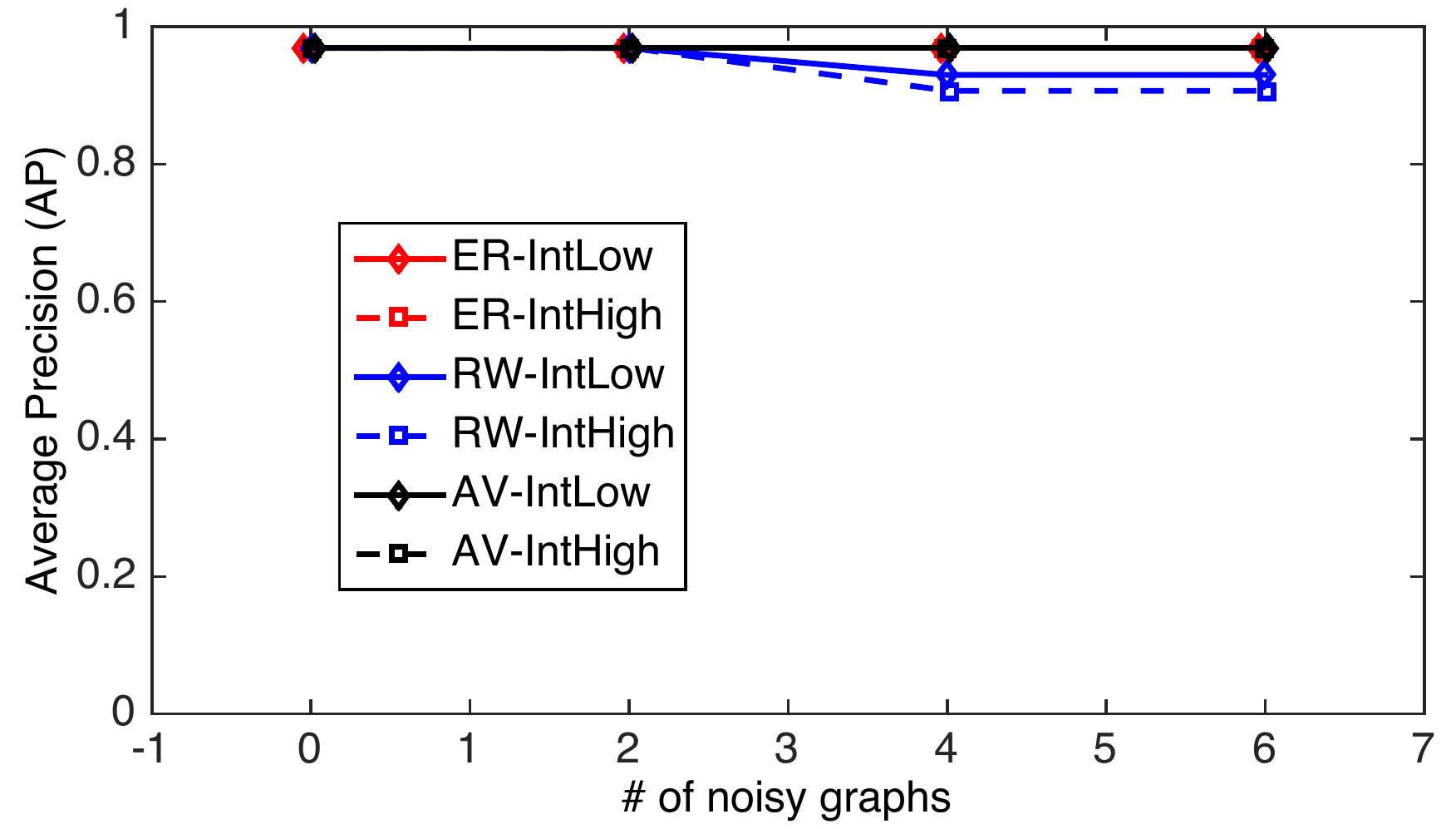} \\
\end{tabular}
\vspace{-0.15in}
\caption{\textit{\method~is robust under varying level (\#graphs), intensity (low/high), models (ER,RW,AV) of noise.}
 \label{fig:noise2}}
\vspace{-0.25in}
\end{figure}

We further investigate the affects of noise using {\em RealityMining} as a running example, as in the absence of noise all methods perform similarly on this multi-graph.
Figure \ref{fig:noise1} (left) shows how the performance of the methods change with increasing number of intrusive graphs (under rewiring and low-intensity). Figure \ref{fig:noise1} (right) shows the same with different noise intensity (under rewiring, 6 intrusive graphs). These clearly show that the competing methods are hindered by noise, while \method's performance remains near-stable. In fact, as Figure \ref{fig:noise2} shows \method~is robust 
under all settings; increasing level and intensity as well as different models of noise.

\begin{figure}[!t]
\centering
\begin{tabular}{c}
\includegraphics[width=0.95\linewidth,height=1.5in]{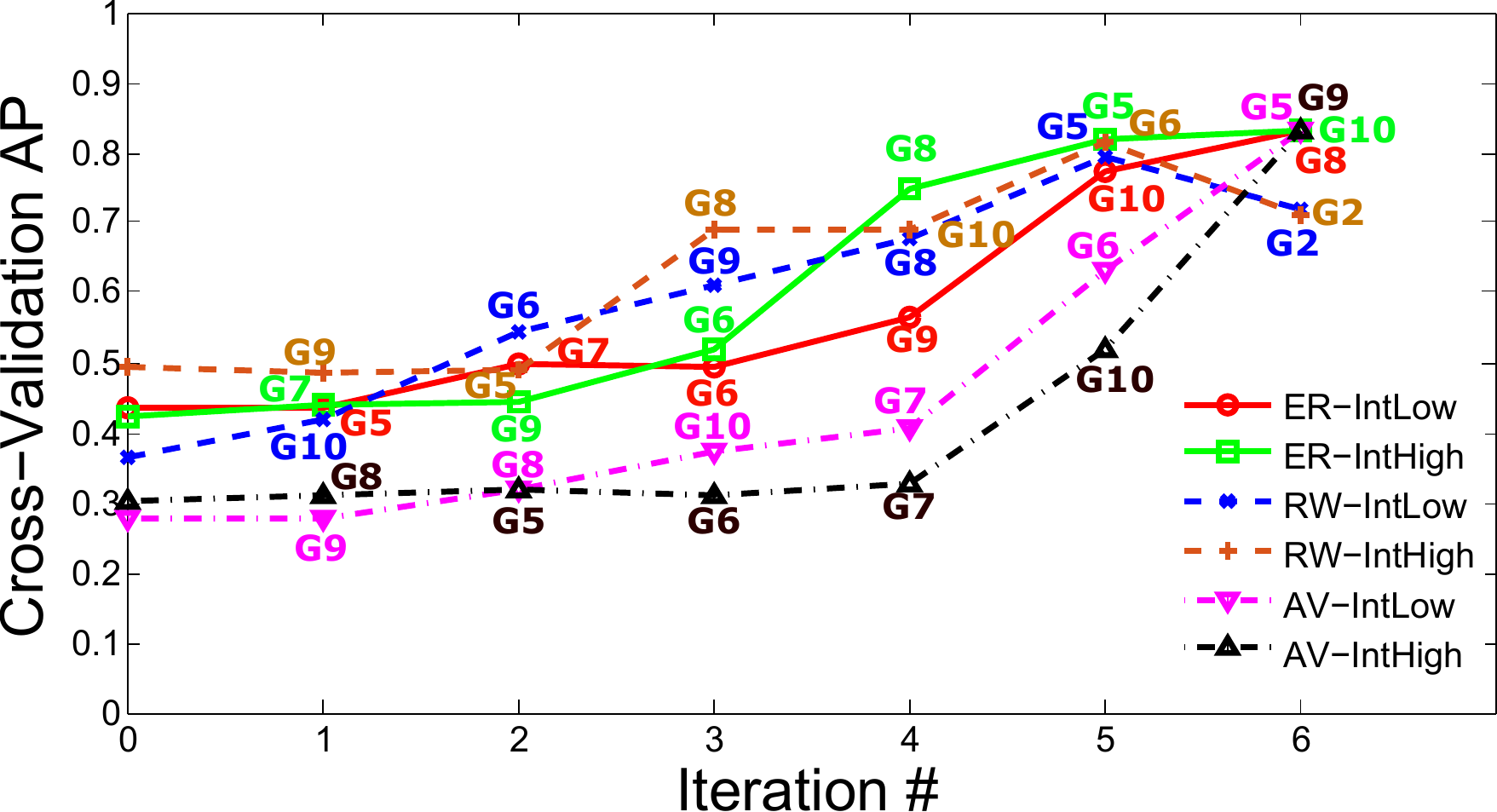} \\
\end{tabular}
\vspace{-0.25in}
\caption{\textit{Removal order of graphs by \method.}
 \label{fig:noise3}}
\vspace{-0.1in}
\end{figure}

Figure \ref{fig:noise3} illustrates the removal order of graphs during the course of \method~(Algorithm \ref{alg:SA}), for each of the six noise settings for \textit{RealityMining} in Table \ref{tab:allresults}. 
The graph ID removed at each iteration is depicted in the figure. We see that simulated-annealing continues to remove graphs as long as cross-validation performance either increases or drops slightly, and stops when it drops significantly. Notice that the proposed method successfully removes most of the injected intrusive graphs $G_5$-$G_{10}$. (See Appendix \ref{sec:order} for a similar figure for the real datasets).

\begin{figure}[h]
	\vspace{-0.1in}
\centering
\begin{tabular}{c}
\includegraphics[width=0.95\linewidth,height=1.2in]{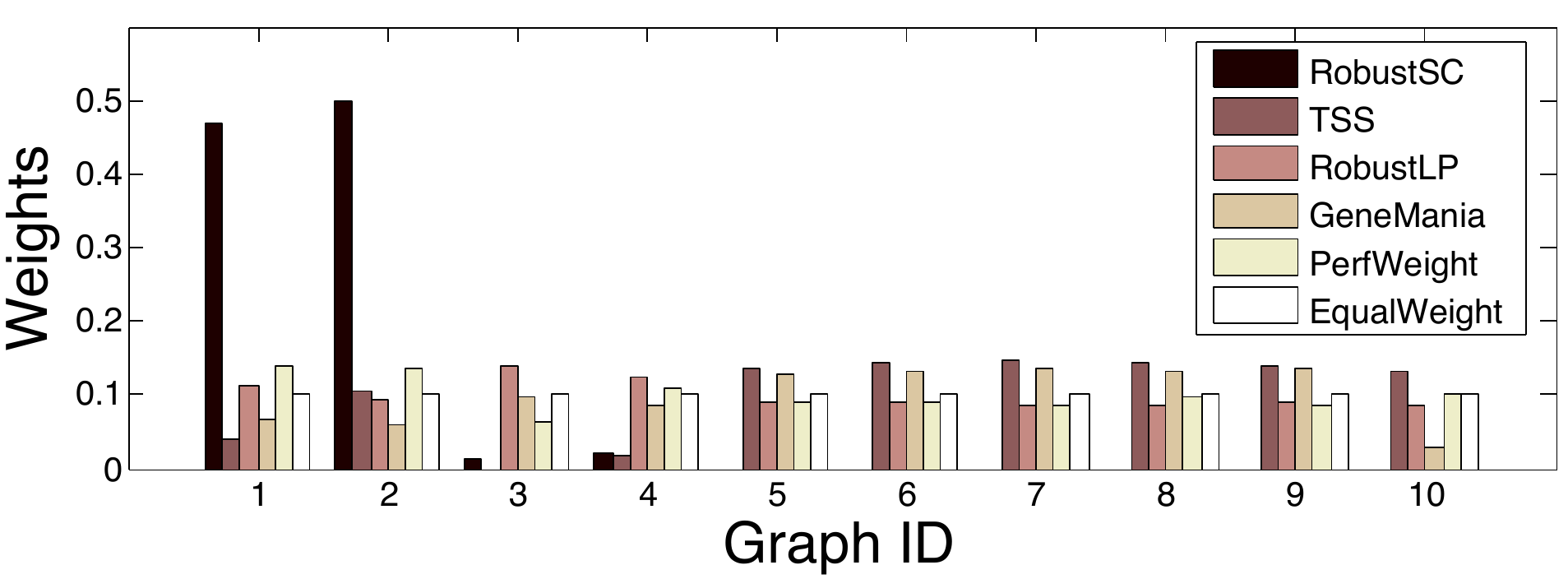} \\
\end{tabular}
\vspace{-0.25in}
\caption{\textit{Inferred graph weights on {\em RealityMining} (+6 injected graphs $G_5$-$G_{10}$ under AV and high-intensity).}
 \label{fig:noise4}}
\vspace{-0.1in}
\end{figure}

We also analyze the inferred weights by each method. Figure \ref{fig:noise4} shows the normalized weights on {\em RealityMining} with 6 injected graphs, under AV with high intensity. (Weight figures for all ten datasets in Table \ref{tab:allresults} are in Appendix \ref{sec:weights}).
Notice that all competing methods give non-zero weights to all the injected graphs $G_5$-$G_{10}$, which hinders their performance. In contrast, \method~puts non-zero weight only on the informative graphs $G_1$-$G_{4}$, particularly large weights on the first two. These are in fact the well-structured and denser informative graphs (See Appendix A.1).

\begin{figure}[h]
\centering
\begin{tabular}{c}
\includegraphics[width=0.95\linewidth,height=1.65in]{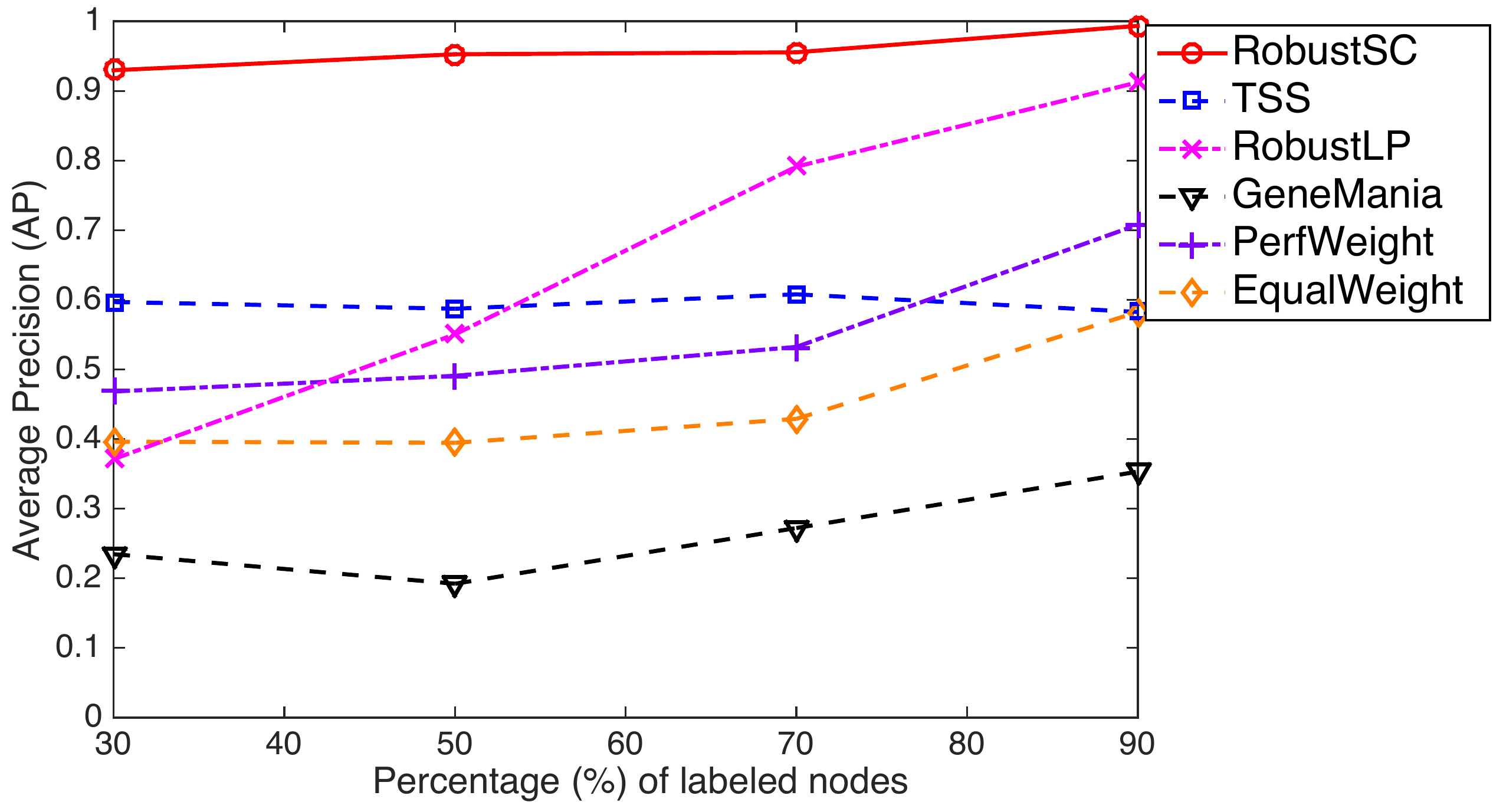} \\
\end{tabular}
\vspace{-0.2in}
\caption{\textit{Performance vs. \% labeled nodes. Competing methods remain hindered by noise despite 90\% data labeled.
		}
 \label{fig:noise5}}
\vspace{-0.25in}
\end{figure}

Finally in Figure \ref{fig:noise5} we show how the performance of the methods change when we increase the labeled set percentage in \textit{RealityMining} from 30\% up to 90\% (6 injected graphs, under rewiring with low intensity; results are avg'ed over 10 runs).
As expected the performance improves for all methods with increasing labeled data.
However, the competing methods cannot achieve improved robustness and reach the same performance level by \method, even when they are provided 90\% data labeled.
As such, robustness is not a function of labeled data consumed.

\section{Related Work}
\label{sec:related}

Semi-supervised classification for graph data aims to classify the unlabeled nodes using a (small) set of available labeled nodes in the graph.
Most methods in the literature (especially for unattributed graphs) are transductive, where the inference is performed over the graph without learning any particular classifier model \cite{macskassy07,conf/icml/BlumC01,zhu2003semi,conf/nips/ZhouBLWS03,conf/colt/BelkinMN04}.
These methods solve a global objective function that enforces smoothness of labels over the graph structure as well as goodness-of-fit to the labeled examples. 
For detailed review of graph-based semi-supervised learning, we refer to \cite{zhu08survey} and \cite{series/synthesis/2014Subramanya}.

While most transductive methods consider a single given graph,  
there also exist methods that aim to predict the node labels by leveraging multiple graphs \cite{LanckrietBCJN04,conf/nips/ArgyriouHP05,tsuda2005fast,Mostafavi08,journals/tnn/KatoKS09,journals/eswa/ShinTS09}. The problem is sometimes referred as multiple kernel learning \cite{conf/icml/BachLJ04}, where data is represented by means of kernel matrices defined by similarities between data pairs.

The SDP/SVM method by Lanckriet \etal~\cite{LanckrietBCJN04} use a weighted sum of kernel matrices, and find the optimal weights within a semi-definite programming (SDP) framework.
Argyriou \etal~\cite{conf/nips/ArgyriouHP05} build on their method to instead combine Laplacian kernel matrices.
Multiple kernel learning has also been used for applications such as protein function prediction \cite{conf/ijcai/YuRDZZ13} and image classification and retrieval \cite{conf/mm/WangJHT10}.
The SDP/SVM  method however is cubic in data size and hence not scalable to large datasets. Moreover it requires valid (i.e., positive definite) kernel matrices as input.

 Mostafavi \etal~\cite{Mostafavi08,MostafaviM10} construct the optimal composite graph by solving a linear regression problem using labeled data, later also used by Luo \etal~\cite{Luo14}. Different from most semi-supervised transductive methods that leverage both labeled and unlabeled data, these use only the labeled data to estimate the graph weights.

The TSS method by Tsuda \etal~\cite{tsuda2005fast} formulates label inference and weight estimation as a convex optimization problem. The weight assignment
mechanism is close to the way that SVM selects support vectors.
Kato \etal~\cite{journals/tnn/KatoKS09} show that the TSS algorithm produces inferior results when some graphs are irrelevant to the task and hence is not robust when noise is present.
They propose a robust method, which alternates
between minimizing their new (nonconvex) objective function with respect to label assignment and with respect to  weight estimation.
Similar alternating optimization of label and weight learning has also been used in \cite{journals/eswa/ShinTS09} and \cite{conf/sdm/Wan15}, where slightly different objectives and weight update rules are used. All of these leverage both labeled and unlabeled data to estimate the graph weights. However, as they introduce nonconvex formulations, they are not guaranteed to find their globally optimum solution.

 Different from the alternating optimization approaches in \cite{journals/tnn/KatoKS09,journals/eswa/ShinTS09,conf/sdm/Wan15}, we estimate the weights directly by solving a single optimization problem. Moreover, we iteratively discard the irrelevant graphs based on a simulated annealing approach guided by the cross-validation performance to prevent them from deteriorating the results.
 Through weeding out the intrusive graphs and learning globally optimal graph weights we produce the most robust results.

\section{Conclusion}
\label{sec:conclude}

In this work we introduce \method, for robust, scalable, and effective \textit{semi-supervised transductive classification for multi-relational graphs}.
The proposed method employs a \textit{convex} formulation that estimates weights for individual graphs, along with a solution that utilizes a weighted combination of them. 
Based on the analysis of weights, we devise a new scheme that iteratively discards intrusive graphs to achieve \textit{robust} performance. Moreover, \method~is \textit{linearly scalable} w.r.t. the size of the combined graph.
Extensive experiments on real-world multi-graphs show that \method~produces competitive results under varying level, intensity, and models of noise where it outperforms
 several baselines and state-of-the-art methods significantly, which are hindered by the presence of noise. 

\hide{

Contributions

\bit

\item We showed that the weights estimated by TSS \cite{} for semi-supervised classification in multi-graphs are hindered by the presence of noisy graphs. In particular both dense high-quality graphs as well as very noisy graphs receive large weights and hence have large impact on the outcome.

\item We introduced a new scheme for filtering out noisy graphs. Our approach iteratively removes one graph at a time that i) receive a large weight, and ii) is estimated to be noisy.

\item We demonstrated the superior performance of our approach on XX synthetic and real world multi-graphs as compared to state of the art methods including TSS \cite{}, Robust \cite{}, GeneMania \cite{}, as well as baselines that ... 

\eit

We provide the open source code of our \method~algorithm for academic use at {{\small \url{www.cs.stonybrook.edu/~juyye/#code}}}.

}

\hide{
\small{
\section*{Acknowledgments}
The authors thank Sivaraman Balakrishnan for their feedback and insightful discussion that helped shape our work.
This material is based upon research supported by the ARO Young Investigator Program under Contract No. W911NF-14-1-0029, NSF CAREER 1452425, IIS 1408287 and IIP1069147, a Facebook Faculty Gift, an R\&D grant from Northrop Grumman Aerospace Systems, and Stony Brook University Office of Vice President for Research. Any
conclusions expressed in this material are of the authors'
and do not necessarily reflect the views, either expressed or implied, of the funding parties.
}}

{\scriptsize{
\bibliographystyle{abbrv}
\bibliography{BIB/refs}
}}

\newpage
\clearpage
\section*{\Large Appendix}
\label{sec:appendix}

\setcounter{section}{0}

\renewcommand\thesection{{\Alph{section}}}

\vspace{0.1in}

\begin{minipage}{1\textwidth}
\section{\large{Description and Visualization of Multi-Graphs}}
\label{sec:viz}

We visualize the adjacency matrices of individual graphs in our real-world multi-grahs as follows. The rows are ordered by class label, such that positive-class instances are listed first, followed by negative-class instances. Edges between the nodes are shown with dots, where black dots depict within-class edges and red dots depict cross-edges. 

\vspace{0.1in}
\subsection{\bf RealityMining}
\textit{RealityMining} \cite{eagle2009inferring} is a 4-view dataset collected through tracking activities in students' cellphones. Two classes of students (business school students as positives and CS students as negatives) are linked via 4 types of activities: phone calls, Bluetooth device scans, text message exchanges and friendship claims.

\end{minipage}

\begin{table}[h!]
	\label{tab:rm}
	\centering
	\begin{tabular}{  m{4cm}  m{4cm} m{4cm}  m{4cm} }
		\begin{minipage}{.3\textwidth}
			\includegraphics[width=35mm, height=35mm]{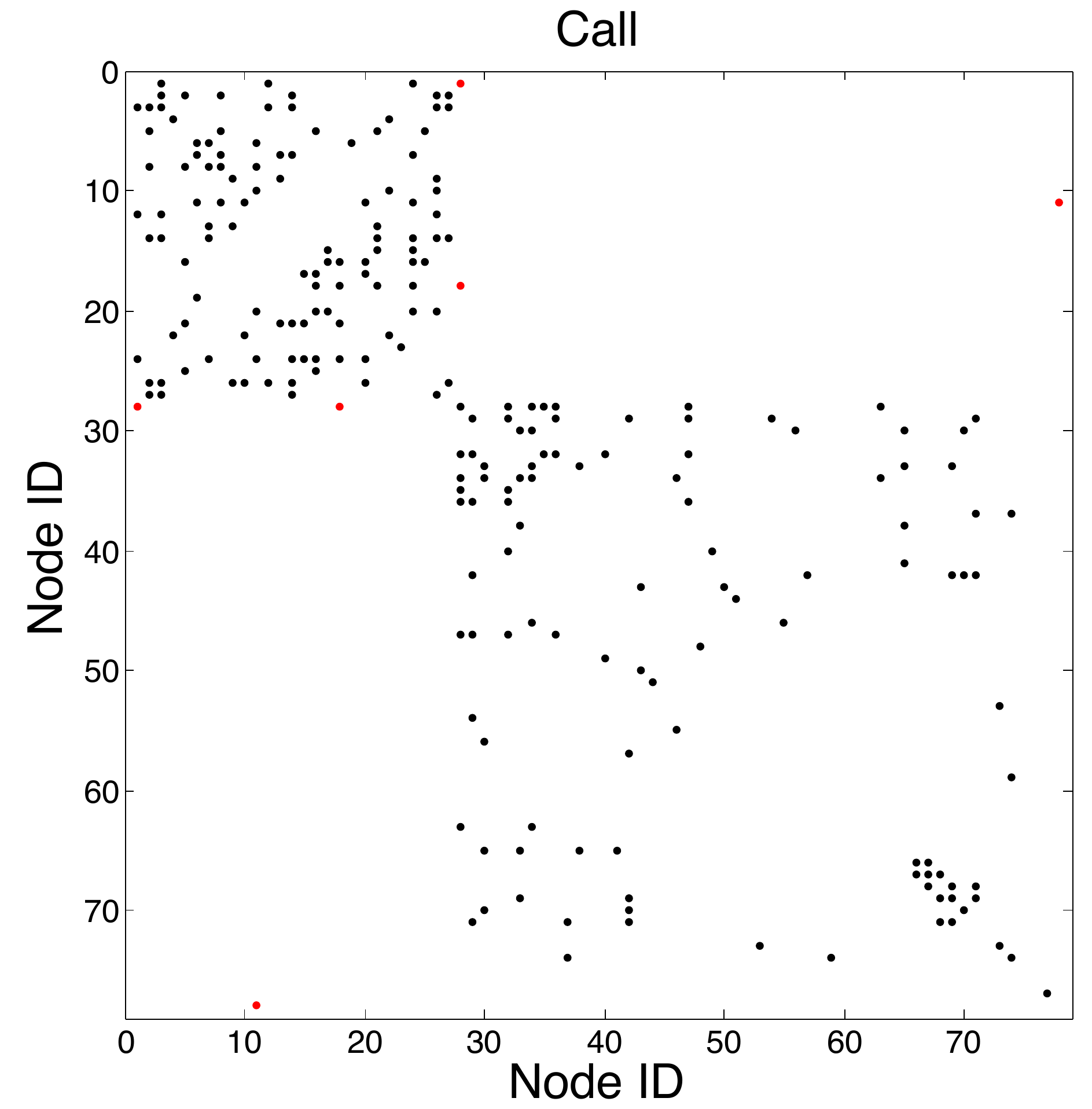}
		\end{minipage}
		&
		\begin{minipage}{.3\textwidth}
			\includegraphics[width=35mm, height=35mm]{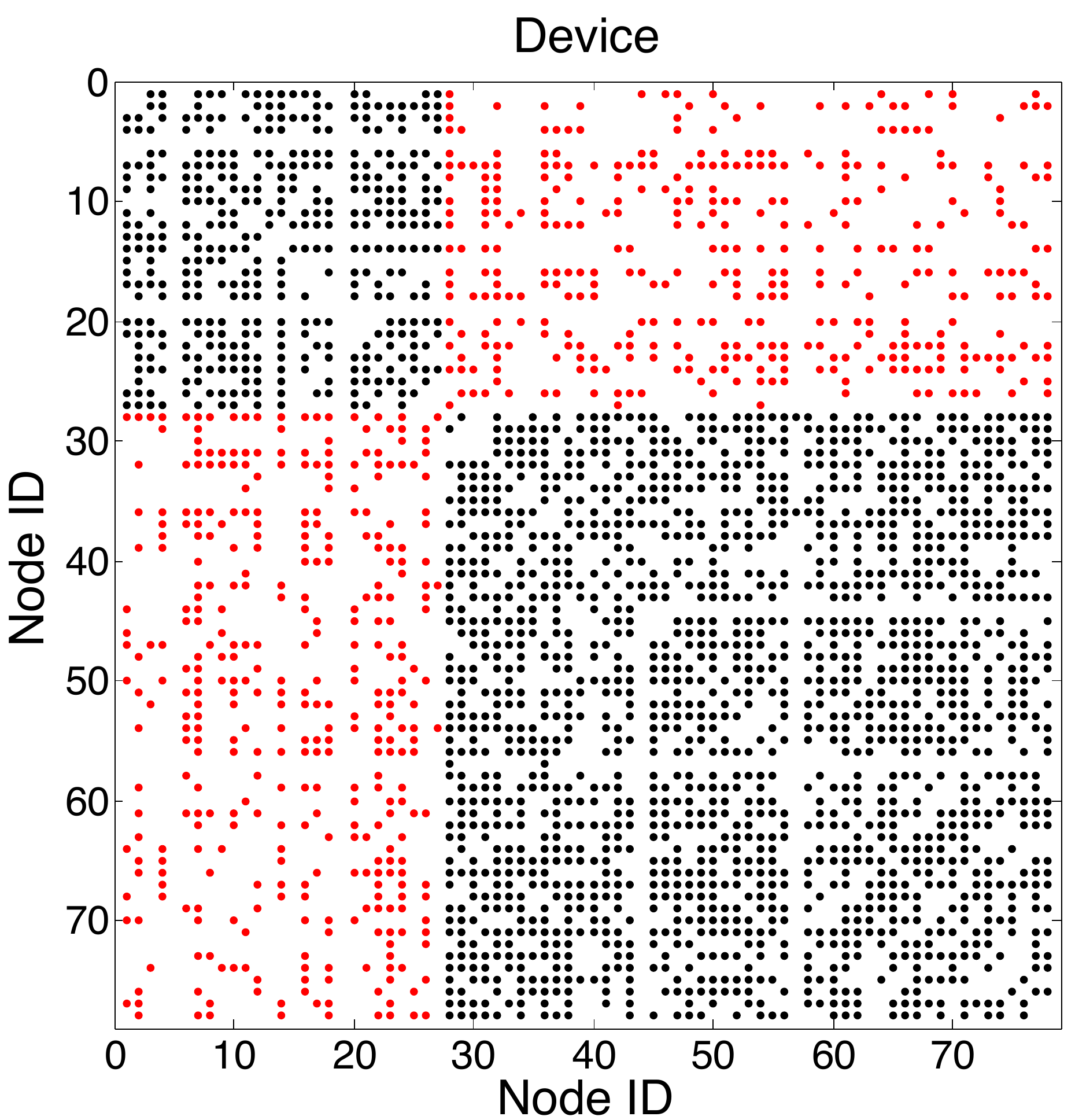}
		\end{minipage}
		&
		\begin{minipage}{.3\textwidth}
			\includegraphics[width=35mm, height=35mm]{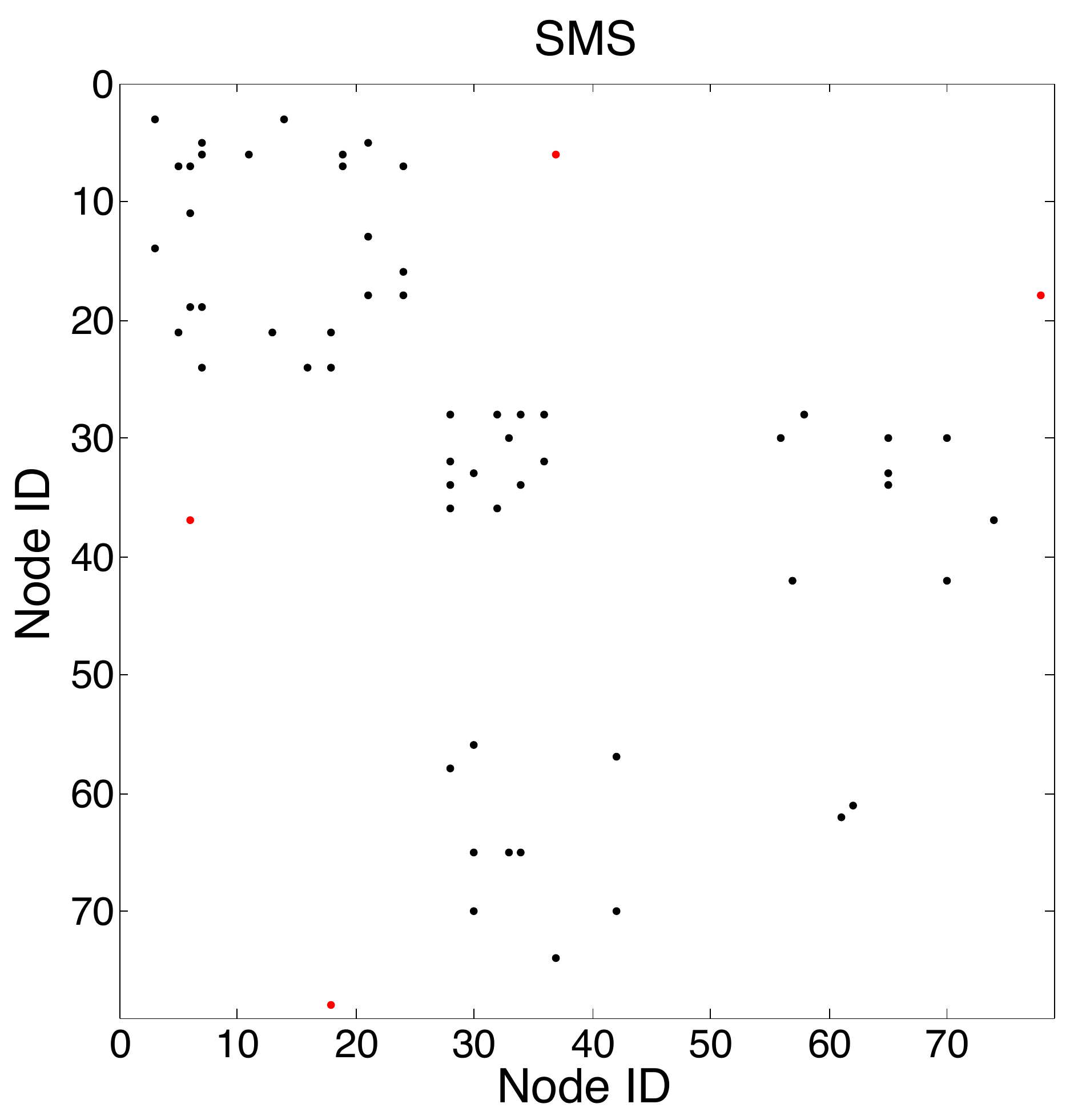}
		\end{minipage}
		&
		\begin{minipage}{.3\textwidth}
			\includegraphics[width=35mm, height=35mm]{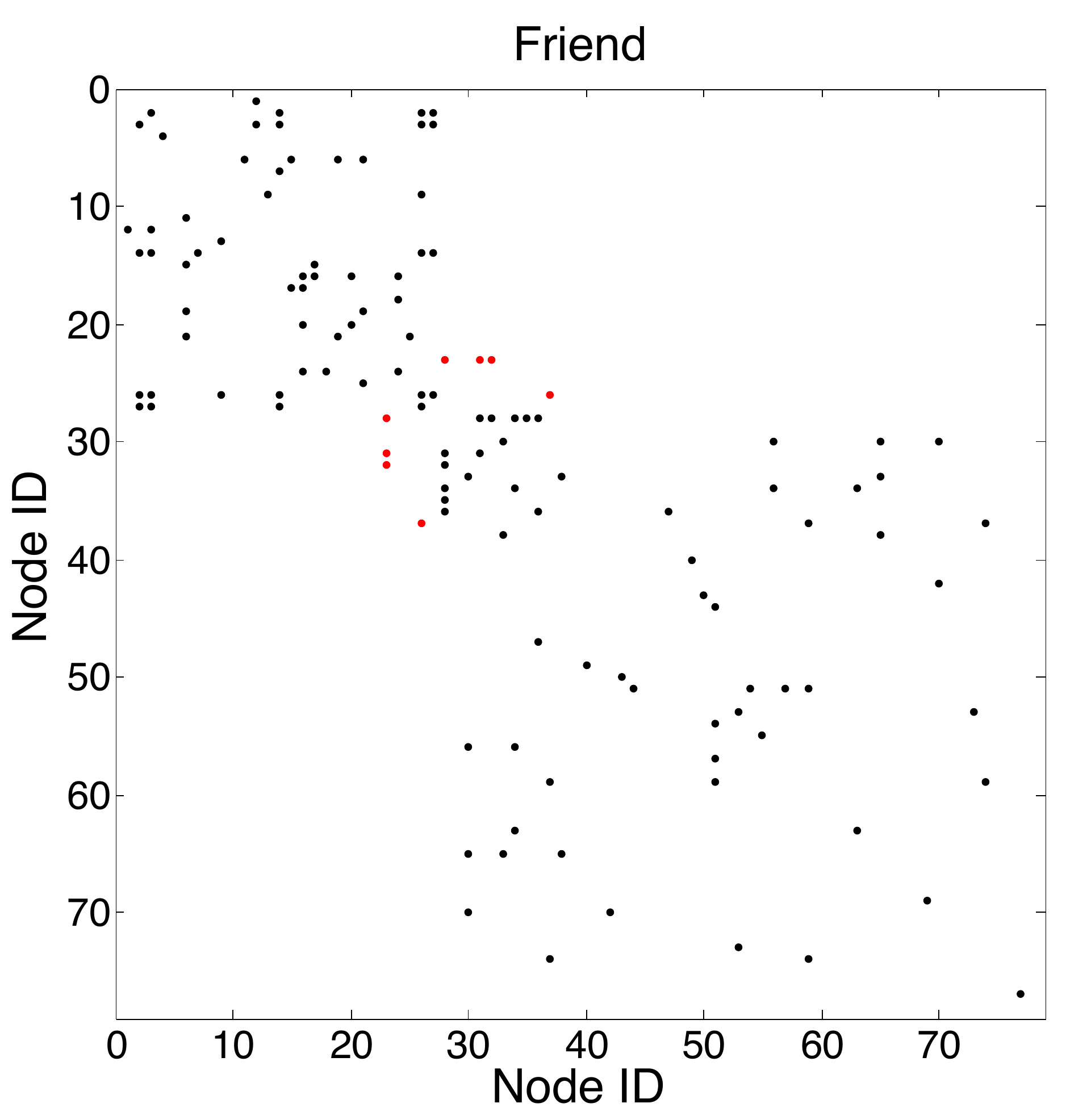}
		\end{minipage}
		\\
	\end{tabular}
\end{table}
\vspace{0.3in}

\begin{minipage}{1\textwidth}
\subsection{Protein} 
\textit{Protein} \cite{tsuda2005fast} consists of 5 relational graphs of yeast proteins. These relations are co-participation in a protein complex, Pfam domain structure similarity, protein-protein interactions, genetic interactions and gene profile similarity. Proteins that exhibit \textit{transport facilitation} function are labeled as positives. The remaining proteins are labeled as negative.
\end{minipage}

\begin{table}[h!]
	\label{tab:protein}
	\centering
	\begin{tabular}{  m{4cm}  m{4cm} m{4cm}  m{4cm} }
		\begin{minipage}{.3\textwidth}
			\includegraphics[width=35mm, height=35mm]{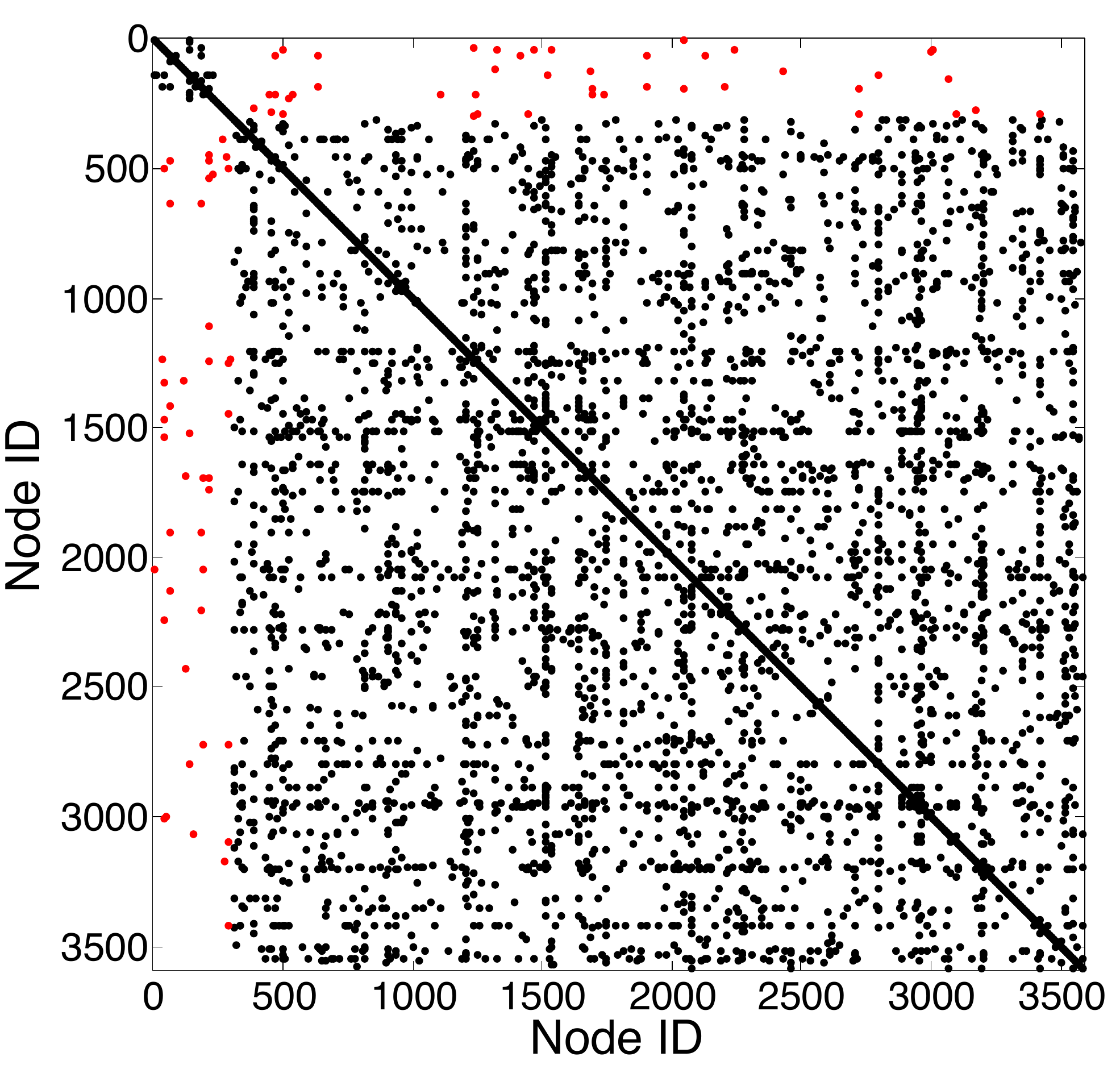}
		\end{minipage}
		&
		\begin{minipage}{.3\textwidth}
			\includegraphics[width=35mm, height=35mm]{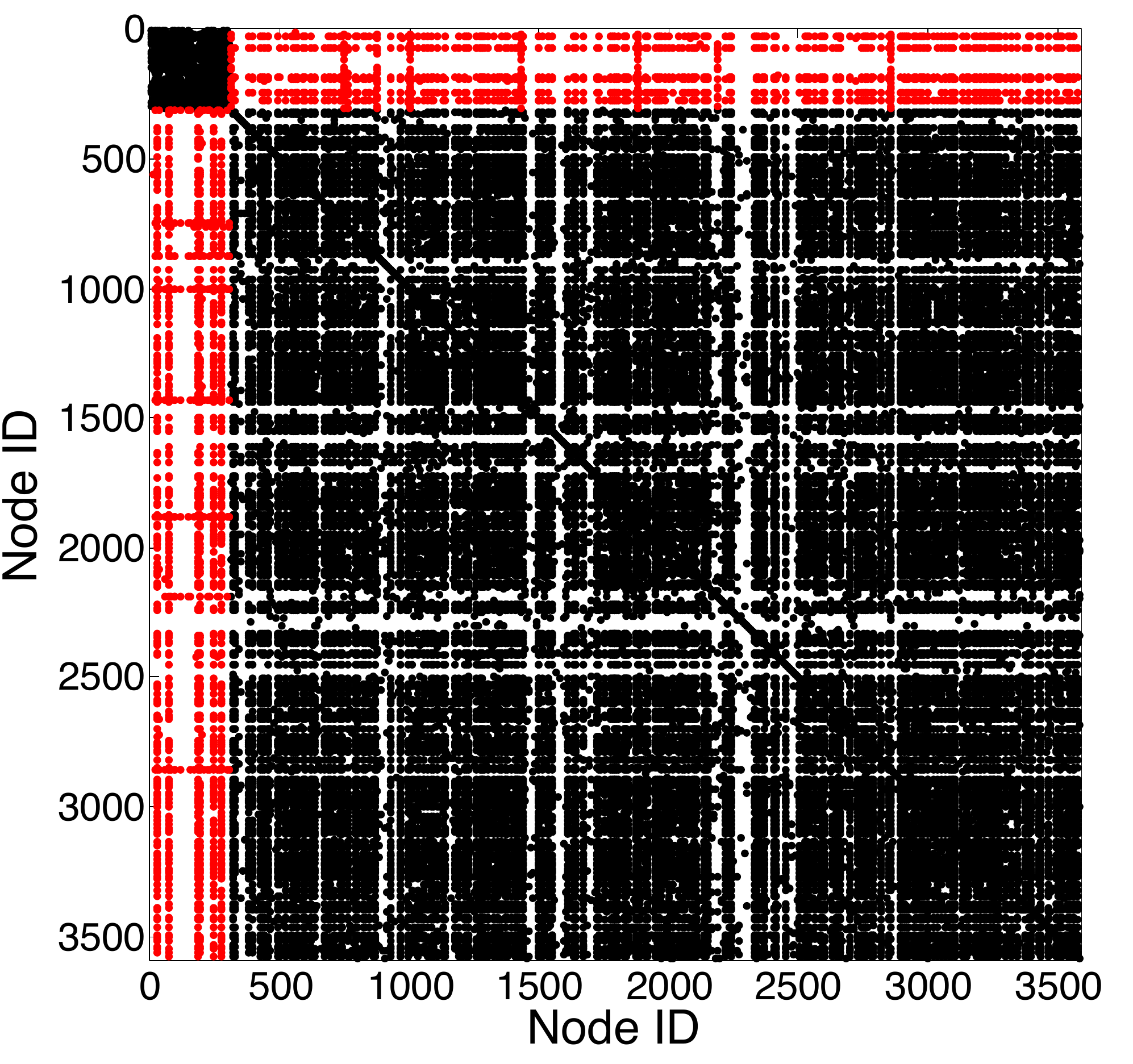}
		\end{minipage}
		&
		\begin{minipage}{.3\textwidth}
			\includegraphics[width=35mm, height=35mm]{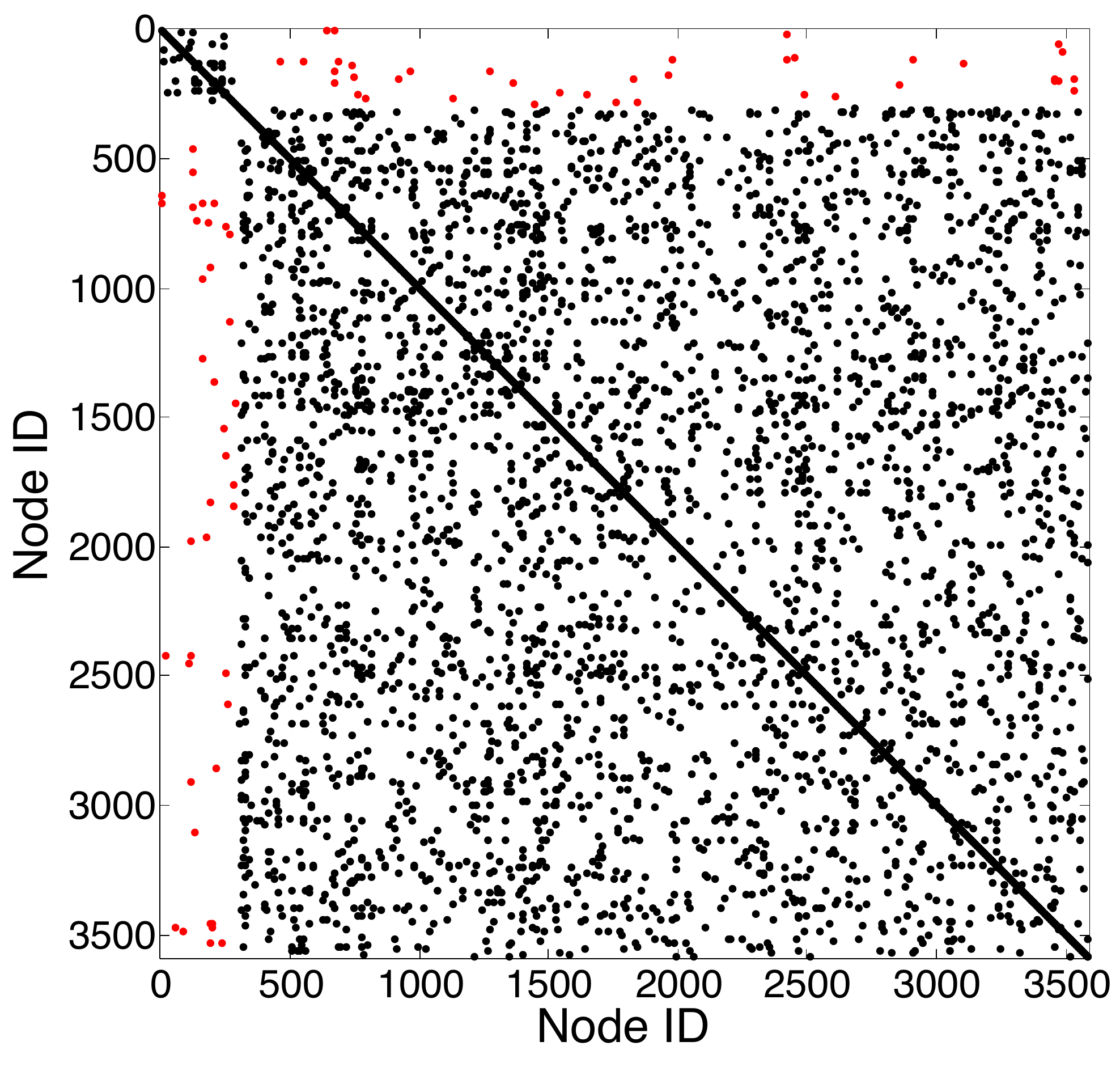}
		\end{minipage}
		&
		\begin{minipage}{.3\textwidth}
			\includegraphics[width=35mm, height=35mm]{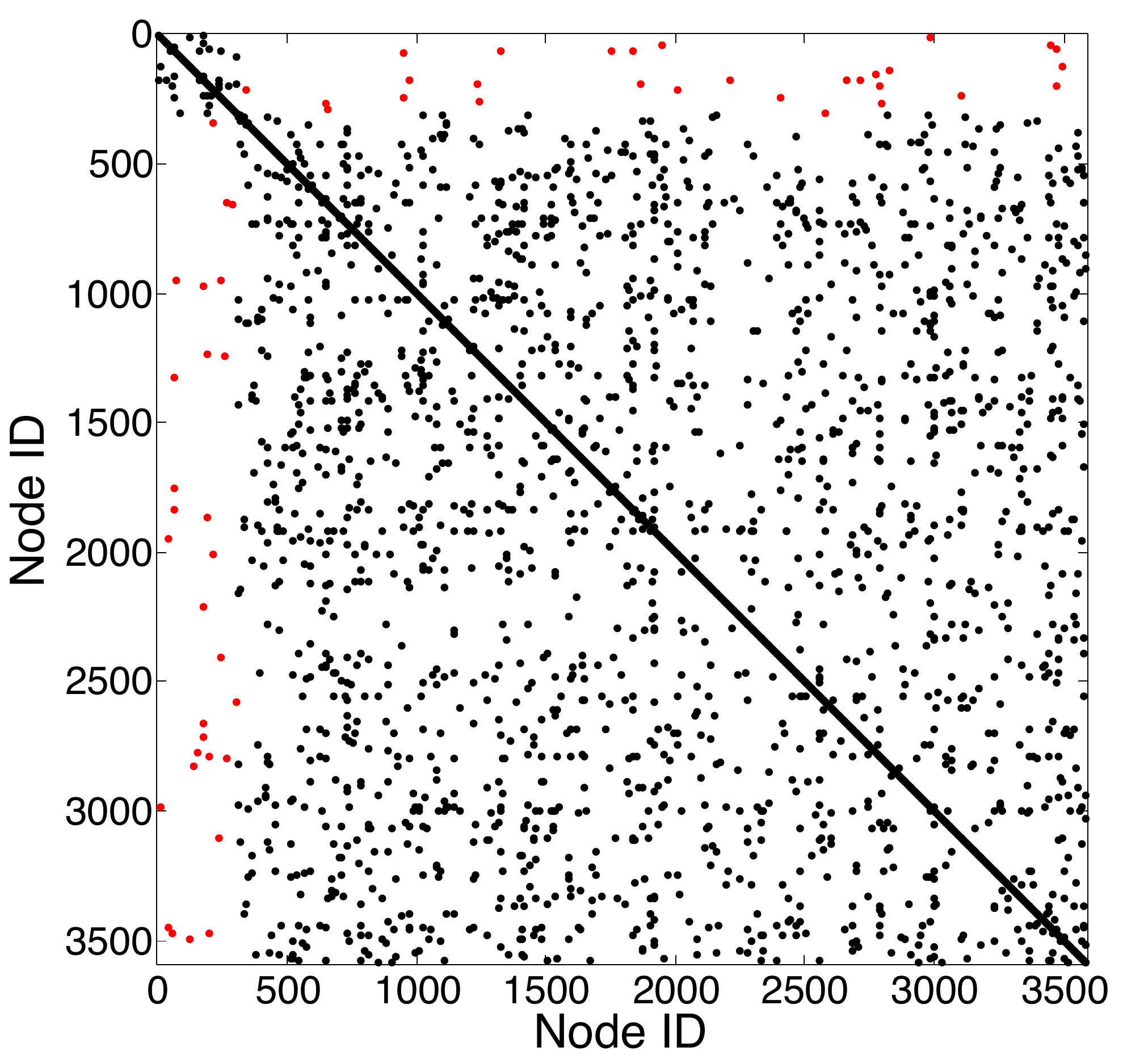}
		\end{minipage}
		\\
		\\
		\begin{minipage}{.3\textwidth}
			\includegraphics[width=35mm, height=35mm]{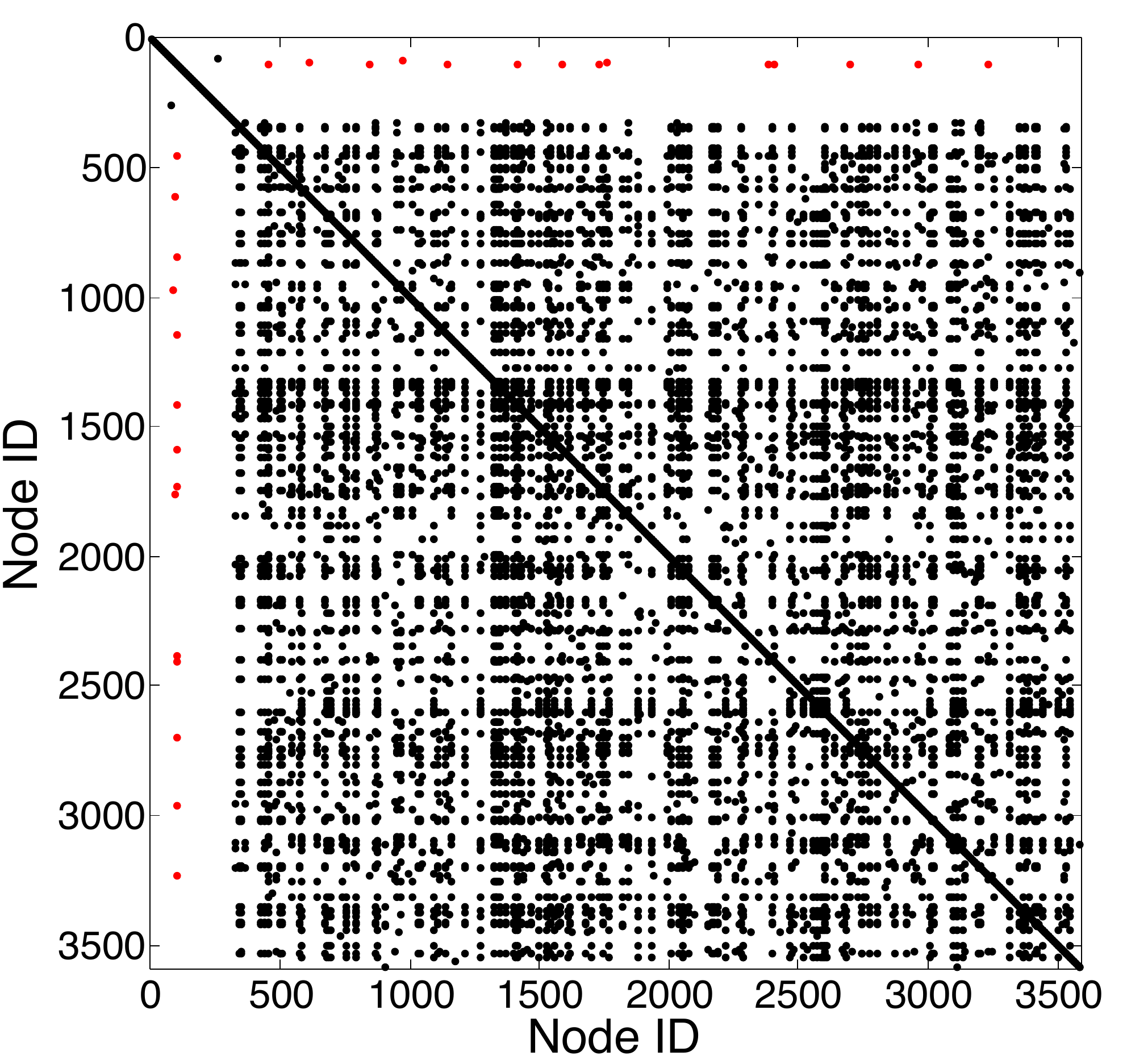}
		\end{minipage}
		\\
	\end{tabular}
\end{table}

\clearpage
\newpage

\begin{minipage}{1\textwidth}
\subsection{\bf Gene1} 
\textit{Gene1} \cite{Mostafavi08} contains 15 relational graphs among yeast genes. All the genes are labeled according to \textit{Gene Ontology (GO)} association file from the Saccharomyces Genome Database. For binary classification setting, we choose the label with maximum number of genes in \textit{Cellular Component (CC)} domain as positive class and consider the remaining genes as negative class. 
\end{minipage}

\begin{table}[h!]
	\label{tab:g1}
	\centering
	\begin{tabular}{  m{4cm}  m{4cm} m{4cm}  m{4cm} }
		\begin{minipage}{.3\textwidth}
			\includegraphics[width=35mm, height=35mm]{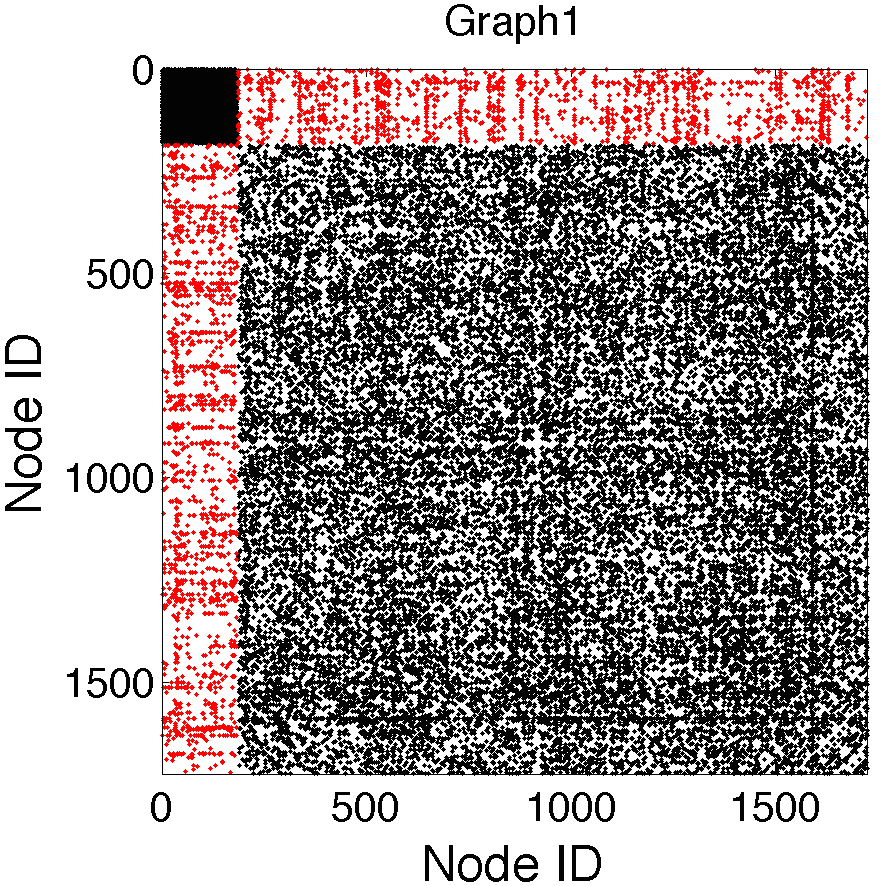}
		\end{minipage}
		&
		\begin{minipage}{.3\textwidth}
			\includegraphics[width=35mm, height=35mm]{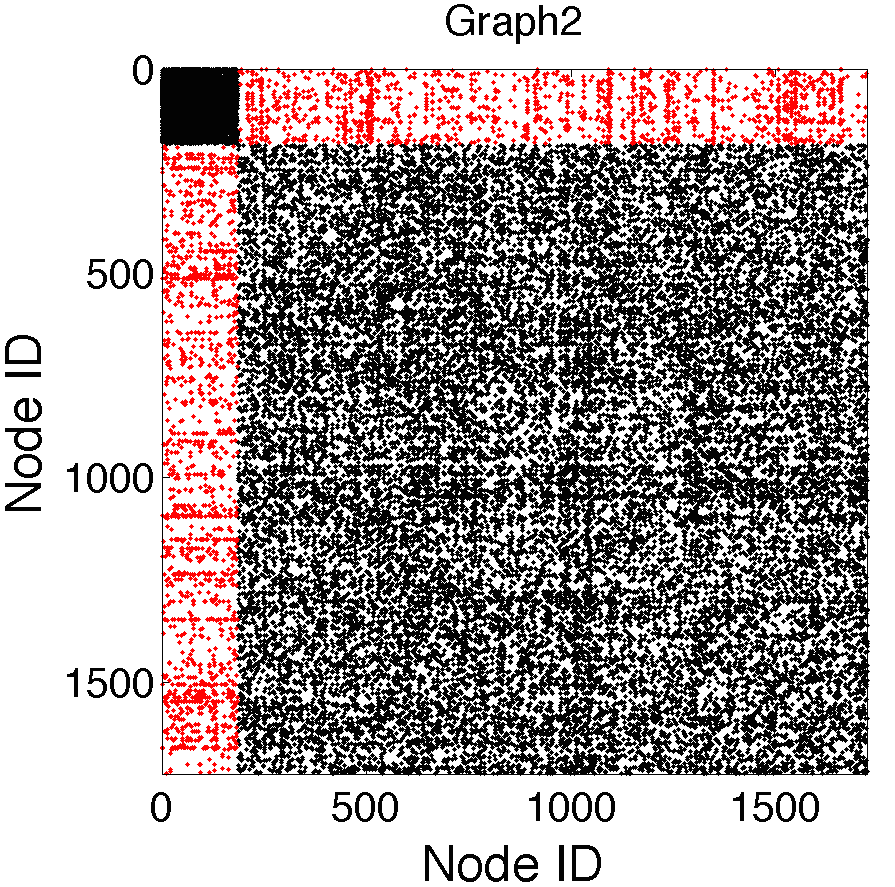}
		\end{minipage}
		&
		\begin{minipage}{.3\textwidth}
			\includegraphics[width=35mm, height=35mm]{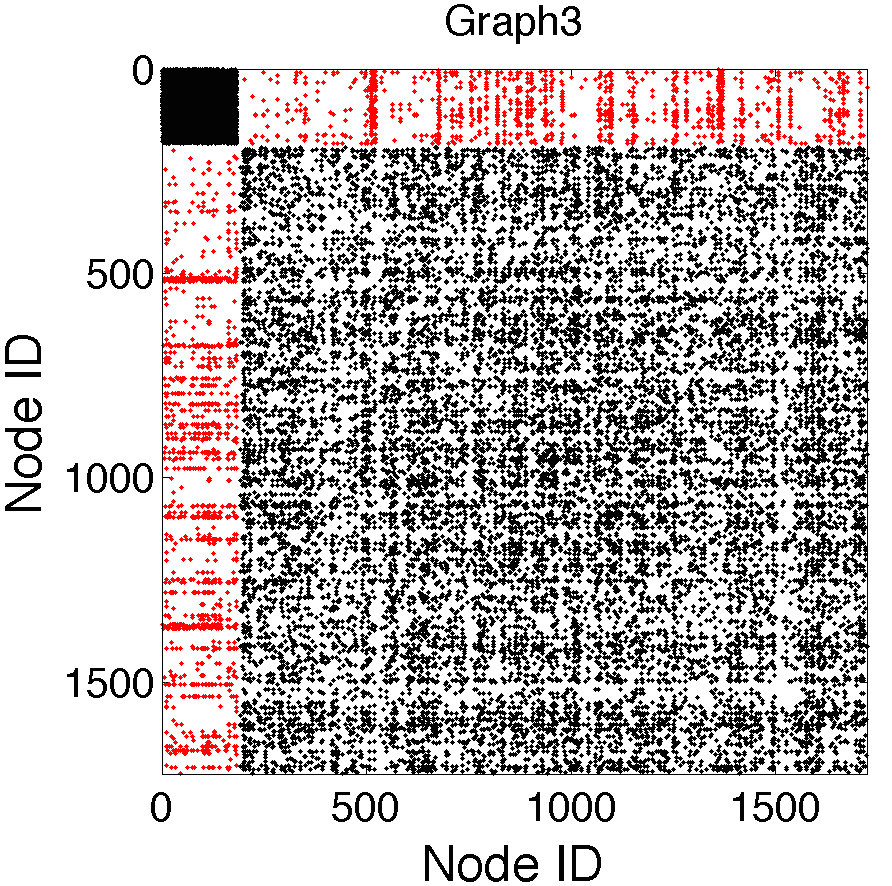}
		\end{minipage}
		&
		\begin{minipage}{.3\textwidth}
			\includegraphics[width=35mm, height=35mm]{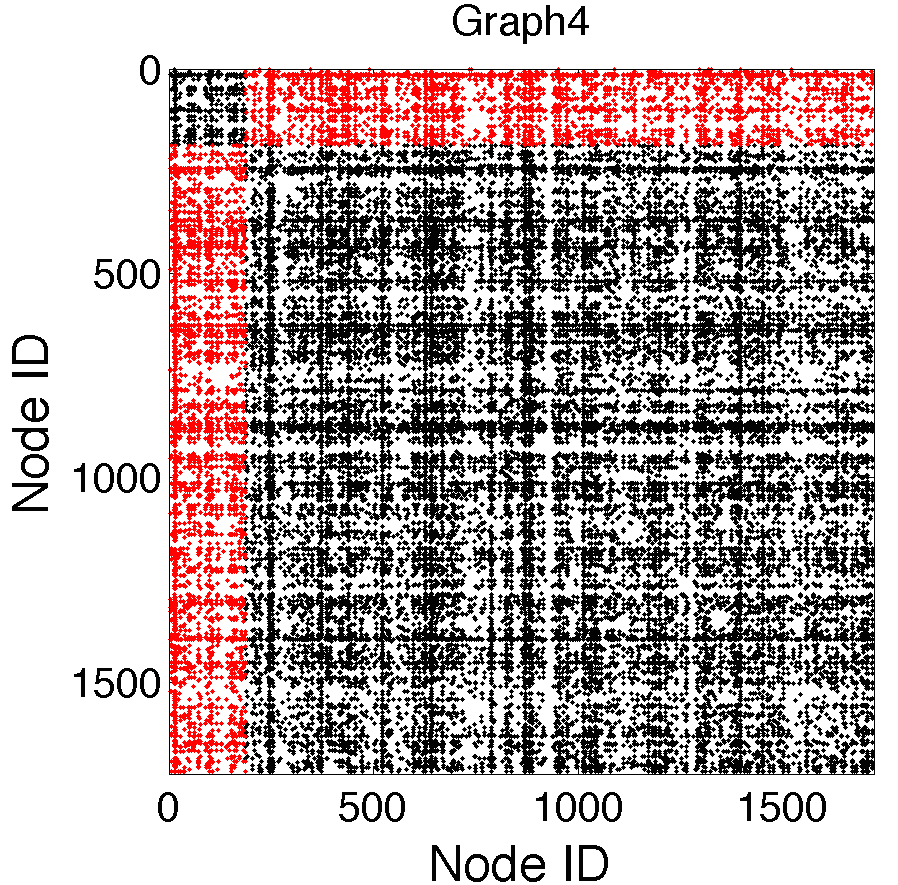}
		\end{minipage}
		\\
		\\
		\begin{minipage}{.3\textwidth}
			\includegraphics[width=35mm, height=35mm]{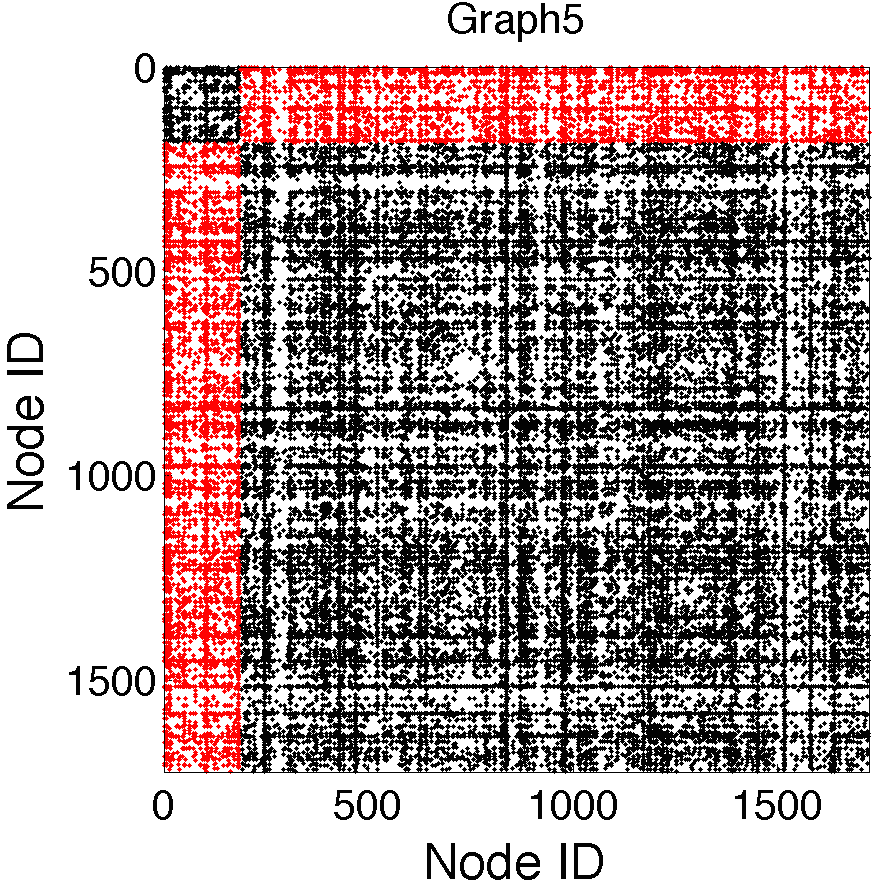}
		\end{minipage}
		&
		\begin{minipage}{.3\textwidth}
			\includegraphics[width=35mm, height=35mm]{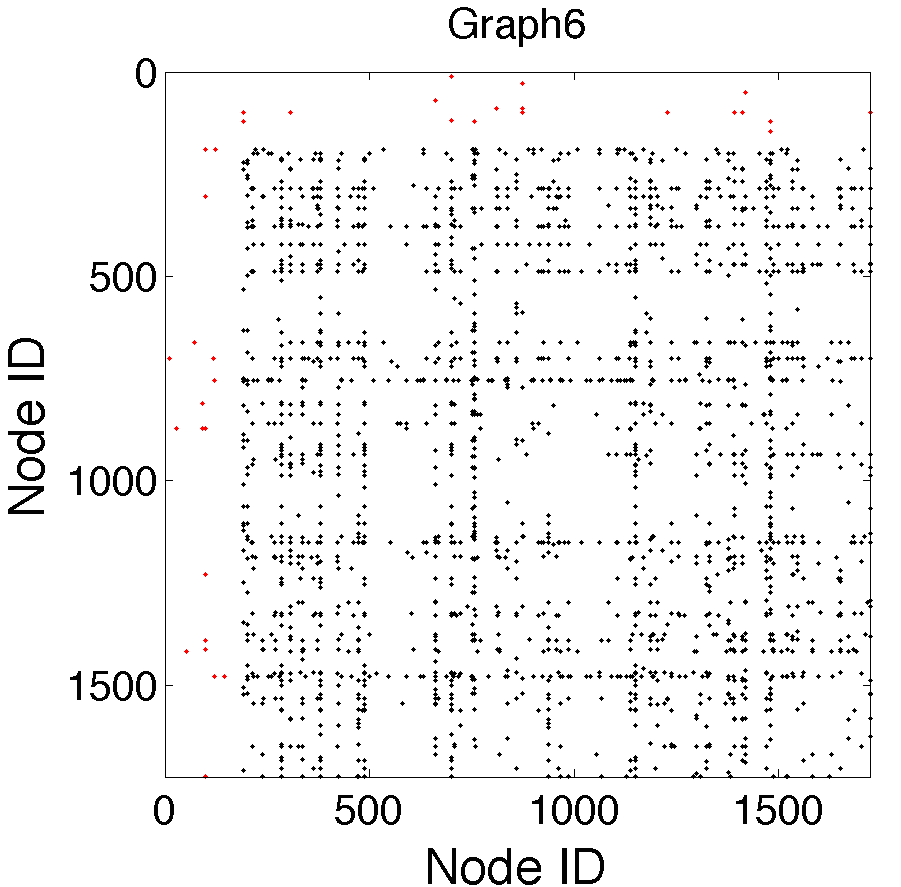}
		\end{minipage}
		&
		\begin{minipage}{.3\textwidth}
			\includegraphics[width=35mm, height=35mm]{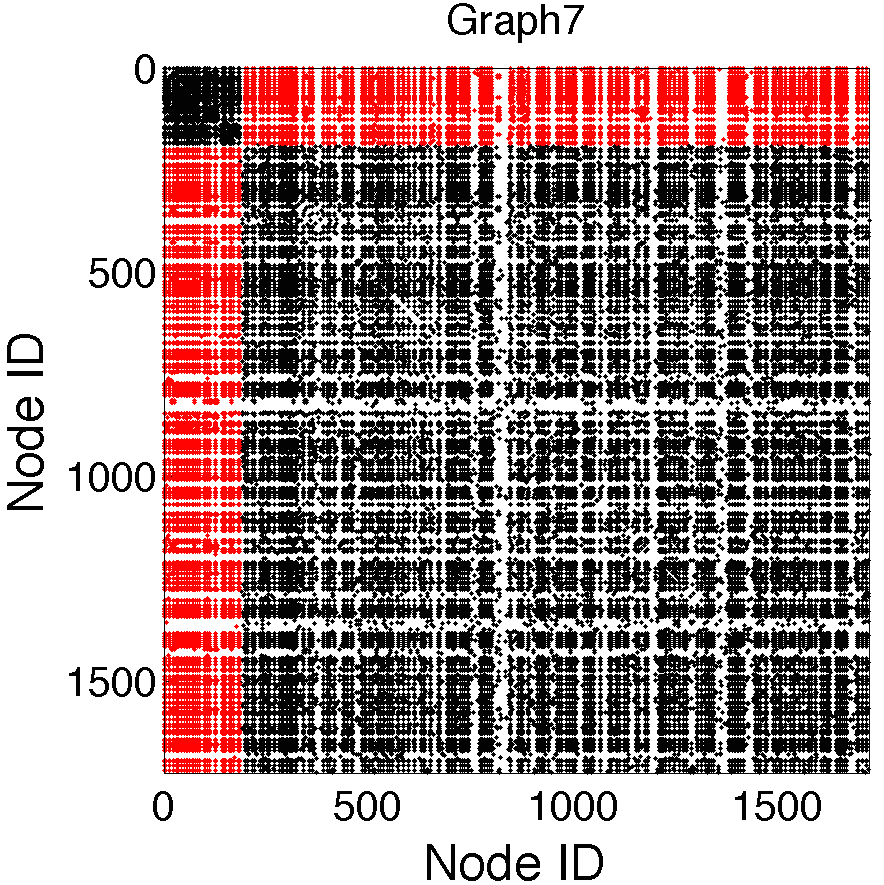}
		\end{minipage}
		&
		\begin{minipage}{.3\textwidth}
			\includegraphics[width=35mm, height=35mm]{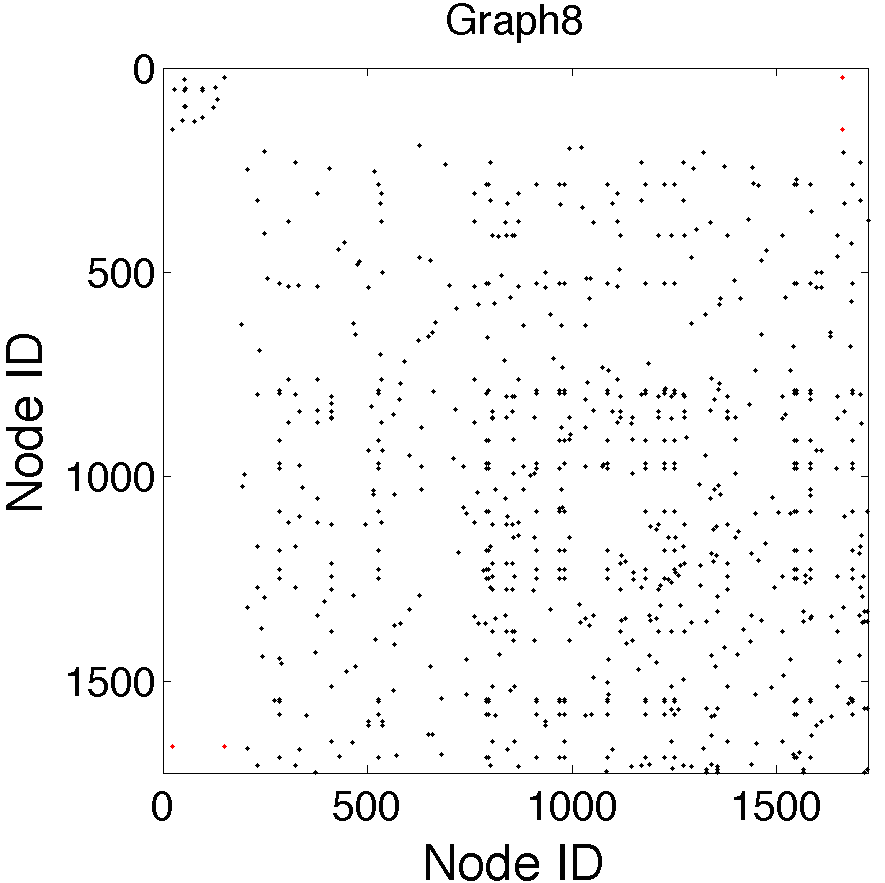}
		\end{minipage}
		\\
		\\
		\begin{minipage}{.3\textwidth}
			\includegraphics[width=35mm, height=35mm]{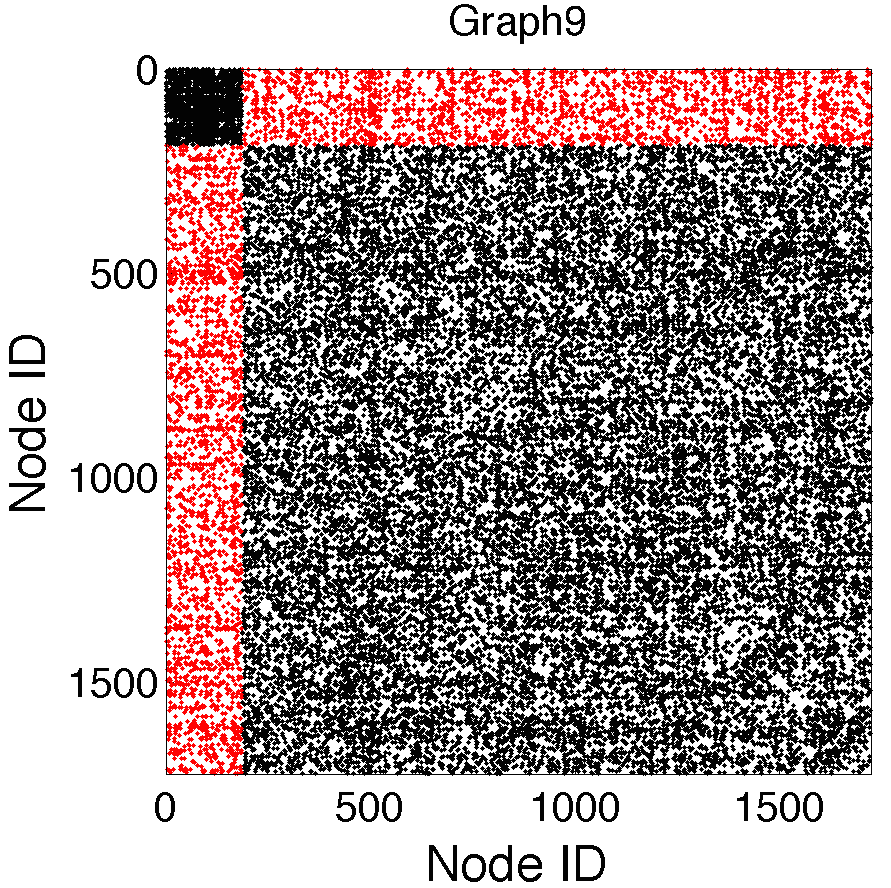}
		\end{minipage}
		&
		\begin{minipage}{.3\textwidth}
			\includegraphics[width=35mm, height=35mm]{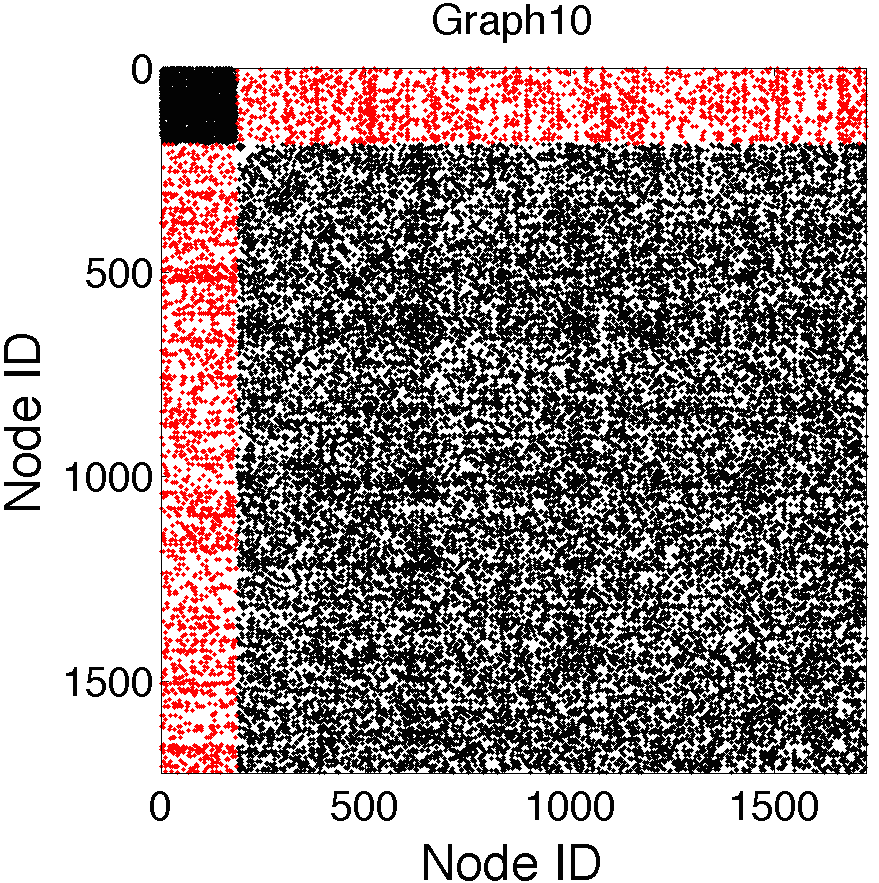}
		\end{minipage}
		&
		\begin{minipage}{.3\textwidth}
			\includegraphics[width=35mm, height=35mm]{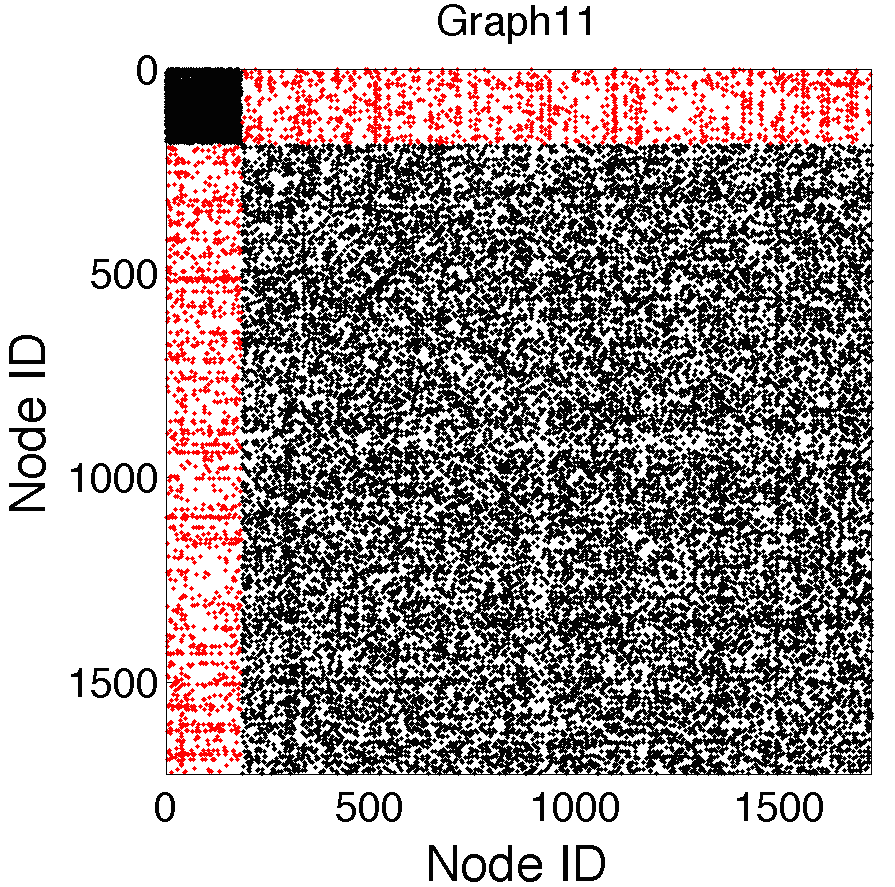}
		\end{minipage}
		&
		\begin{minipage}{.3\textwidth}
			\includegraphics[width=35mm, height=35mm]{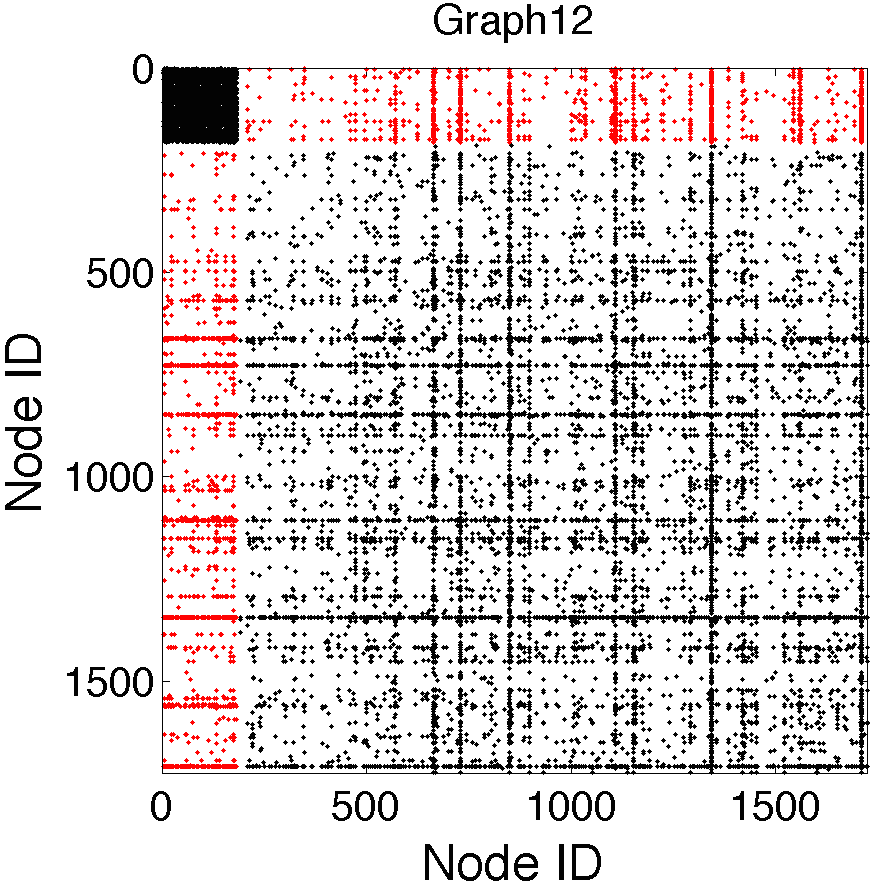}
		\end{minipage}
		\\
		\\
		\begin{minipage}{.3\textwidth}
			\includegraphics[width=35mm, height=35mm]{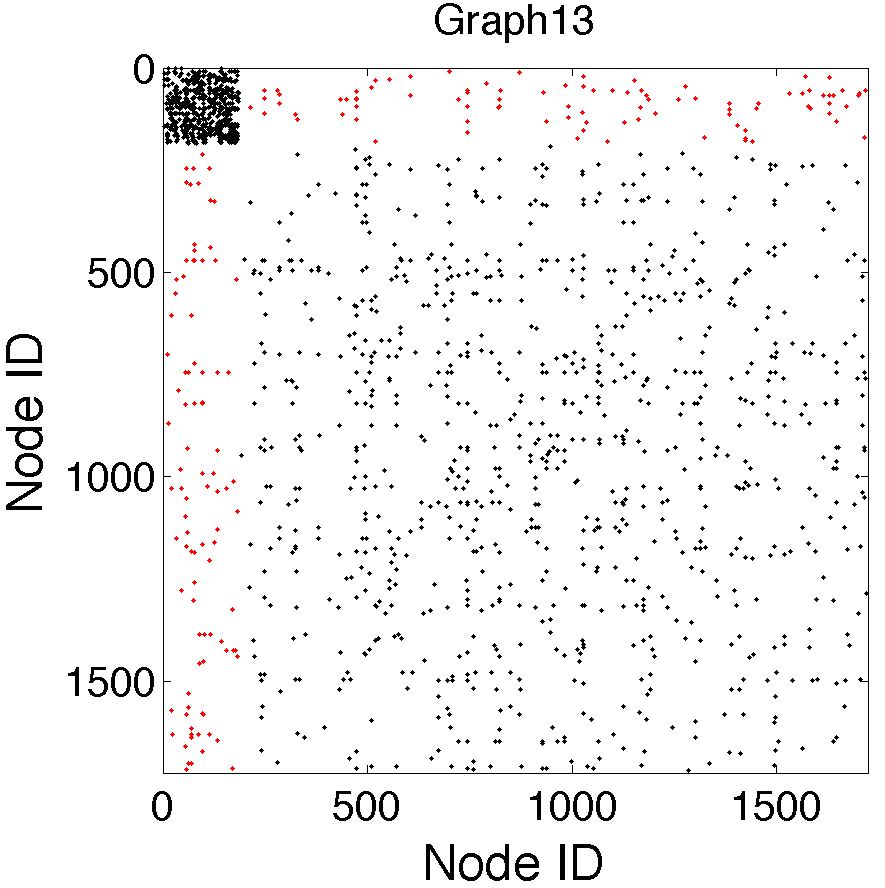}
		\end{minipage}
		&
		\begin{minipage}{.3\textwidth}
			\includegraphics[width=35mm, height=35mm]{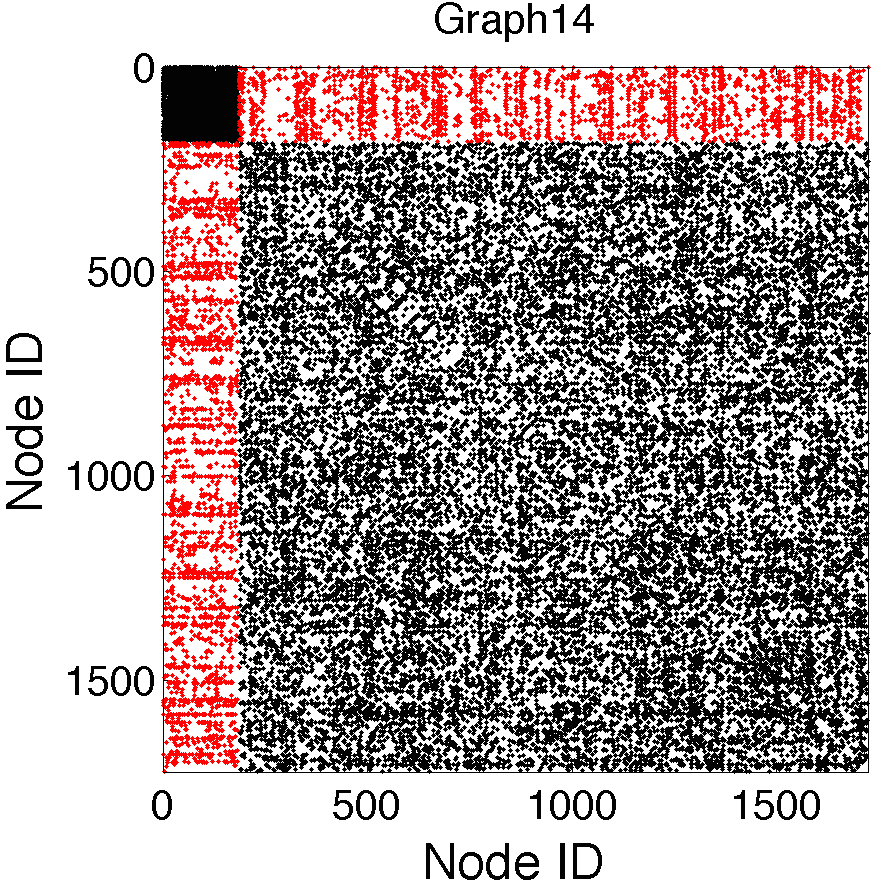}
		\end{minipage}
		&
		\begin{minipage}{.3\textwidth}
			\includegraphics[width=35mm, height=35mm]{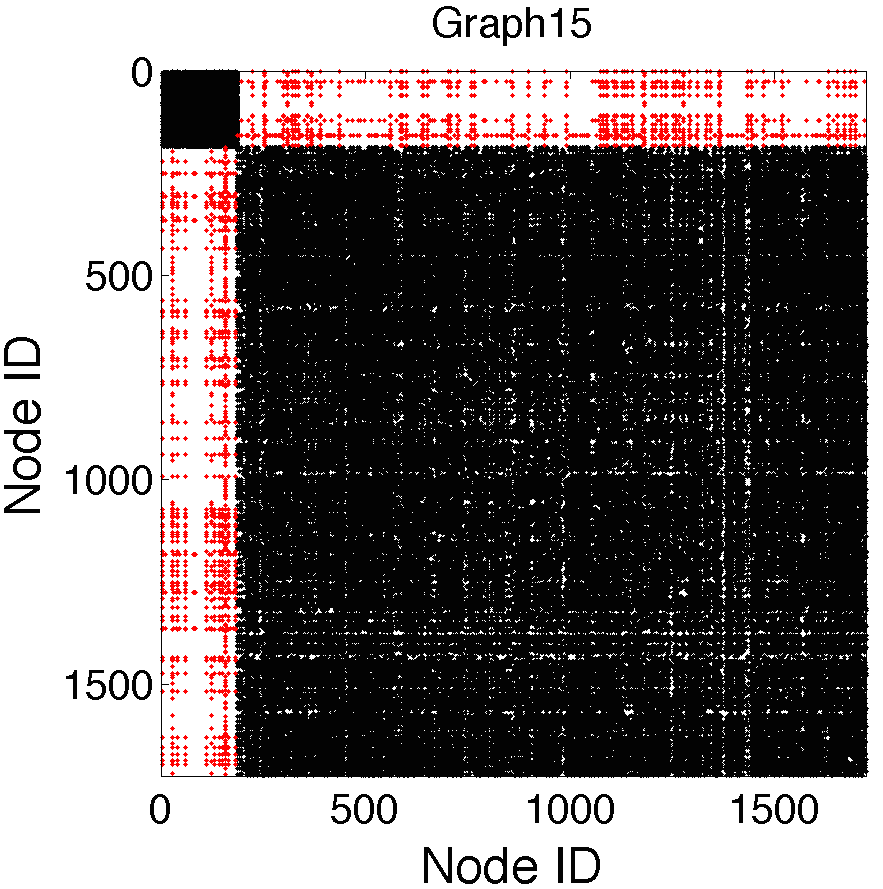}
		\end{minipage}
		\\
	\end{tabular}
\end{table}

\clearpage
\newpage

\begin{minipage}{1\textwidth}
\subsection{\bf Gene2}
We construct \textit{Gene2} in a similar way as for \textit{Gene1}. The major difference is that the labels in this dataset come from \textit{Molecular Function (MF)} domain.

\end{minipage}

\begin{table}[h!]
	\label{tab:g2}
	\centering
	\begin{tabular}{  m{4cm}  m{4cm} m{4cm}  m{4cm} }
		\begin{minipage}{.3\textwidth}
			\includegraphics[width=35mm, height=35mm]{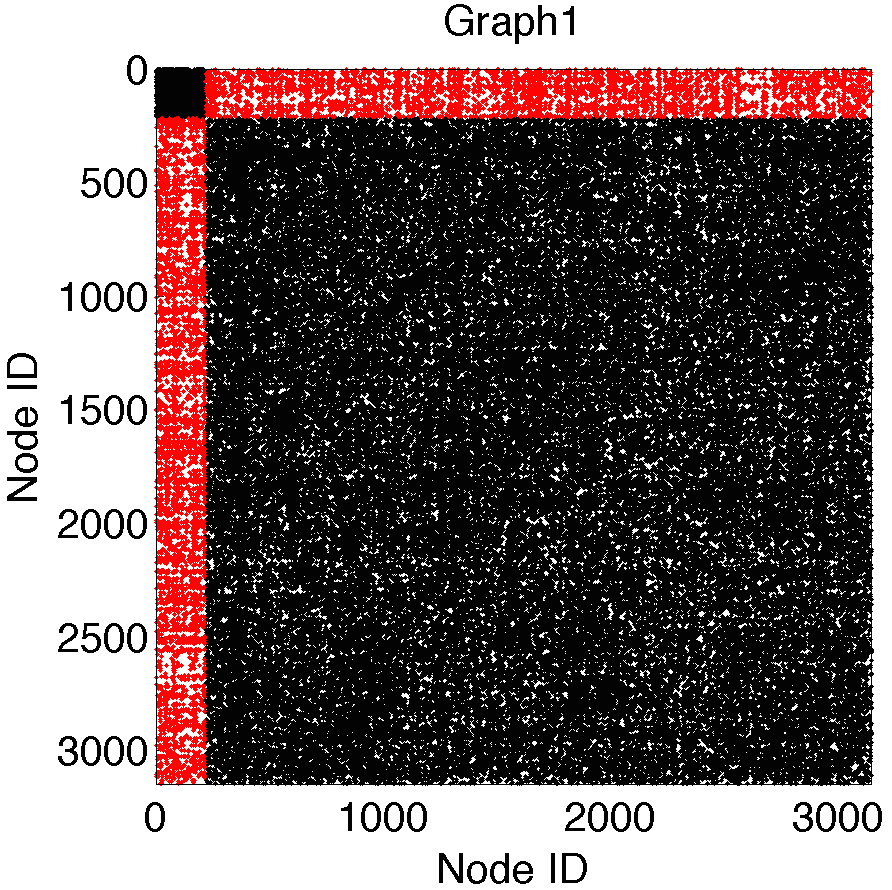}
		\end{minipage}
		&
		\begin{minipage}{.3\textwidth}
			\includegraphics[width=35mm, height=35mm]{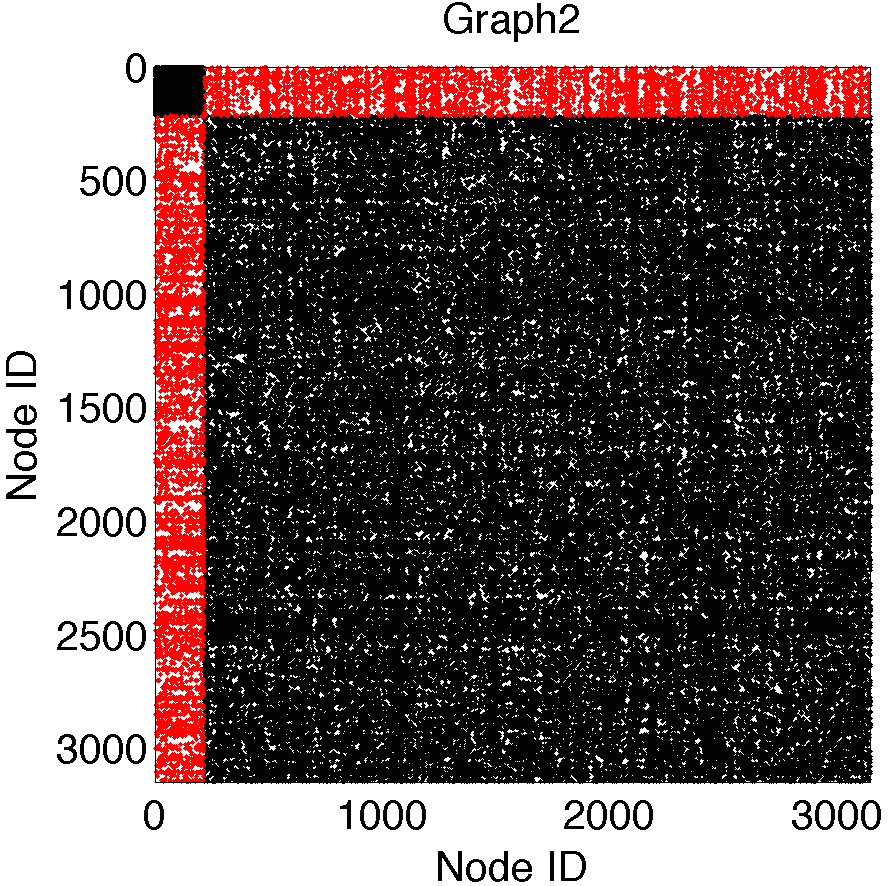}
		\end{minipage}
		&
		\begin{minipage}{.3\textwidth}
			\includegraphics[width=35mm, height=35mm]{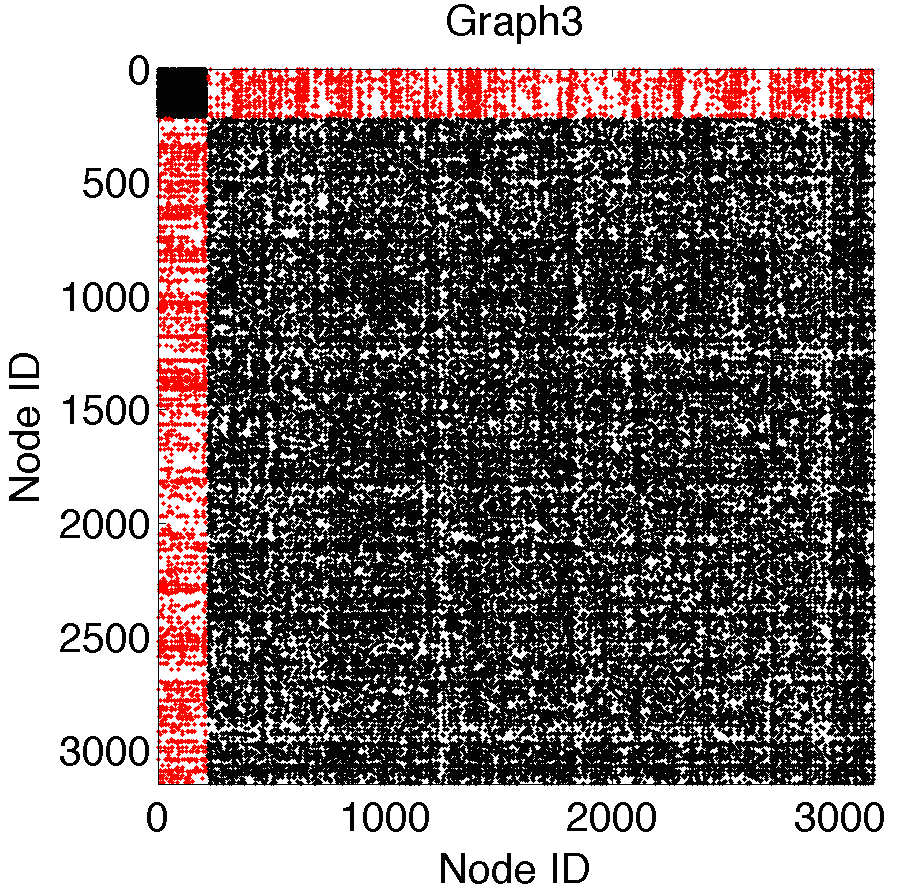}
		\end{minipage}
		&
		\begin{minipage}{.3\textwidth}
			\includegraphics[width=35mm, height=35mm]{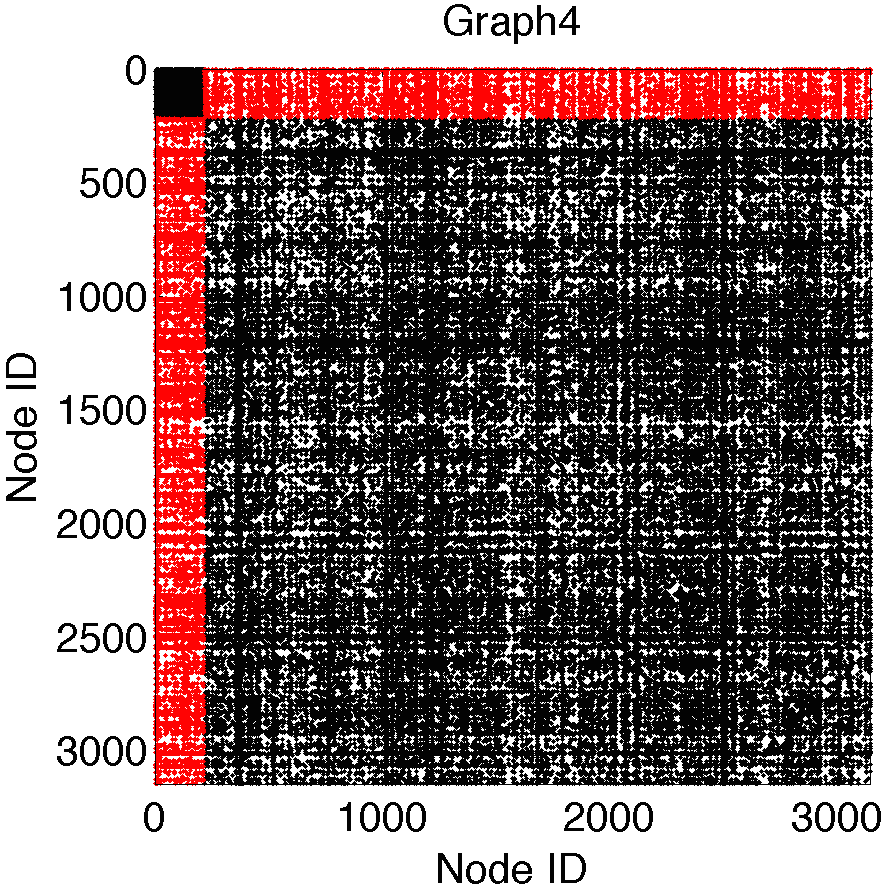}
		\end{minipage}
		\\
		\\
		\begin{minipage}{.3\textwidth}
			\includegraphics[width=35mm, height=35mm]{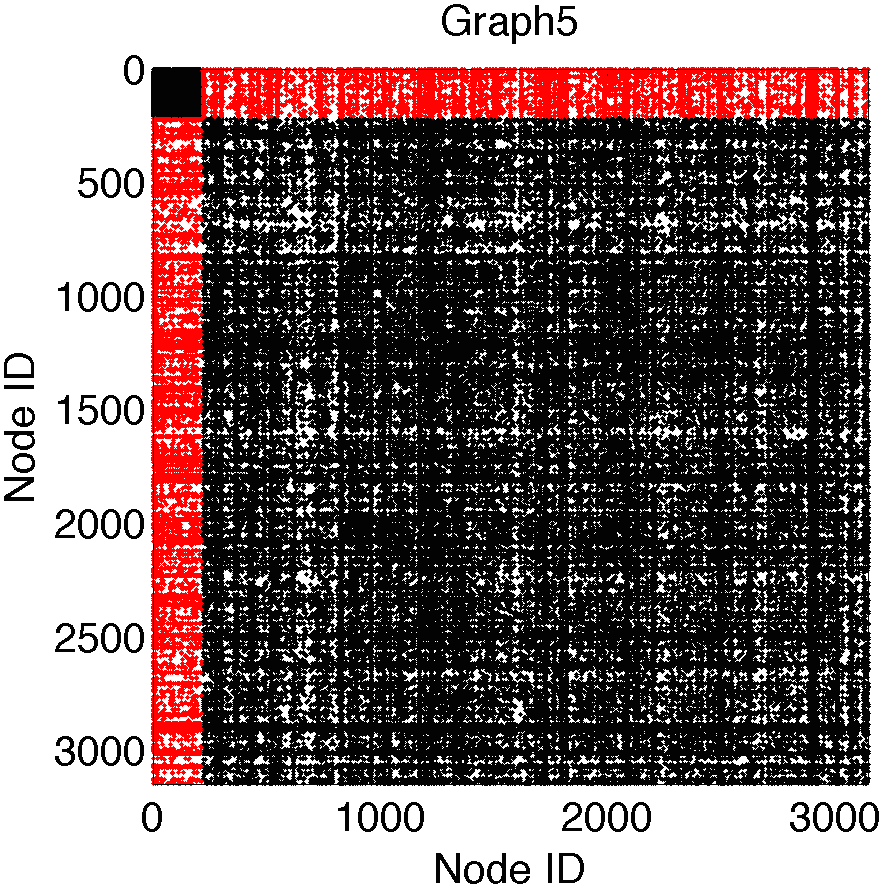}
		\end{minipage}
		&
		\begin{minipage}{.3\textwidth}
			\includegraphics[width=35mm, height=35mm]{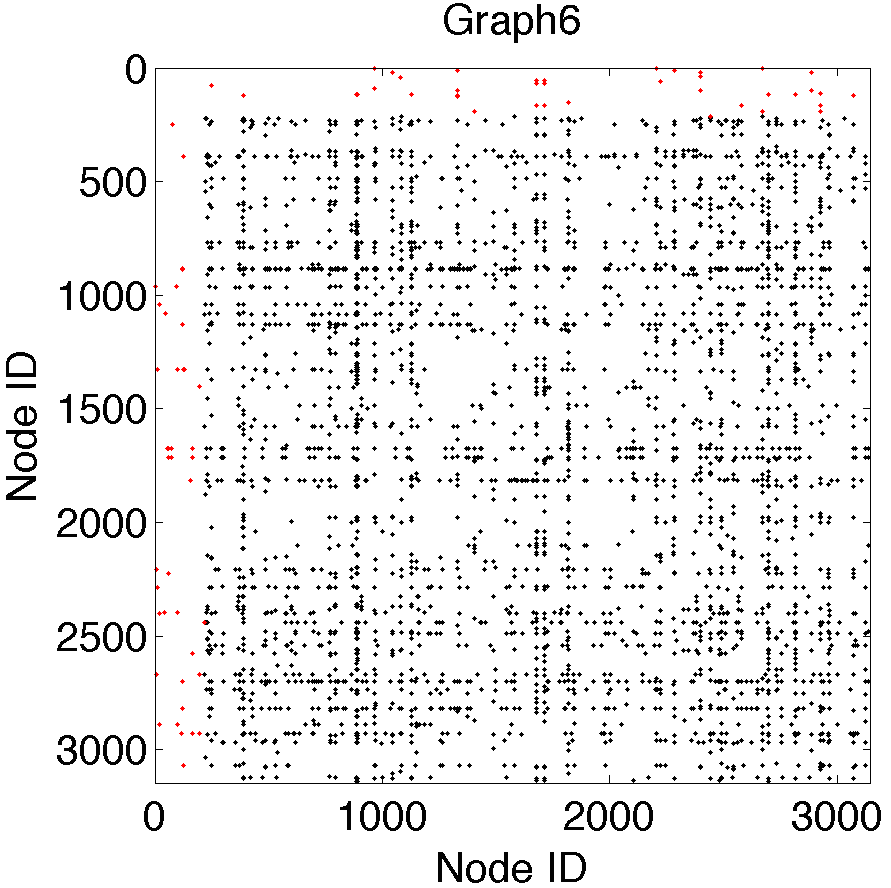}
		\end{minipage}
		&
		\begin{minipage}{.3\textwidth}
			\includegraphics[width=35mm, height=35mm]{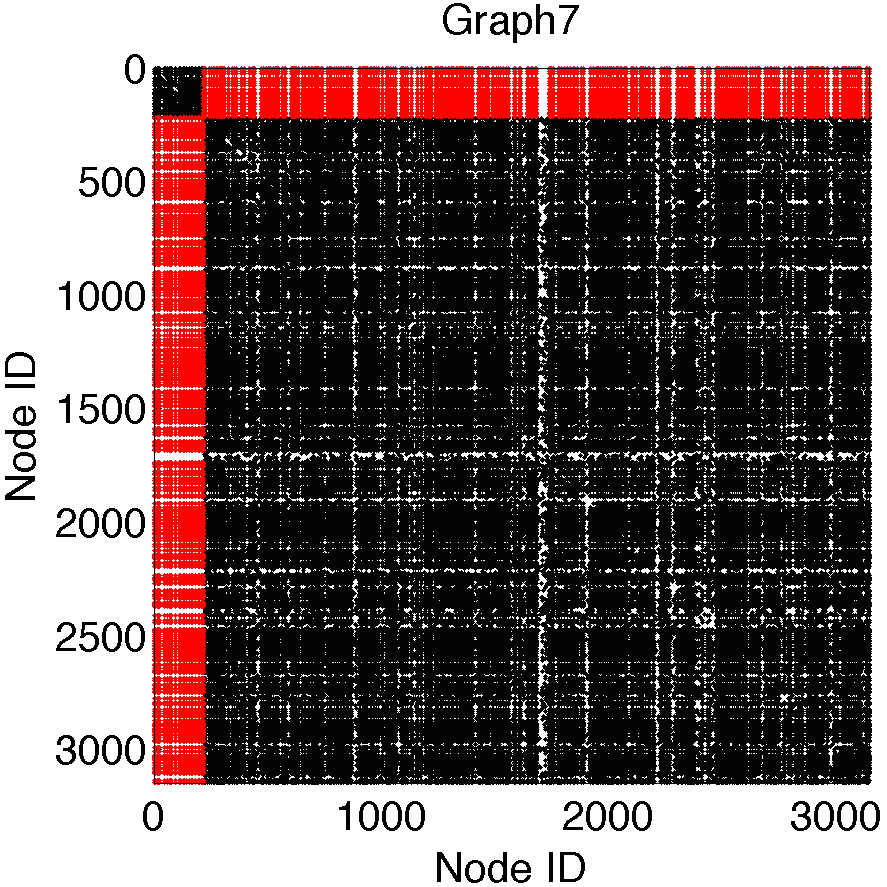}
		\end{minipage}
		&
		\begin{minipage}{.3\textwidth}
			\includegraphics[width=35mm, height=35mm]{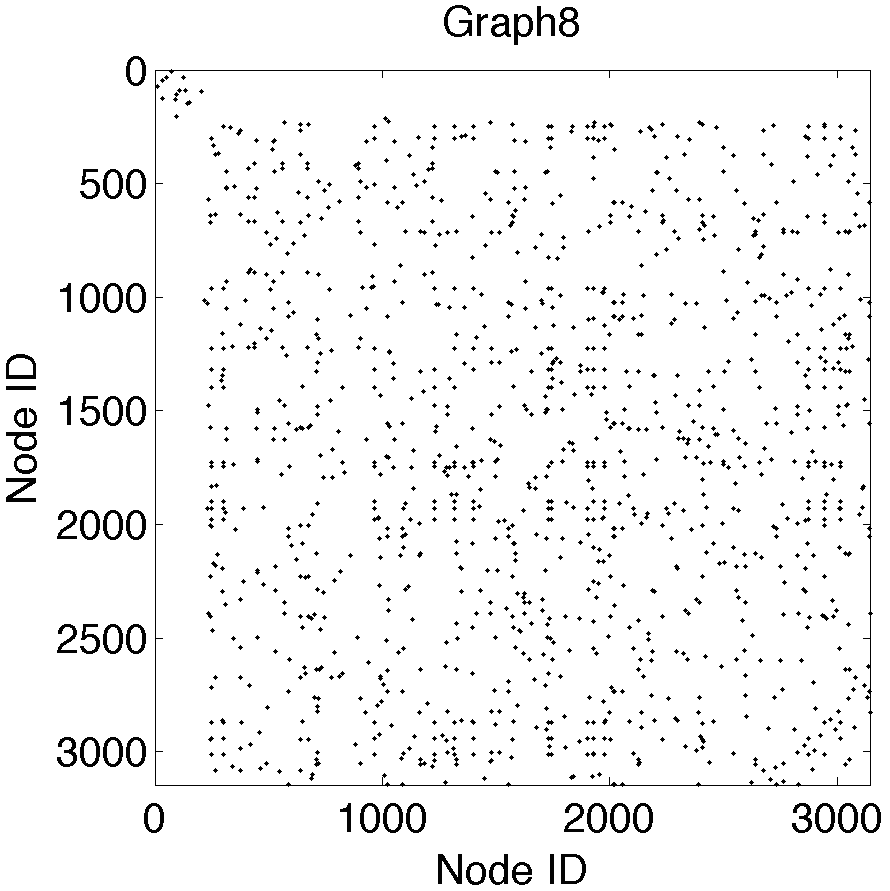}
		\end{minipage}
		\\
		\\
		\begin{minipage}{.3\textwidth}
			\includegraphics[width=35mm, height=35mm]{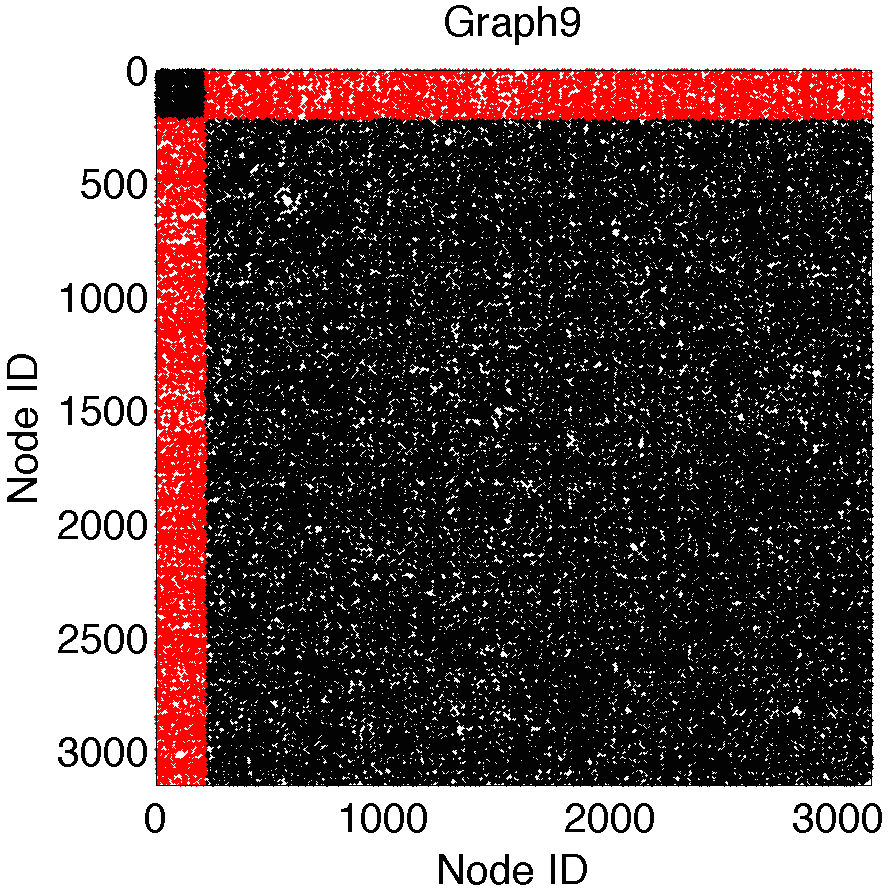}
		\end{minipage}
		&
		\begin{minipage}{.3\textwidth}
			\includegraphics[width=35mm, height=35mm]{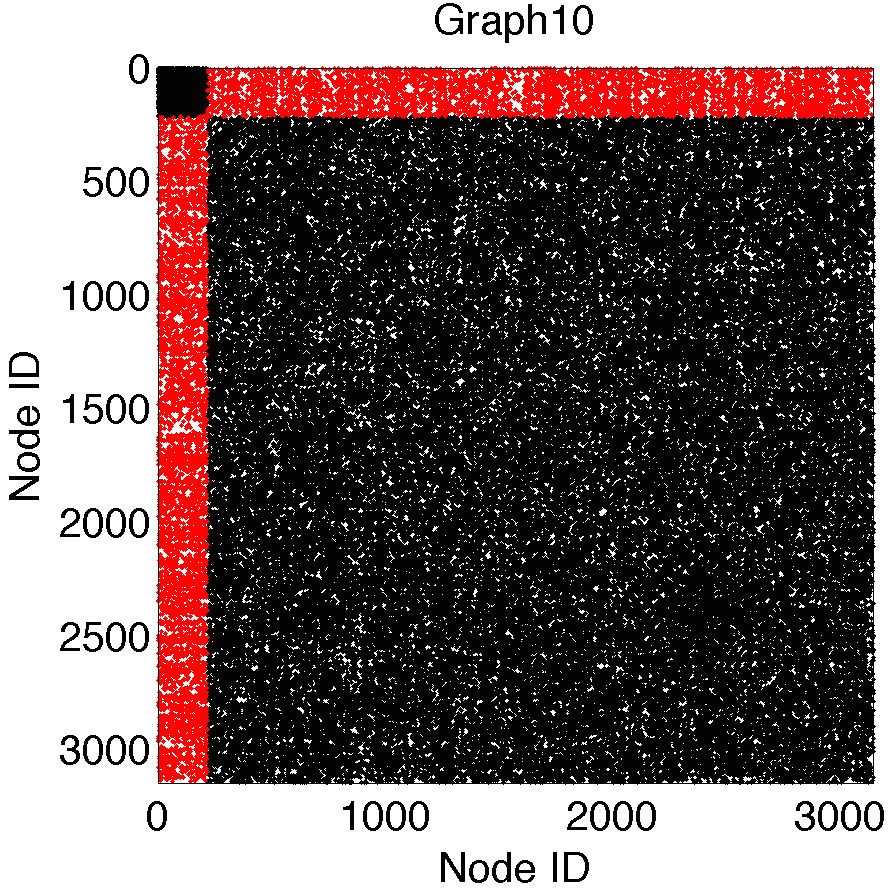}
		\end{minipage}
		&
		\begin{minipage}{.3\textwidth}
			\includegraphics[width=35mm, height=35mm]{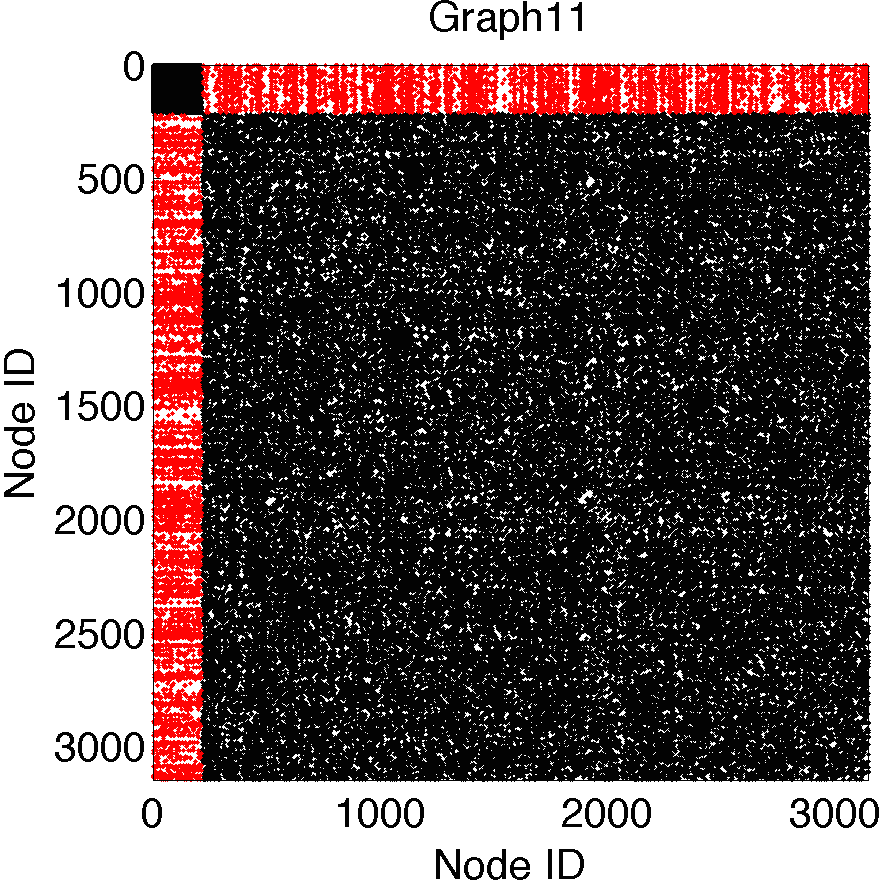}
		\end{minipage}
		&
		\begin{minipage}{.3\textwidth}
			\includegraphics[width=35mm, height=35mm]{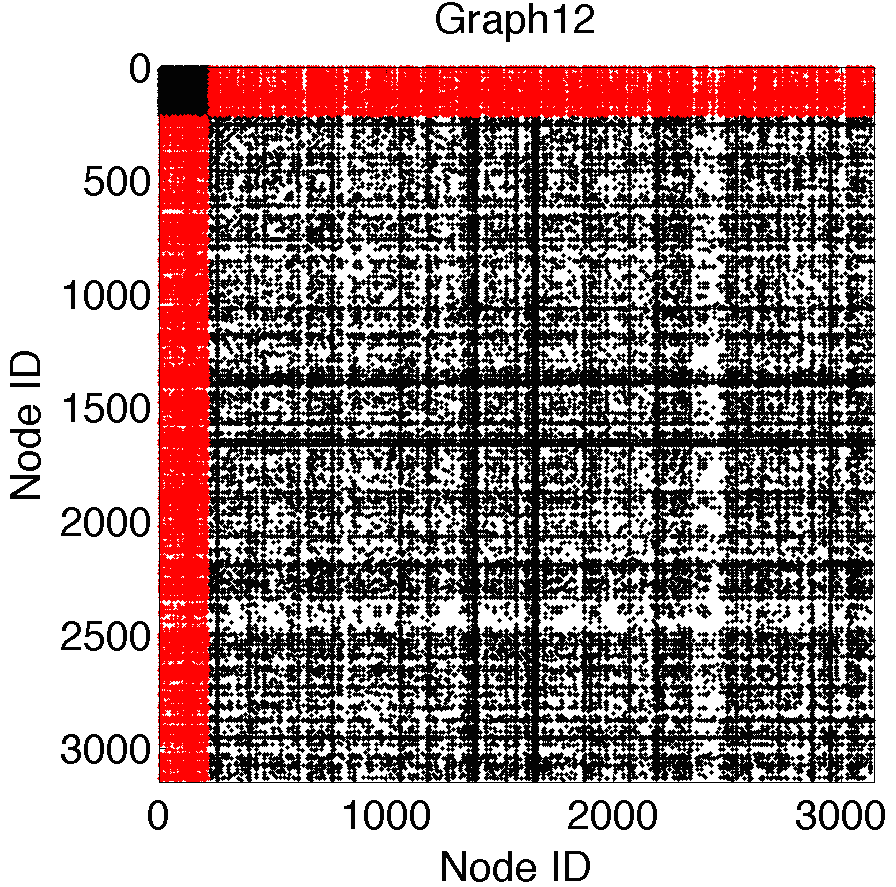}
		\end{minipage}
		\\
		\\
		\begin{minipage}{.3\textwidth}
			\includegraphics[width=35mm, height=35mm]{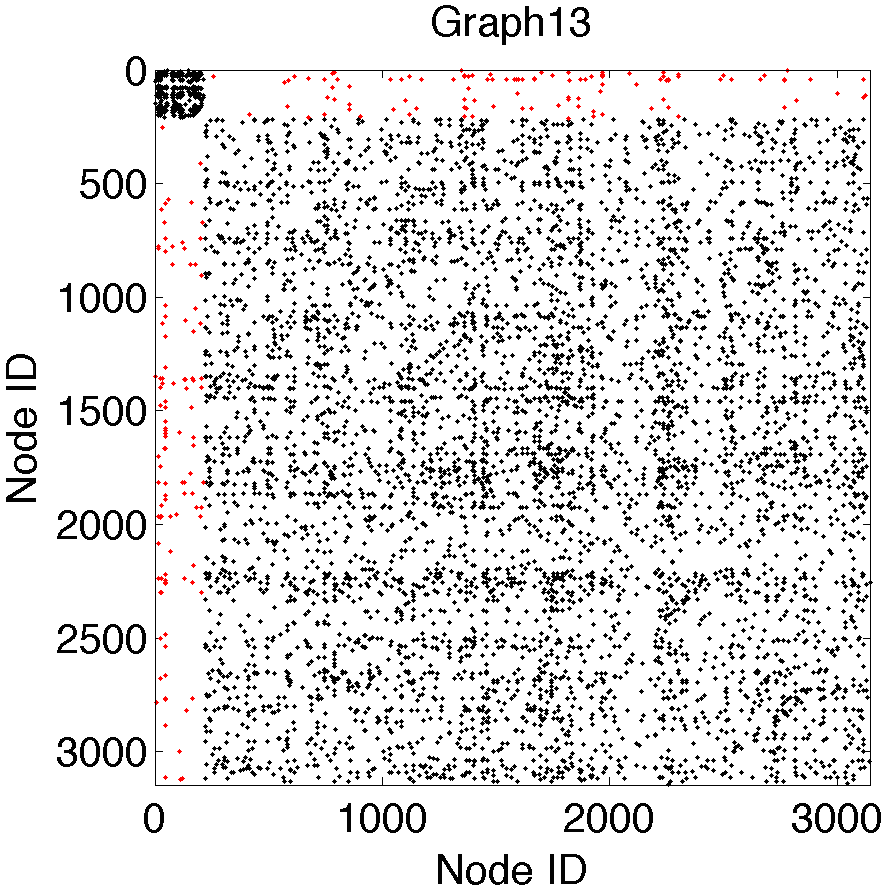}
		\end{minipage}
		&
		\begin{minipage}{.3\textwidth}
			\includegraphics[width=35mm, height=35mm]{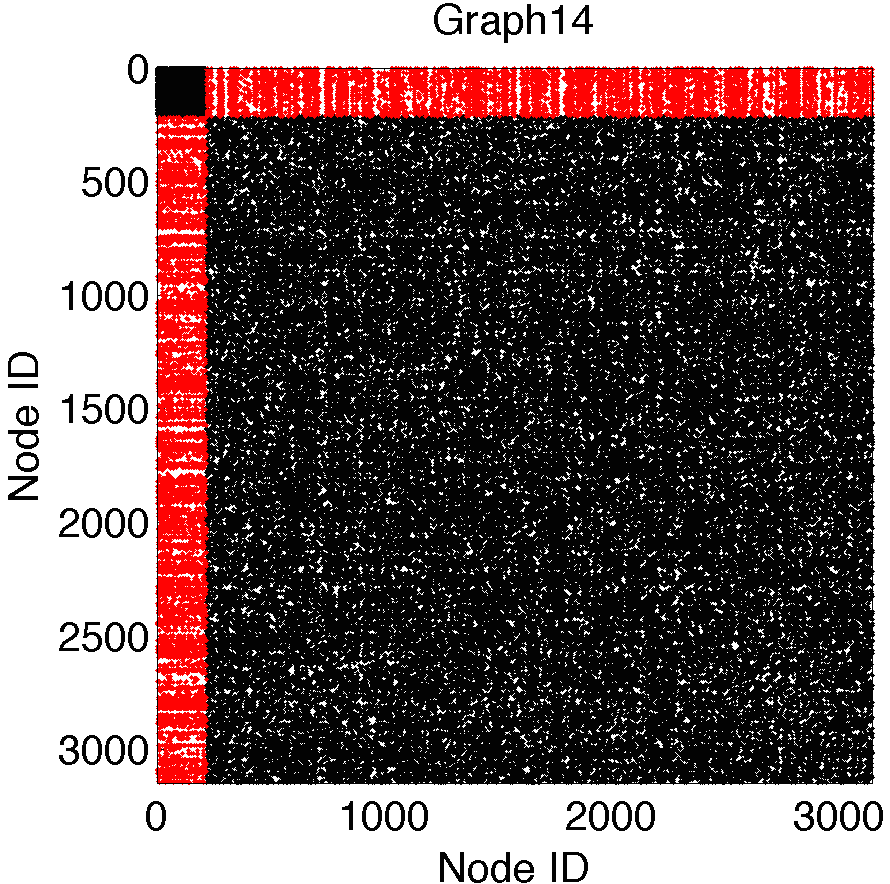}
		\end{minipage}
		&
		\begin{minipage}{.3\textwidth}
			\includegraphics[width=35mm, height=35mm]{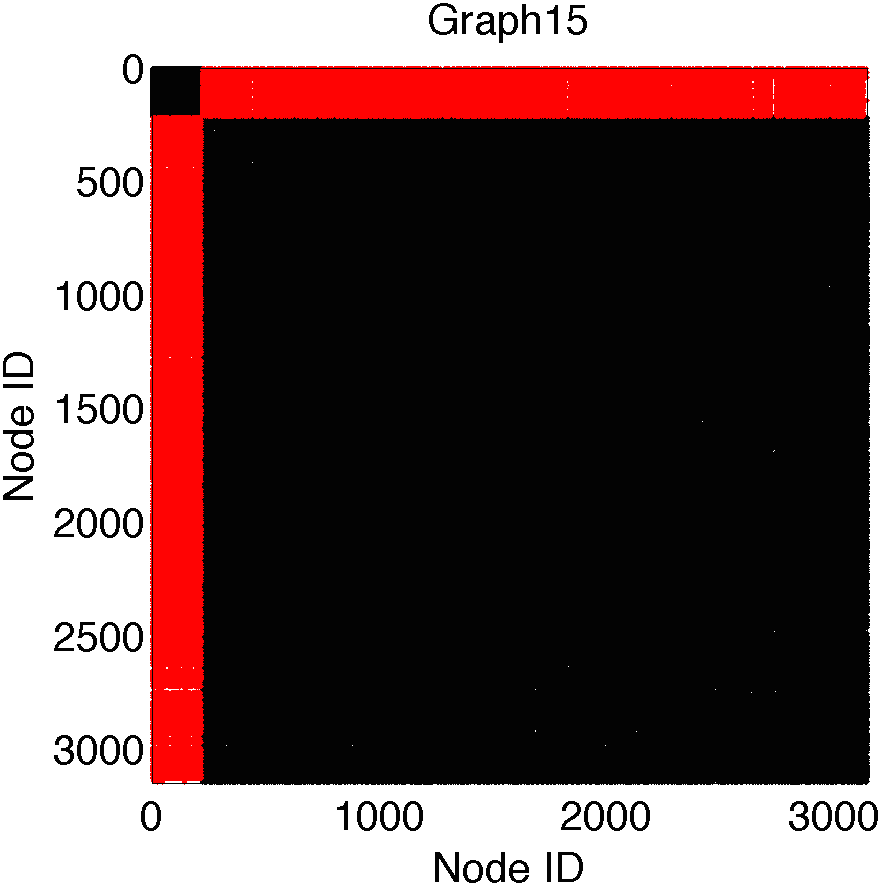}
		\end{minipage}
		\\
	\end{tabular}
\end{table}
\clearpage
\newpage

\begin{minipage}{1\textwidth}	
\section{\large{Precision-Recall Plots on Real-World Multi-graphs}}
\label{sec:ap}
\end{minipage}

\begin{table}[H]
	\begin{tabular}{  m{4cm}  m{4cm} m{4cm}  m{4cm} }
		\begin{minipage}{.3\textwidth}
			\includegraphics[width=42mm, height=33mm]{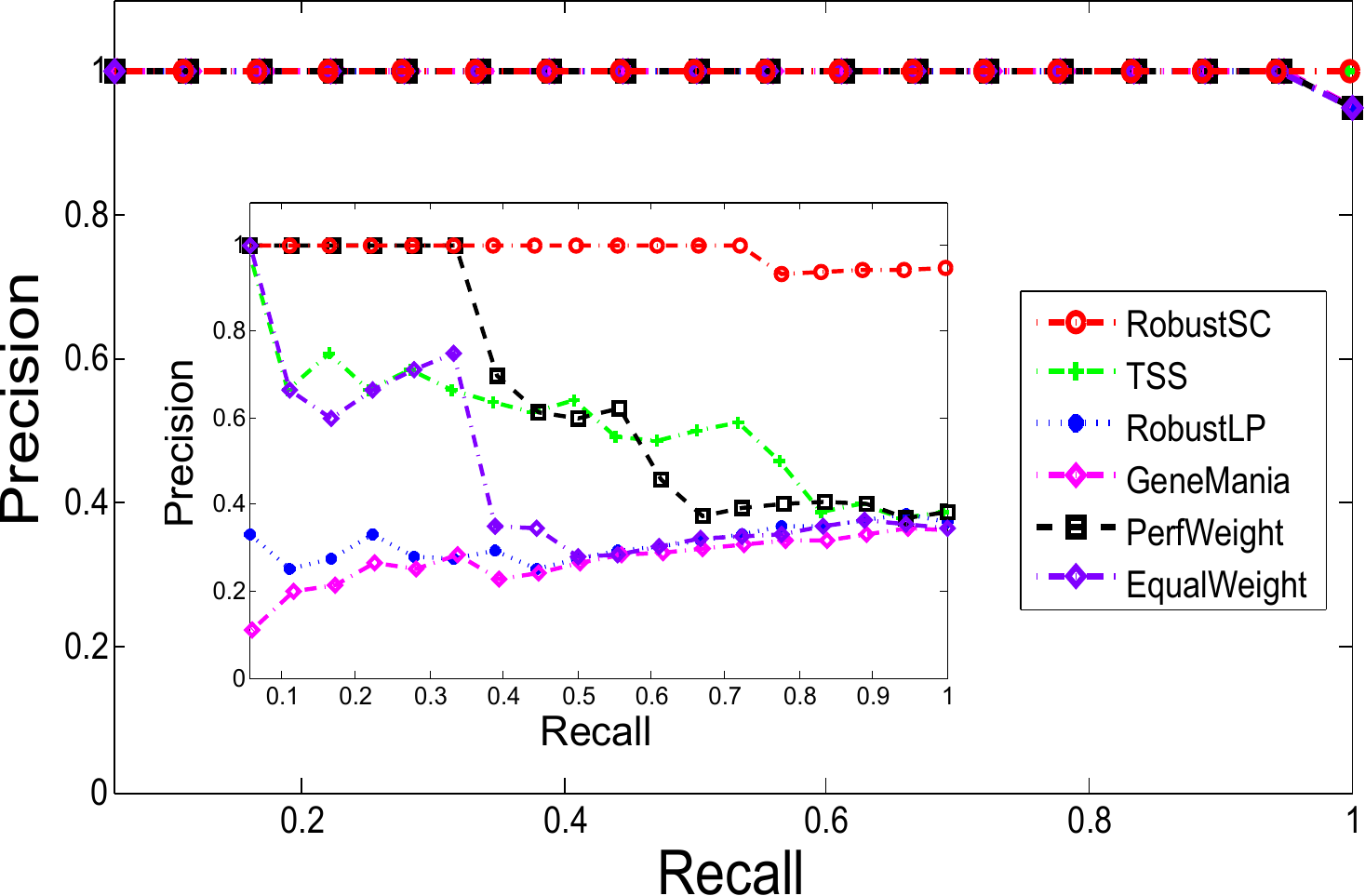}
			
		\end{minipage}
		&
		\begin{minipage}{.3\textwidth}
			\includegraphics[width=42mm, height=35mm]{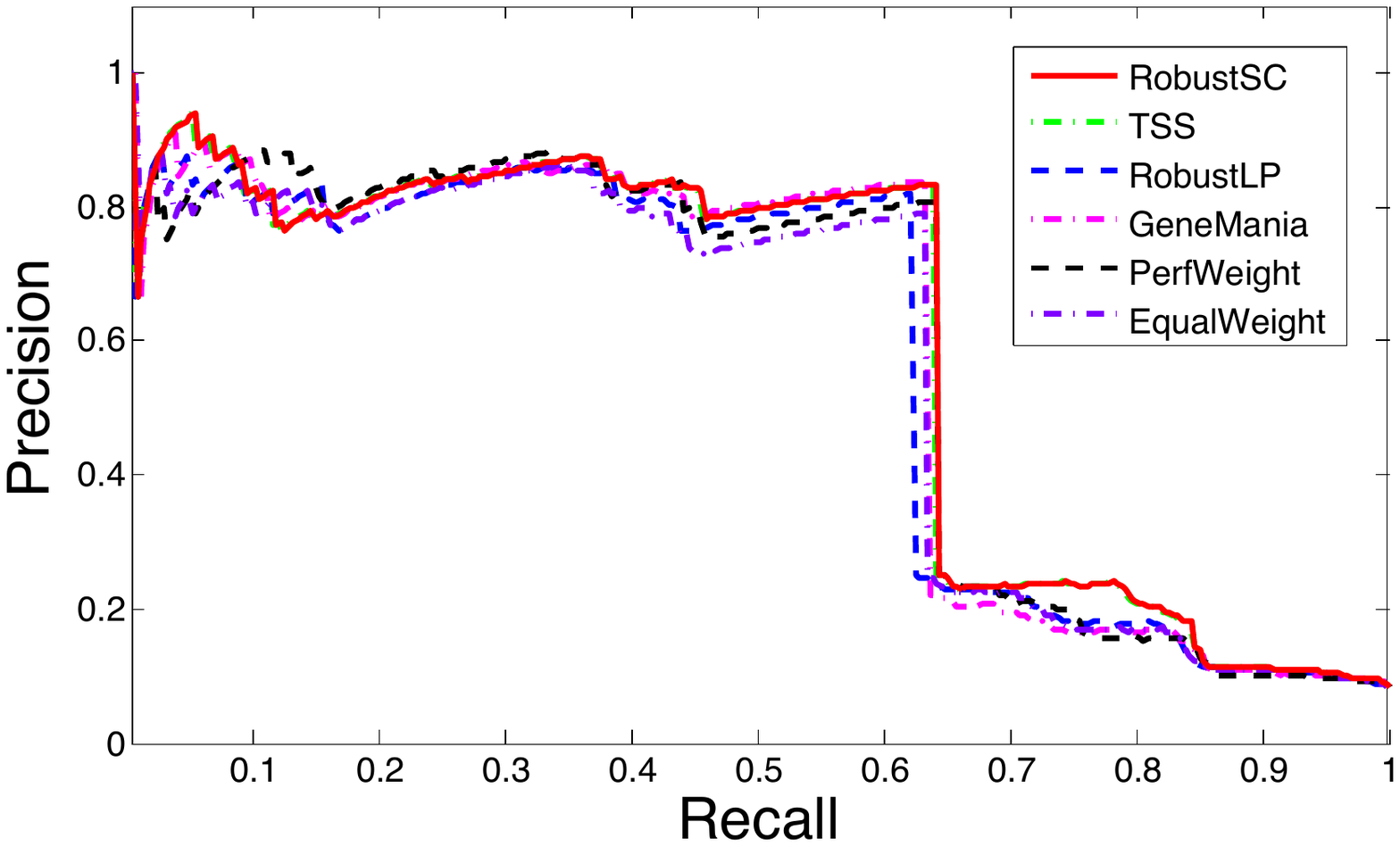}
		\end{minipage}
		&
		\begin{minipage}{.3\textwidth}
			\includegraphics[width=42mm, height=35mm]{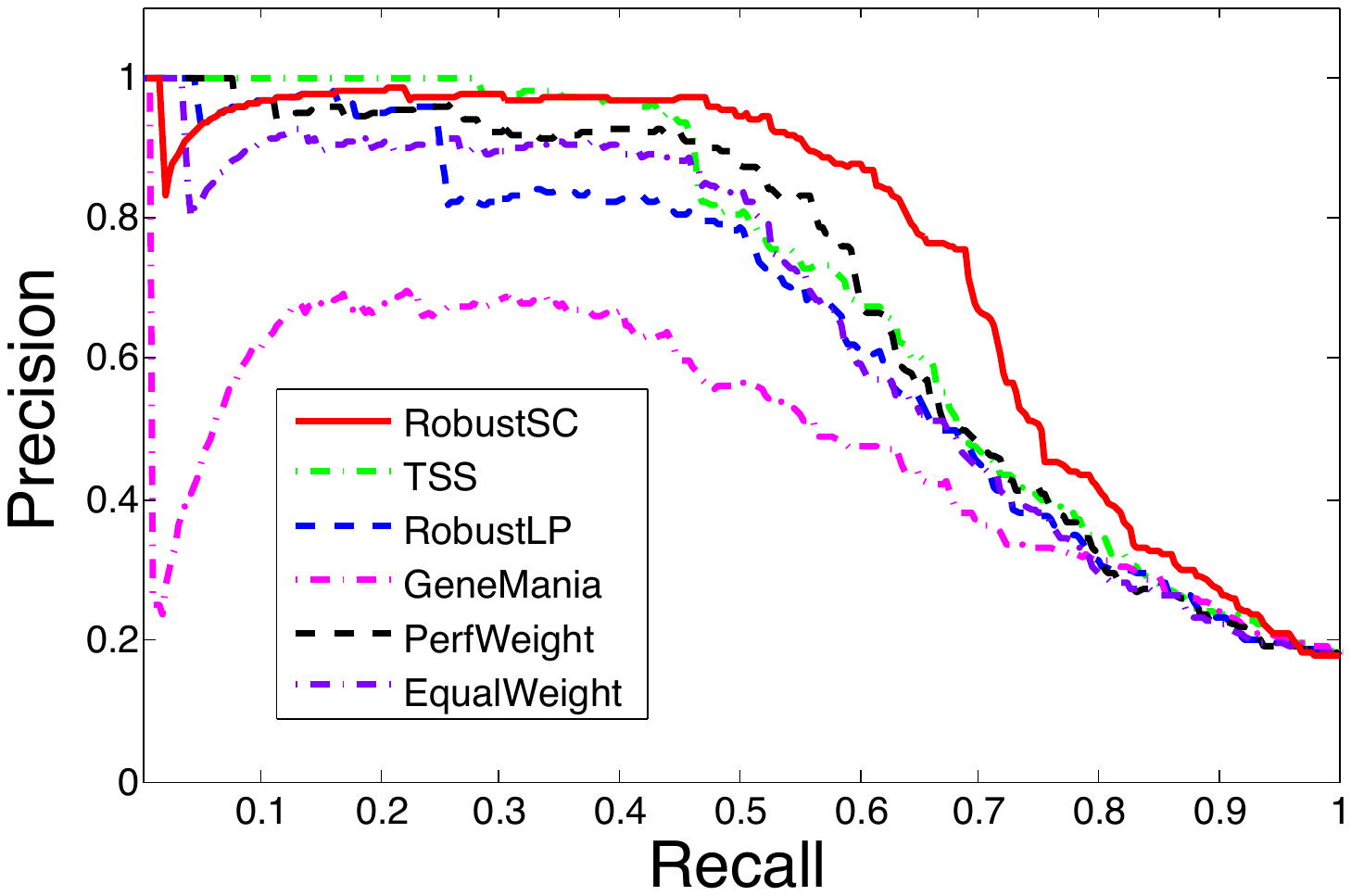}
		\end{minipage}
		&
		\begin{minipage}{.3\textwidth}
			\includegraphics[width=42mm, height=35mm]{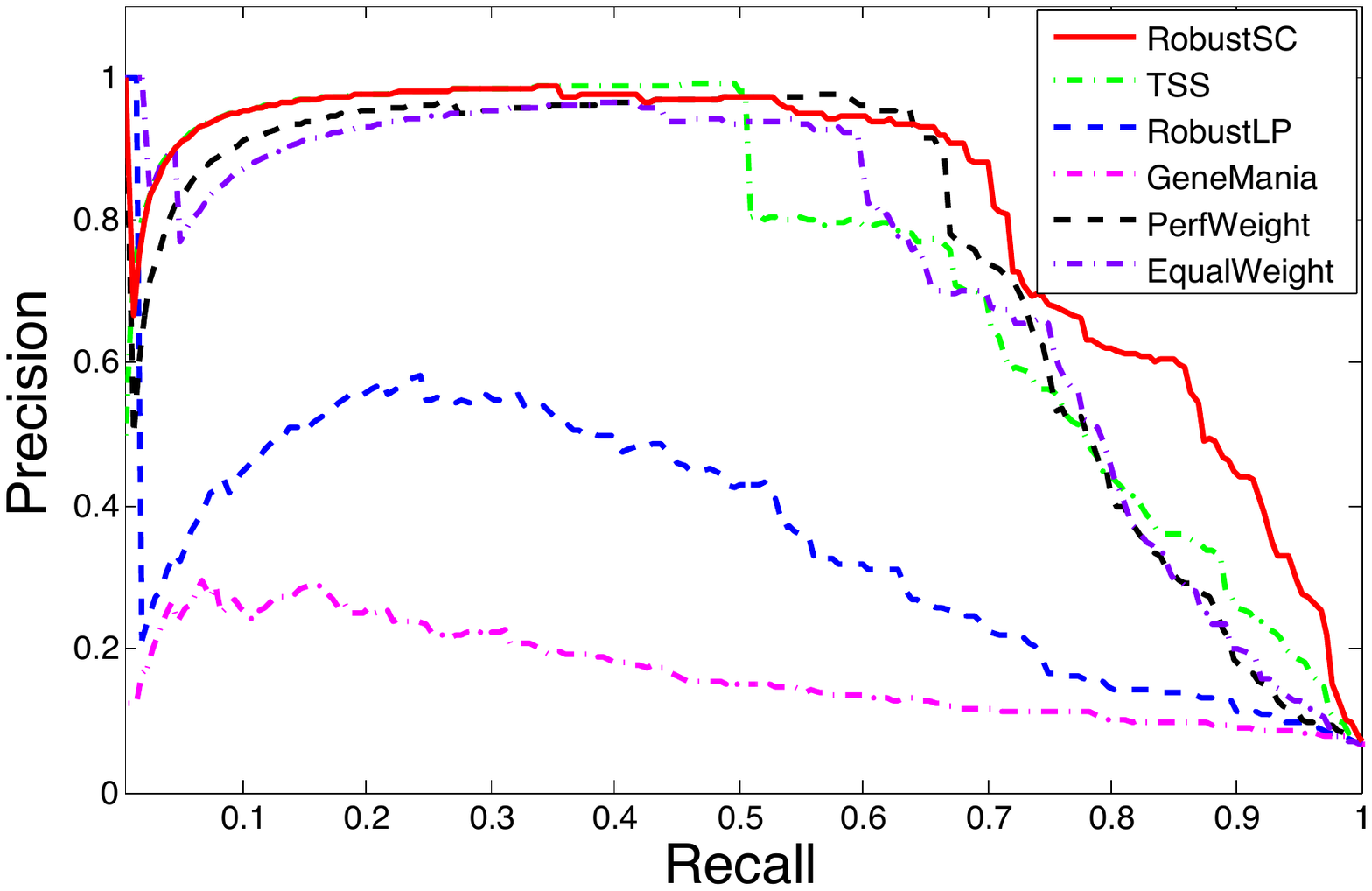}
		\end{minipage}
		\\\centering
		(a) \textit{RealityMining} &
		\centering
		(b) \textit{Protein} &
		\centering
		(c) \textit{Gene1} &
		\centering
		(d) \textit{Gene2}
	\end{tabular}
	\begin{minipage}{1\textwidth}
	\caption{Precision vs. Recall of competing methods in four real-world multi-graphs: (a) {\em RealityMining} (4 graphs), (b) {\em Protein} (5 graphs),
		(c) {\em Gene1} (15 graphs), and (d) {\em Gene2} (15 graphs).
		Inset plot in (a) shows performance when 6 rewired graphs with low intensity are injected to {\em RealityMining} multi-graph---noise hinders existing methods dramatically, whereas \method~remains near-stable.}
	\end{minipage}
\end{table}

\begin{minipage}{1\textwidth}
\section{\large{Additional Noise-Test Results}}
\label{sec:rm}

We provide additional noise-testing results on RealityMining with 2 and 4 injected graphs, under 3 noise models and 2 intensity levels in Table \ref{tab:more}.
\end{minipage}
\begin{table}[H]
		\vspace{-0.1in}
		\begin{minipage}{1\textwidth}
		\caption{Performance of all methods on real-world multi-graphs. RM also injected with 2 and 4 intrusive graphs with various settings. Values depict mean and standard deviation Average Precision (AP) (over 10 runs with different labeled set). 	\label{tab:more}}
	\end{minipage}
	{\fontsize{8}{9.5}
	\selectfont
		\begin{tabular}{p{0.38in}|p{0.3in}|p{0.57in}|p{0.3in}|>{\bfseries} p{0.54in}|p{0.65in}|p{0.65in}|p{0.54in}|p{0.62in}|p{0.65in}}
			\hline   Dataset 					& \#Graphs 	    & NoiseModel    & Intensity & {\sc Proposed}        & \perf 						& \eql 						&  \tss 									&  \rob 					&  \gene \\ 
			\hline   \multirow{13}{*}{RM} &  4+2			&  Adversarial    & Low		   & 0.970$\pm$0.004    &  0.707$\pm$0.050 &  0.554$\pm$0.049 &  0.525$\pm$0.075 			&  0.851$\pm$0.025 &  0.470$\pm$0.097 \\ 
			\cline{2-10}  						&  4+2				&  Adversarial    & High		& 0.970$\pm$0.004   &  0.611$\pm$0.049 & 0.484$\pm$0.045   & 0.554$\pm$0.048  			&   0.809$\pm$0.044 & 0.359$\pm$0.068   \\
			\cline{2-10}   						&  4+2				&  Rewire 			& Low   	 & 0.970$\pm$0.004   &  0.695$\pm$0.054 & 0.669$\pm$0.050  & 0.718$\pm$0.061 				& 0.873$\pm$0.021  & 0.509$\pm$0.086      \\ 
			\cline{2-10}   						&  4+2				&  Rewire 			& High        & 0.970$\pm$0.004   &  0.563$\pm$0.064 &  0.537$\pm$0.062 &  0.646$\pm$0.066 				&  0.824$\pm$0.045 &  0.290$\pm$0.050\\ 
			\cline{2-10}   						&  4+2				&  Erdos-Renyi  &  Low		   & 0.970$\pm$0.004  &  0.905$\pm$0.024 &  0.841$\pm$0.042 & 0.657$\pm$0.081  				&  0.928$\pm$0.011 &  0.773$\pm$0.067 \\ 
			\cline{2-10}   						&  4+2				&  Erdos-Renyi  &  High		   & 0.970$\pm$0.004  &  0.942$\pm$0.016 &  0.920$\pm$0.022  & 0.895$\pm$0.042  			&  0.930$\pm$0.015 &  0.883$\pm$0.055\\ 
			\cline{2-10}
			\cline{2-10}   						&  4+4				&  Adversarial    & Low		   & 0.970$\pm$0.004    &  0.531$\pm$0.042 &  0.372$\pm$0.040 &  0.390$\pm$0.047 			&  0.427$\pm$0.063 &  0.260$\pm$0.029 \\ 
			\cline{2-10}  						&  4+4				&  Adversarial    & High		& 0.970$\pm$0.004   &  0.383$\pm$0.045 & 0.297$\pm$0.025   & 0.576$\pm$0.057  			&   0.339$\pm$0.051 & 0.215$\pm$0.010   \\
			\cline{2-10}   						&  4+4				&  Rewire 			& Low   	 & 0.930$\pm$0.004   &  0.610$\pm$0.062 & 0.503$\pm$0.058  & 0.561$\pm$0.062 				& 0.505$\pm$0.066  & 0.319$\pm$0.046      \\ 
			\cline{2-10}   						&  4+4				&  Rewire 			& High        & 0.907$\pm$0.004   &  0.437$\pm$0.054 &  0.349$\pm$0.044 &  0.542$\pm$0.062 				&  0.334$\pm$0.048 &  0.217$\pm$0.011\\ 
			\cline{2-10}   						&  4+4				&  Erdos-Renyi  &  Low		   & 0.970$\pm$0.004  &  0.867$\pm$0.029 &  0.770$\pm$0.045 & 0.482$\pm$0.084  				&  0.869$\pm$0.034 &  0.698$\pm$0.094 \\ 
			\cline{2-10}   						&  4+4				&  Erdos-Renyi  &  High		   & 0.970$\pm$0.004  &  0.942$\pm$0.016 &  0.917$\pm$0.024  & 0.659$\pm$0.078  			&  0.925$\pm$0.026 &  0.834$\pm$0.068\\ 
			\hline 
		\end{tabular}}
\end{table}

\begin{minipage}{1\textwidth}
\section{\large{Graph Removal Order of \method~on Real Datasets}}
\label{sec:order}
\end{minipage}

\begin{figure}[H]
	\centering
	\begin{minipage}{1\textwidth}
		\centering
	\begin{tabular}{c}
		\includegraphics[width=0.55\linewidth,height=1.7in]{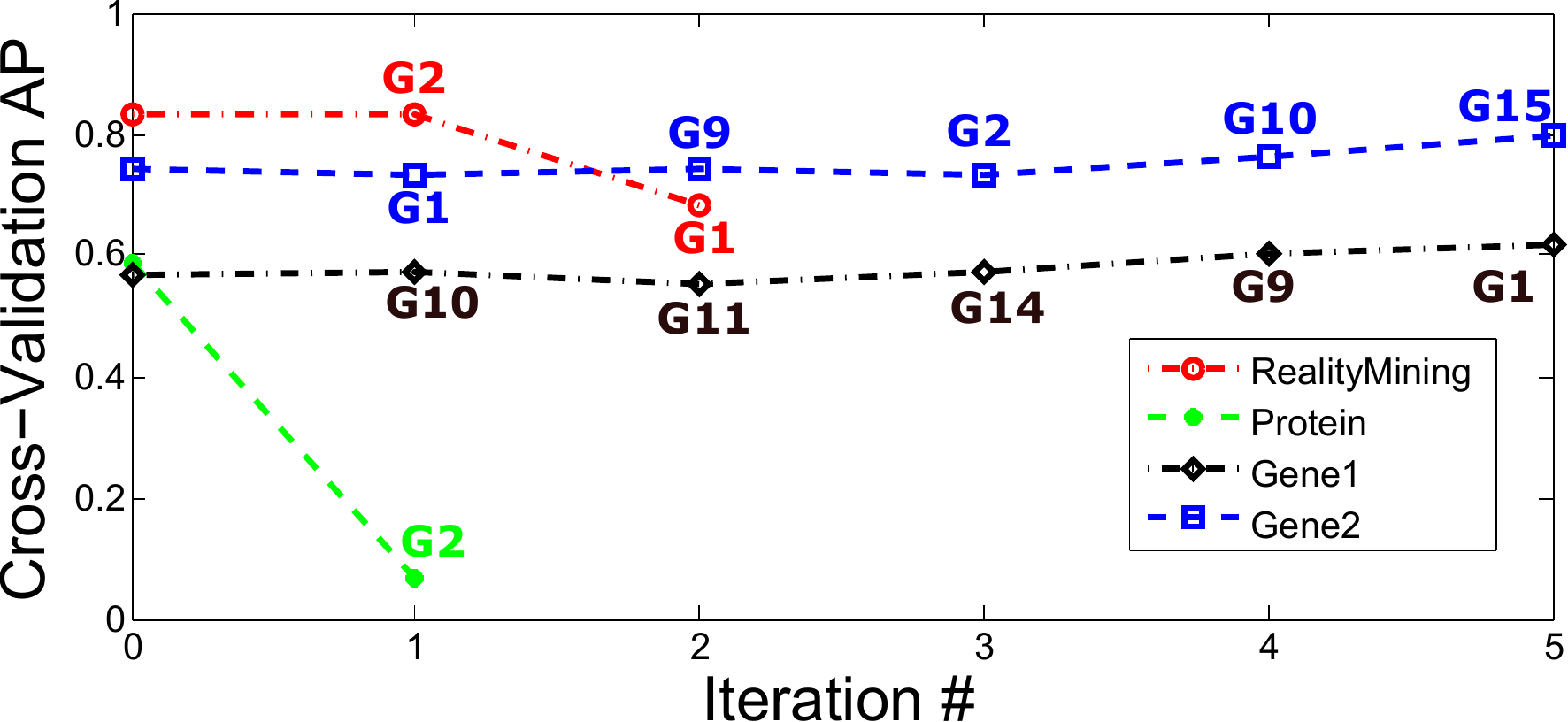} \\
	\end{tabular}
\end{minipage}
	\vspace{-0.1in}
	\begin{minipage}{1\textwidth}
	\caption{The removal order of individual graphs from real-world multi-graphs during the course of \method, which removes $G_x$ from {\em RealityMining} (See A.1) and $G_y$ from {\em Protein} (See A.2). It also removes 5 (uninformative) graphs each from both {\em Gene1} (A.3) and {\em Gene2} (A.4), where the cross-validation performance increases slightly.
				\label{fig:order}}
			\end{minipage}
\end{figure}

\clearpage
\newpage

\begin{minipage}{1\textwidth}
\section{\large{Inferred Weights on All Datasets}} 
\label{sec:weights}

In the following we provide the inferred weights by all the six methods on all the ten datasets as listed in Table \ref{tab:allresults}.
\end{minipage}

\begin{figure}[H]
	\begin{tabular}{cc}
		\includegraphics[width=0.85\linewidth,height=1.3in]{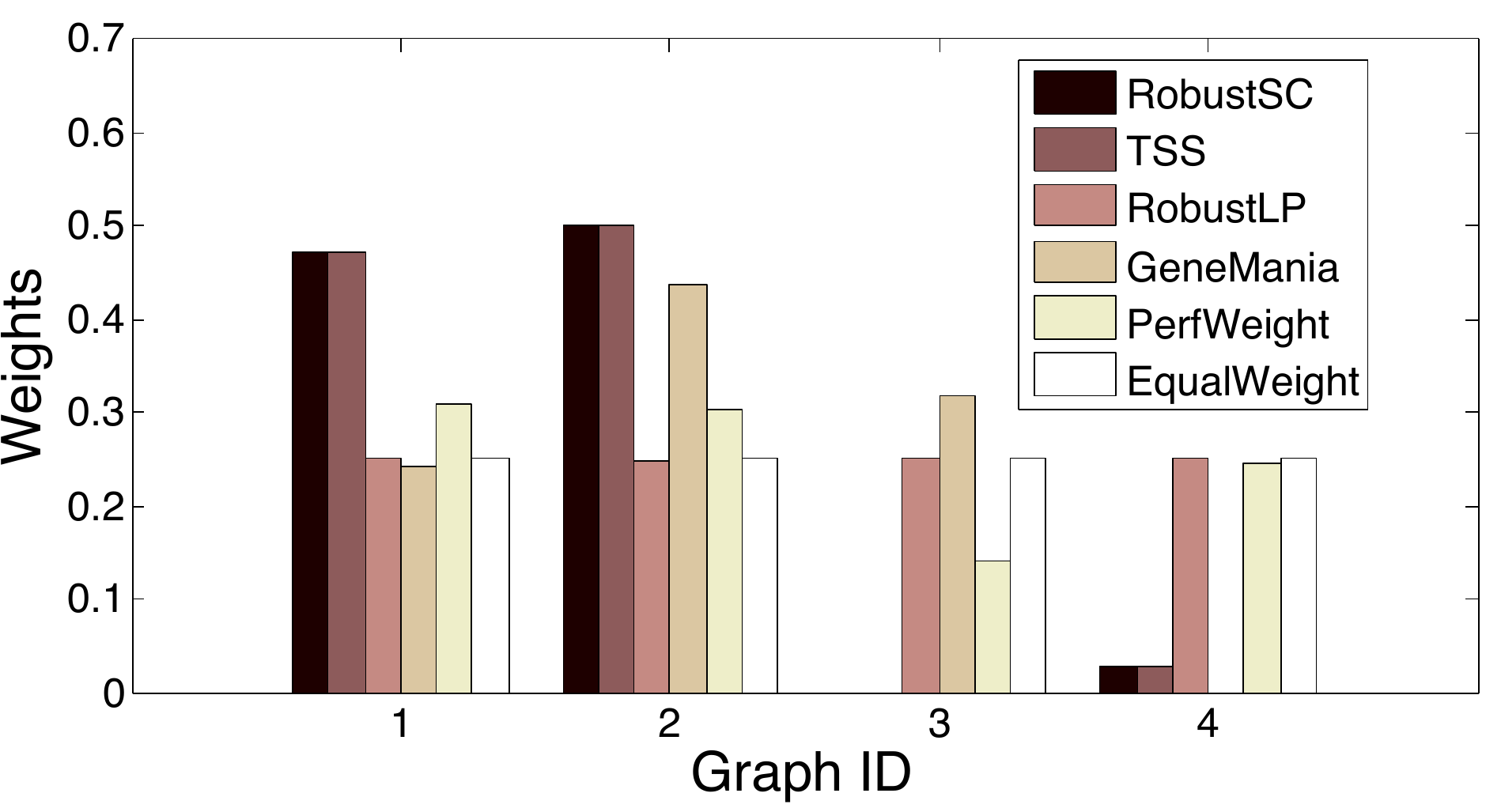} &
		\includegraphics[width=0.85\linewidth,height=1.3in]{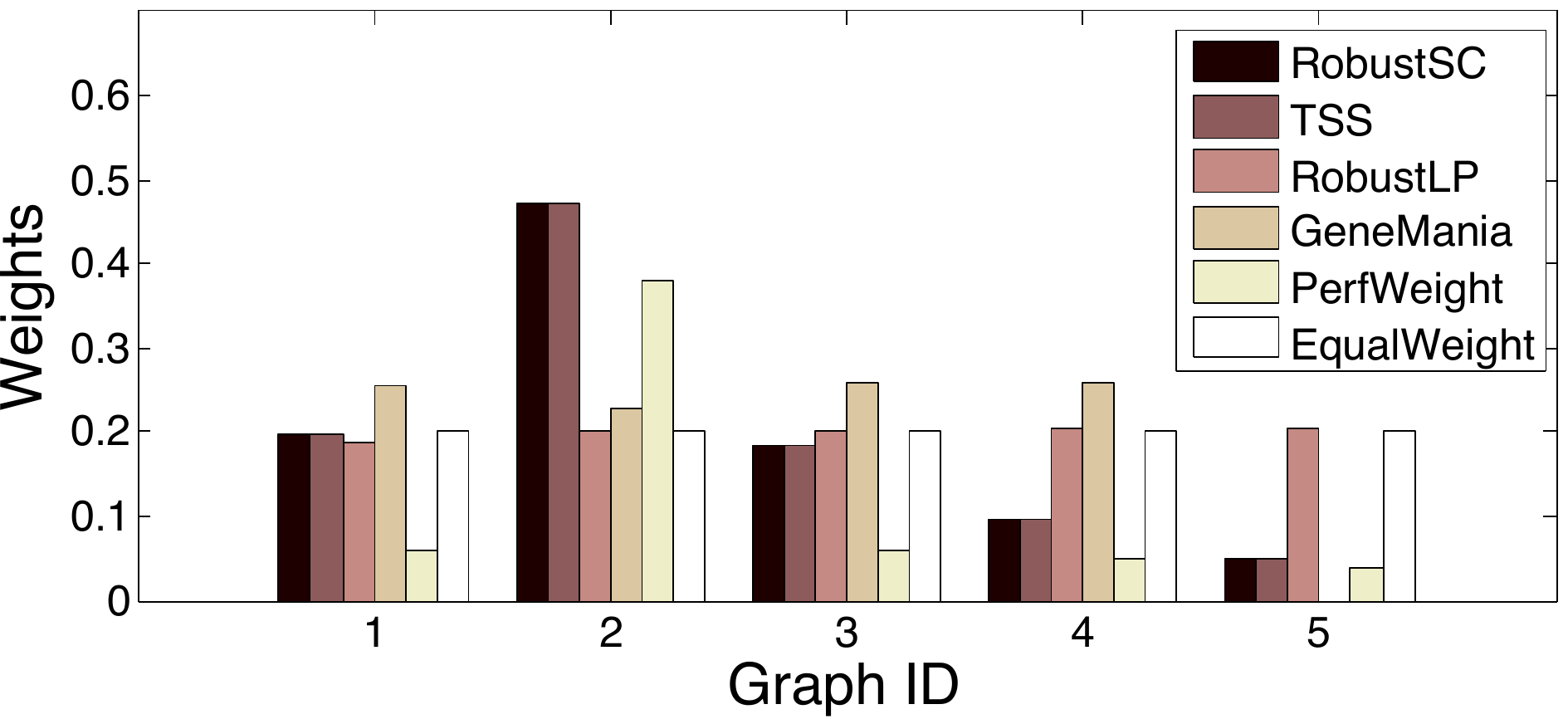} 		\\
		(1) {\em RealityMining} -- no noise	& 
		(2) {\em Protein} -- no noise \\	\\	
				\includegraphics[width=0.99\linewidth,height=1.3in]{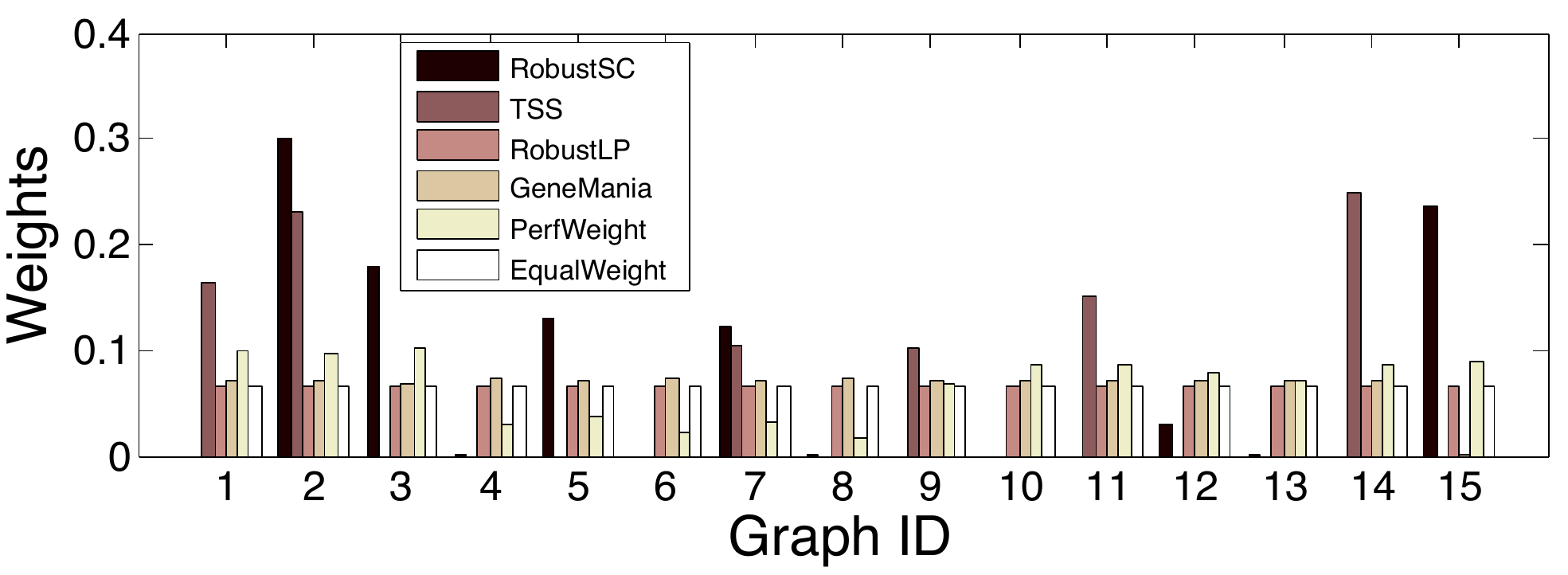} &
				\includegraphics[width=0.99\linewidth,height=1.3in]{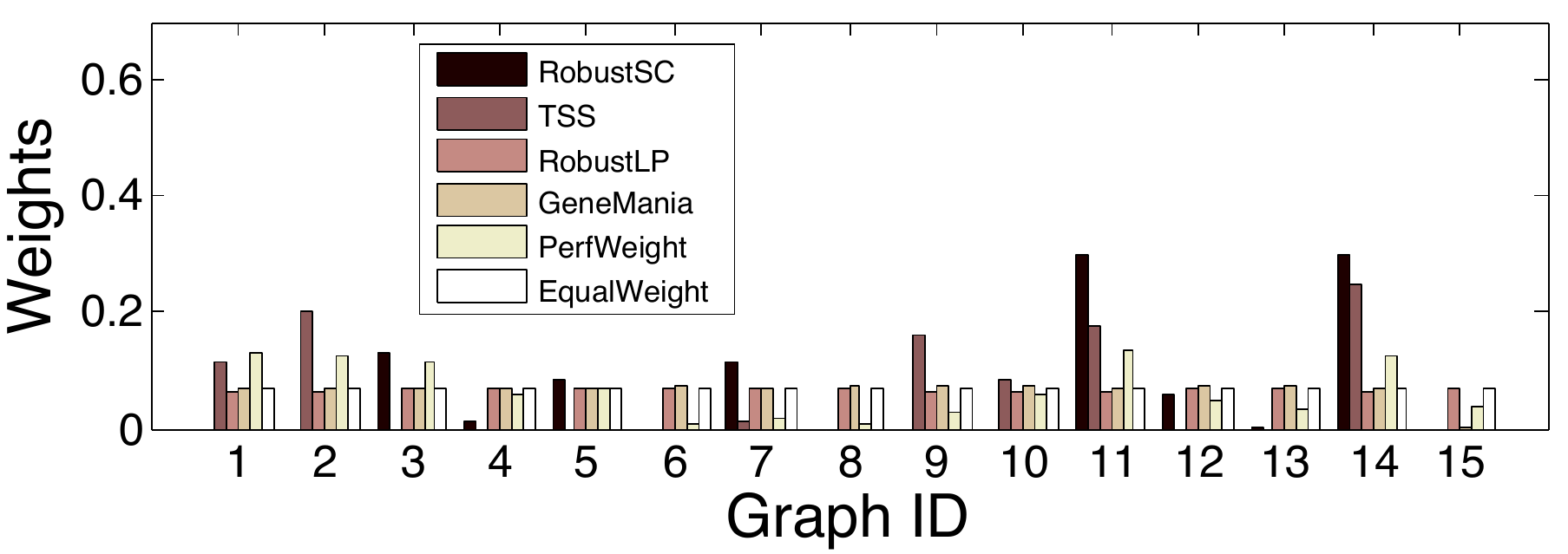} \\
				(3) {\em Gene1} -- no noise	& 
				(4) {\em Gene2} -- no noise \\	\\
					\includegraphics[width=0.99\linewidth,height=1.3in]{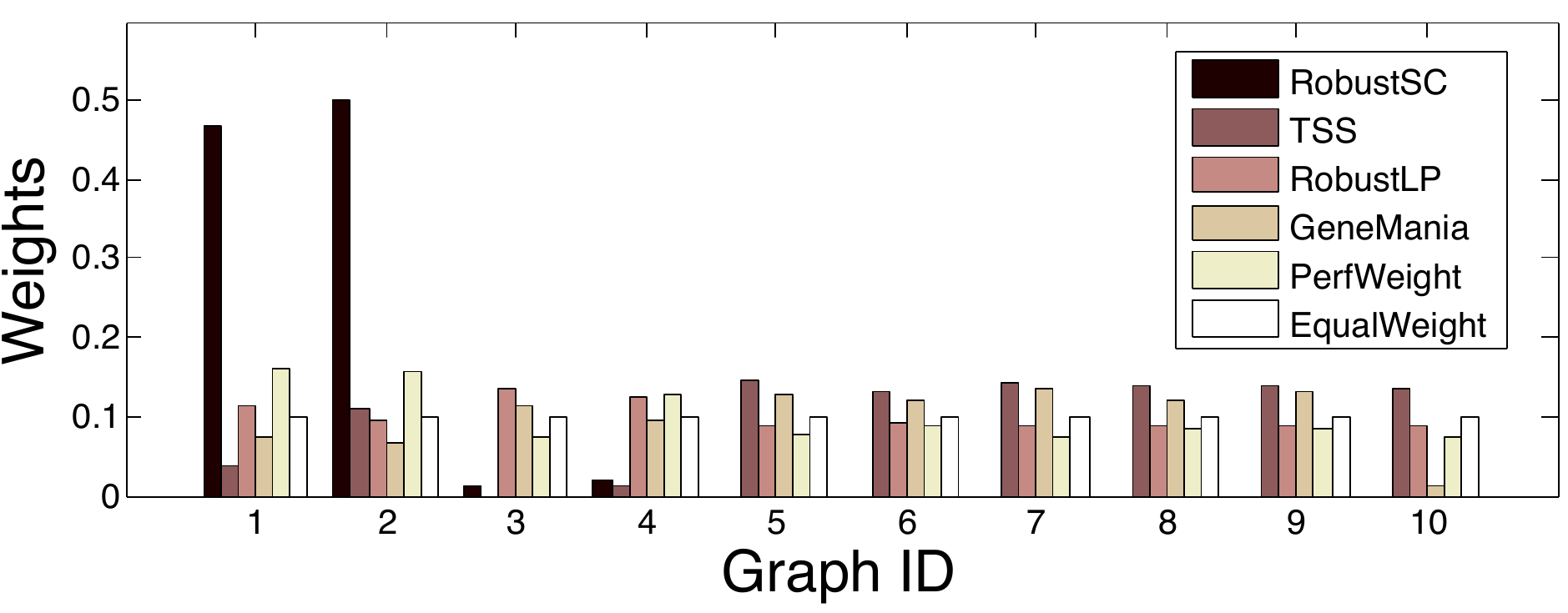} &
					\includegraphics[width=0.99\linewidth,height=1.3in]{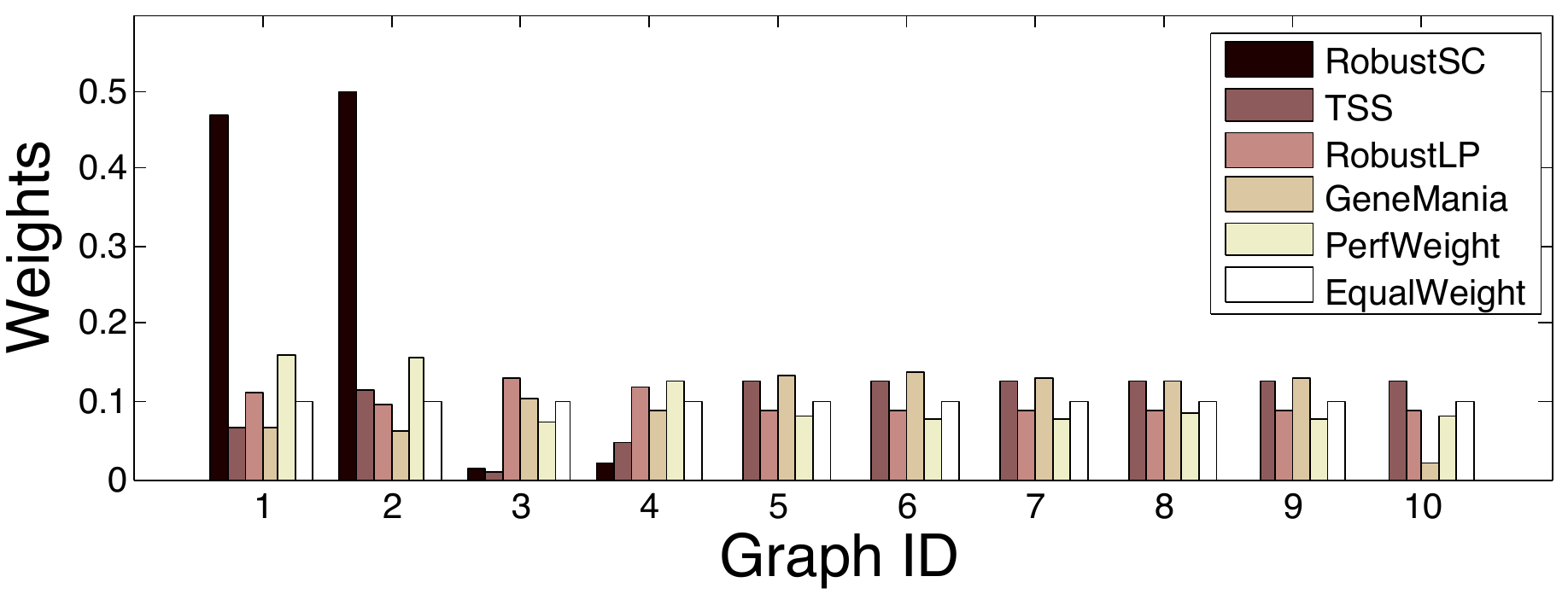} \\
					(5) {\em RealityMining} -- 6 injected, under ER and low intensity	&	
					(6) {\em RealityMining} -- 6 injected, under ER and high intensity	\\	
					\includegraphics[width=0.99\linewidth,height=1.3in]{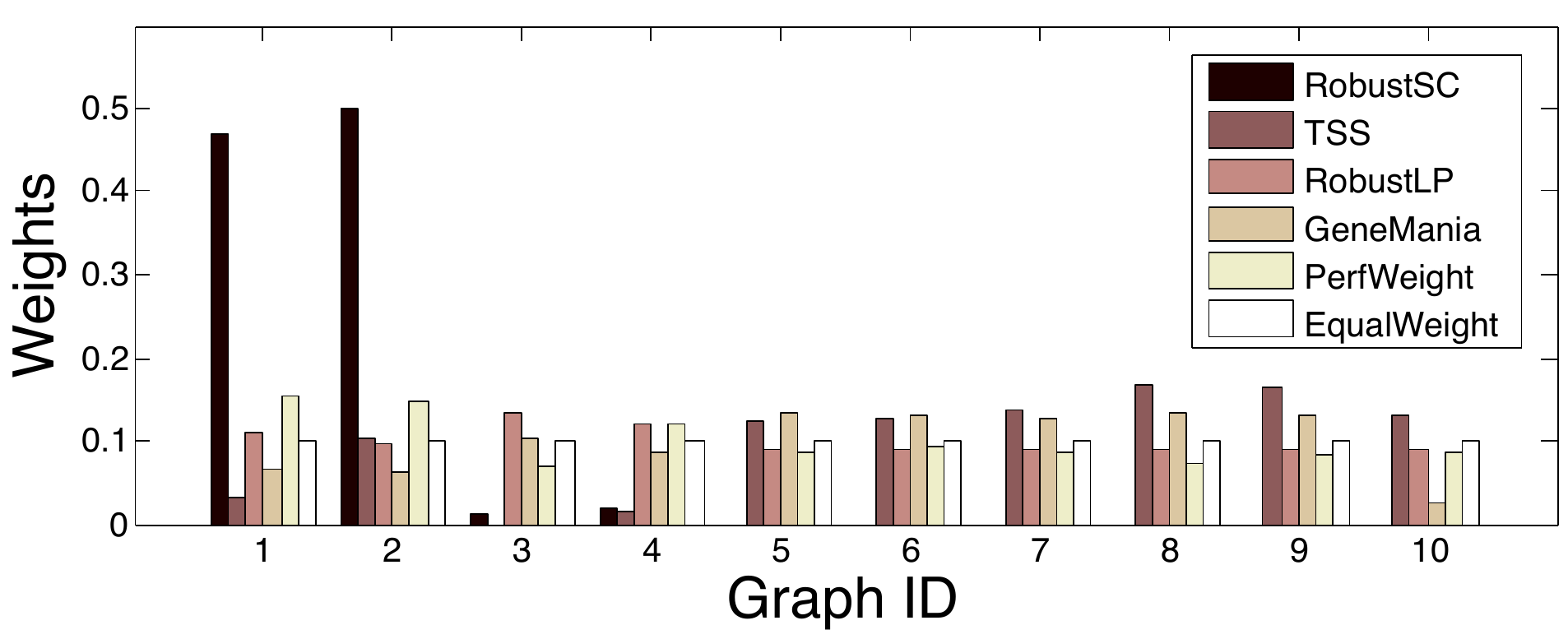} &
					\includegraphics[width=0.99\linewidth,height=1.3in]{FIG/perf/weights/RM-6-AV-Int2.pdf} \\
					(7) {\em RealityMining} -- 6 injected, under AV and low intensity	&
					(8) {\em RealityMining} -- 6 injected, under AV and high intensity	\\	
					\includegraphics[width=0.99\linewidth,height=1.3in]{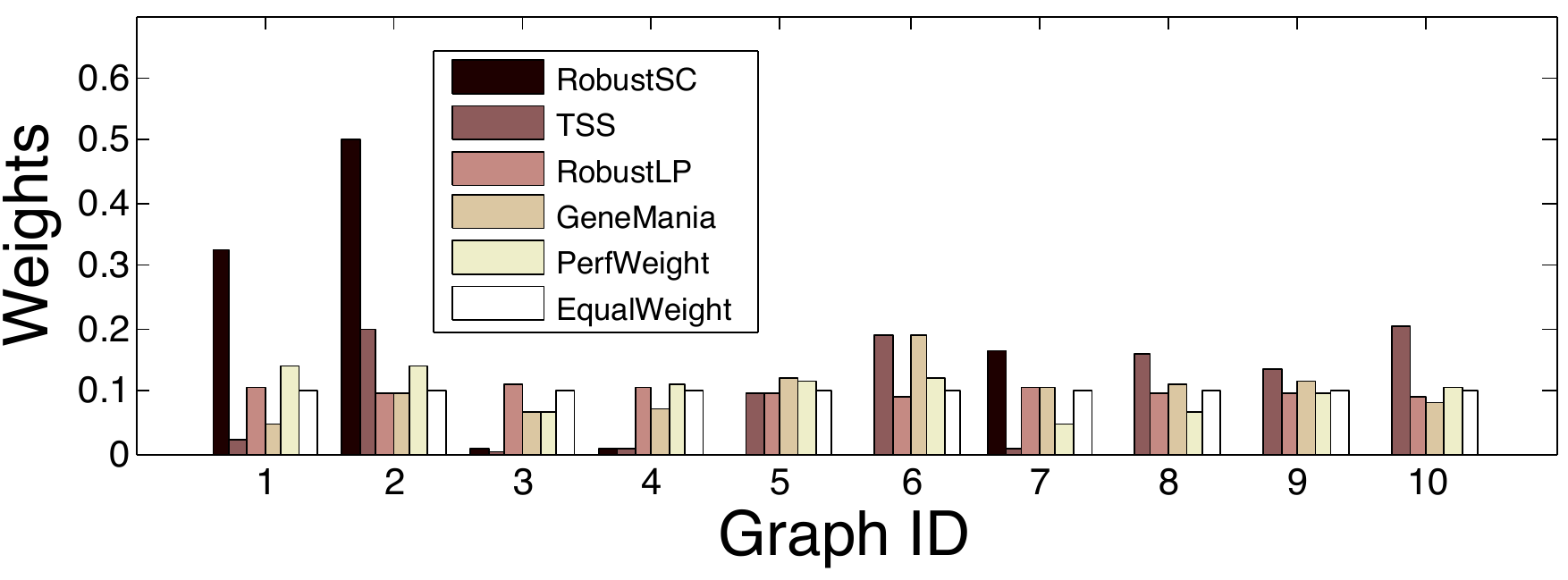} &
					\includegraphics[width=0.99\linewidth,height=1.3in]{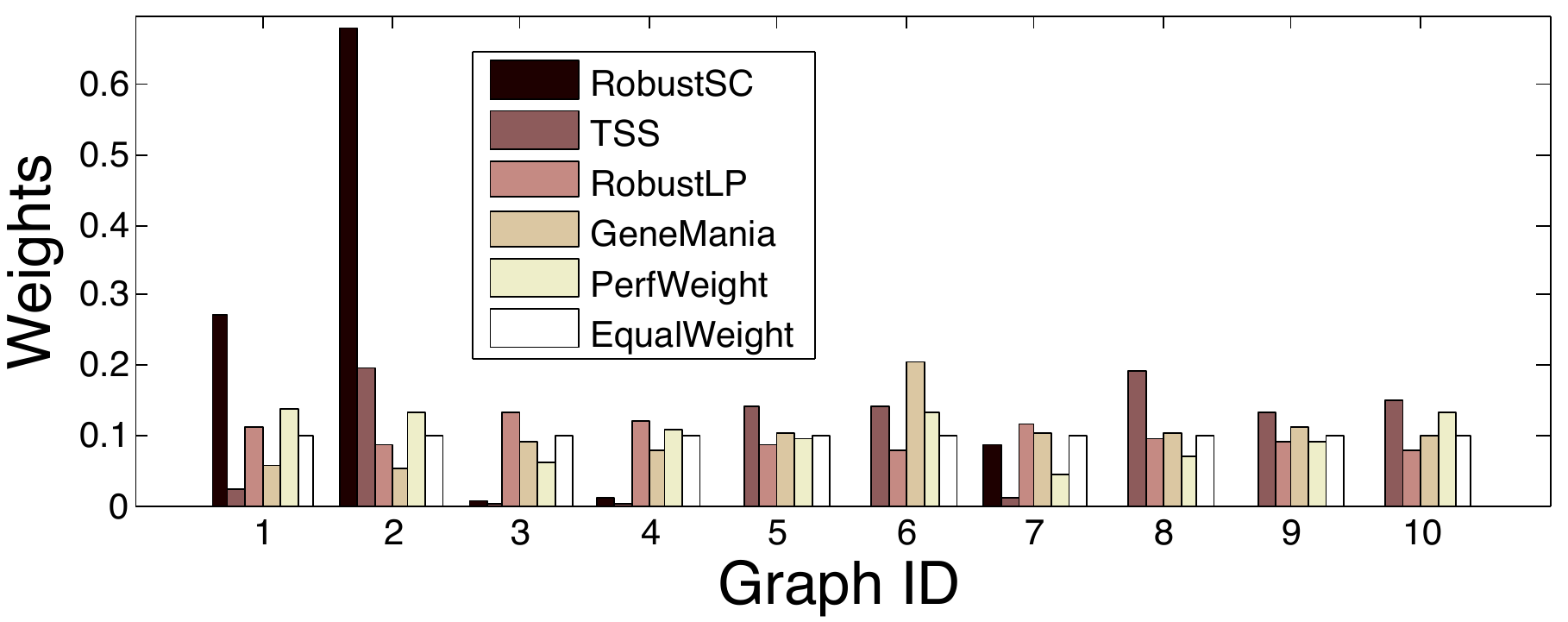} \\
					(9) {\em RealityMining} -- 6 injected, under RW and low intensity	&		
					(10) {\em RealityMining} -- 6 injected, under RW and high intensity	\\	
								
				\end{tabular}
	\vspace{-0.1in}
	\begin{minipage}{1\textwidth}
	\caption{{Inferred graph weights (normalized) on all ten datasets by all six methods.}\label{fig:allweights}}
	\end{minipage}	
\end{figure}

\end{document}